\documentclass[myepj-spec,pdftex]{mySvjour}
\usepackage{graphicx}
\usepackage[pdfpagemode=UseNone]{hyperref}
\usepackage{color}
\usepackage{xspace}
\usepackage[square,sort&compress,numbers]{natbib}   
\usepackage{verbatim}
\usepackage{amsmath,amssymb,amsfonts}
\usepackage{tabularx}
\usepackage{multicol}
\usepackage{footnote}
\usepackage{listings}
\usepackage[gen]{eurosym}
\usepackage[switch, modulo]{lineno}
\usepackage{longtable}
\usepackage[utf8]{inputenc}
\usepackage[T1]{fontenc} 
\usepackage{hyperref} 
\usepackage{url} 
\usepackage{booktabs} 
\usepackage{amsfonts}  
\usepackage{nicefrac} 
\usepackage{microtype}
\usepackage{xcolor}
\usepackage{graphicx}
\usepackage{float}
\usepackage{caption}
\usepackage{subcaption}
\usepackage{amsmath}

\usepackage{listings}
\usepackage{xcolor}

\lstdefinestyle{mystyle}{
    backgroundcolor=\color{white},
    basicstyle=\ttfamily\small,
    breaklines=true,
    captionpos=b,
    columns=flexible,
    numbers=left,
    numbersep=5pt,
    frame=single,
    language=Python,
    numberstyle=\tiny,
    keywordstyle=\color{blue},
    commentstyle=\color{green},
    stringstyle=\color{red},
}

\usepackage{lmodern,bm}                
\usepackage[T1]{sansmath} 
\SetMathAlphabet{\mathsfbf}{sans}{\sansmathencoding}{\sfdefault}{bx}{sl}
\usepackage{etoolbox}
\AtBeginEnvironment{sansmath}{}{}{}

\usepackage{lipsum}

\definecolor{darkblue1}{rgb}{0,0,.2}
\definecolor{darkblue}{rgb}{0,0,.2}
\definecolor{darkred}{rgb}{0.5,0,0}
\pagecolor{white}
\color{black}
\hypersetup{breaklinks=true, 
	colorlinks=true, 
	linkcolor=darkblue1, 
	menucolor=darkblue1, 
	urlcolor=darkblue1,
	citecolor=darkblue1,
	pdftitle={},
	pdfauthor={},
	pdfsubject={},
	pdfkeywords={},
	pdfproducer={}
}

\usepackage{siunitx}

\newcommand{\bi}{\begin{itemize}}
\newcommand{\ei}{\end{itemize}}

\newcommand{\ben}{\begin{enumerate}}
\newcommand{\een}{\end{enumerate}} 

\newcommand{\bt}[1]{\begin{table}[tb]\begin{tabular}{#1} \hline\hline  \\[-1.0em]}
\newcommand{\et}[2]{\hline\hline \end{tabular} \caption{#1} \label{#2} \end{table}}

\newcommand{\be}{\begin{equation}}
\newcommand{\ee}{\end{equation}}

\newcommand{\bea}{\begin{eqnarray}}
\newcommand{\eea}{\end{eqnarray}}

\newcommand{\bc}{}

\newcommand{\mev}{\ensuremath{\mathrm{\,Me\kern -0.1em V}}\xspace}
\newcommand{\gev}{\ensuremath{\mathrm{\,Ge\kern -0.1em V}}\xspace}

\begin{document}
			
			\begin{flushright}
				\normalsize
			\end{flushright}
			
			\vspace{-2cm}
			
			\title{\Large\boldmath Enforcing Fundamental Relations via Adversarial Attacks on Input Parameter Correlations}
			%

\author{AI Safety Project Group (Lucie Flek$^2$ \and Alexander Jung$^4$ \and Akbar Karimi$^2$ \and \underline{Timo Saala$^1$} \and Alexander Schmidt$^4$ \and Matthias Schott$^1$ \and Philipp Soldin$^3$ \and Christopher Wiebusch$^3$)}
\institute{Institute of Physics, University of Bonn, Germany 
\and Bonn-Aachen Institute of Technology, University of Bonn, Germany
\and Institute of Experimental physics III B, RWTH Aachen University, Germany
\and Institute of Experimental physics III A, RWTH Aachen University, Germany}

			
			\abstract{Correlations between input parameters play a crucial role in many scientific classification tasks, since these are often related to fundamental laws of nature. For example, in high energy physics, one of the common deep learning use-cases is the classification of signal and background processes in particle collisions. In many such cases, the fundamental principles of the correlations between observables are often better understood than the actual distributions of the observables themselves. In this work, we present a new adversarial attack algorithm called \textit{Random Distribution Shuffle Attack (RDSA)}, emphasizing the correlations between observables in the network rather than individual feature characteristics. Correct application of the proposed novel attack can result in a significant improvement in classification performance - particularly in the context of data augmentation - when using the generated adversaries within adversarial training. Given that correlations between input features are also crucial in many other disciplines. We demonstrate the RDSA effectiveness on six classification tasks, including two particle collision challenges (using CERN Open Data), hand-written digit recognition (MNIST784), human activity recognition (HAR), weather forecasting (Rain in Australia), and ICU patient mortality (MIMIC-IV), demonstrating a general use case beyond fundamental physics for this new type of adversarial attack algorithms.}	
	\maketitle

\tableofcontents

\section{Introduction}

Deep learning models are widely used in high energy physics, from detector simulations \cite{adelmann2022new} and object reconstruction \cite{Ngairangbam_2020} to triggering events at the Large Hadron Collider \cite{Migliorini_2022}. In recent years, adversarial learning techniques have also found their way into fundamental physics to improve network robustness by generating minimally altered data that lead to misclassification. These techniques have been successfully employed in other domains \cite{szegedy2014intriguing}. Common adversarial algorithms include the Fast Gradient Sign Method (FGSM) \cite{goodfellow2015explaining}, Deep Fool \cite{moosavidezfooli2016deepfool}, and Projected Gradient Descent (PGD) \cite{madry2019deep}. FGSM and PGD create adversarial examples by exploiting gradient information in the network's last layer to perturb inputs minimally. These attacks aim to maximize task error while minimizing perceived perturbation, often quantified using norms like $L_\infty$, $L_2$, or $L_1$. Retraining networks with adversarial examples can lead to improvements in robustness and classification performance \cite{goodfellow2015explainingharnessingadversarialexamples}.

\begin{figure}[thb!]
  \centering
  \begin{minipage}[b]{0.49\textwidth}
    \includegraphics[width=\textwidth]{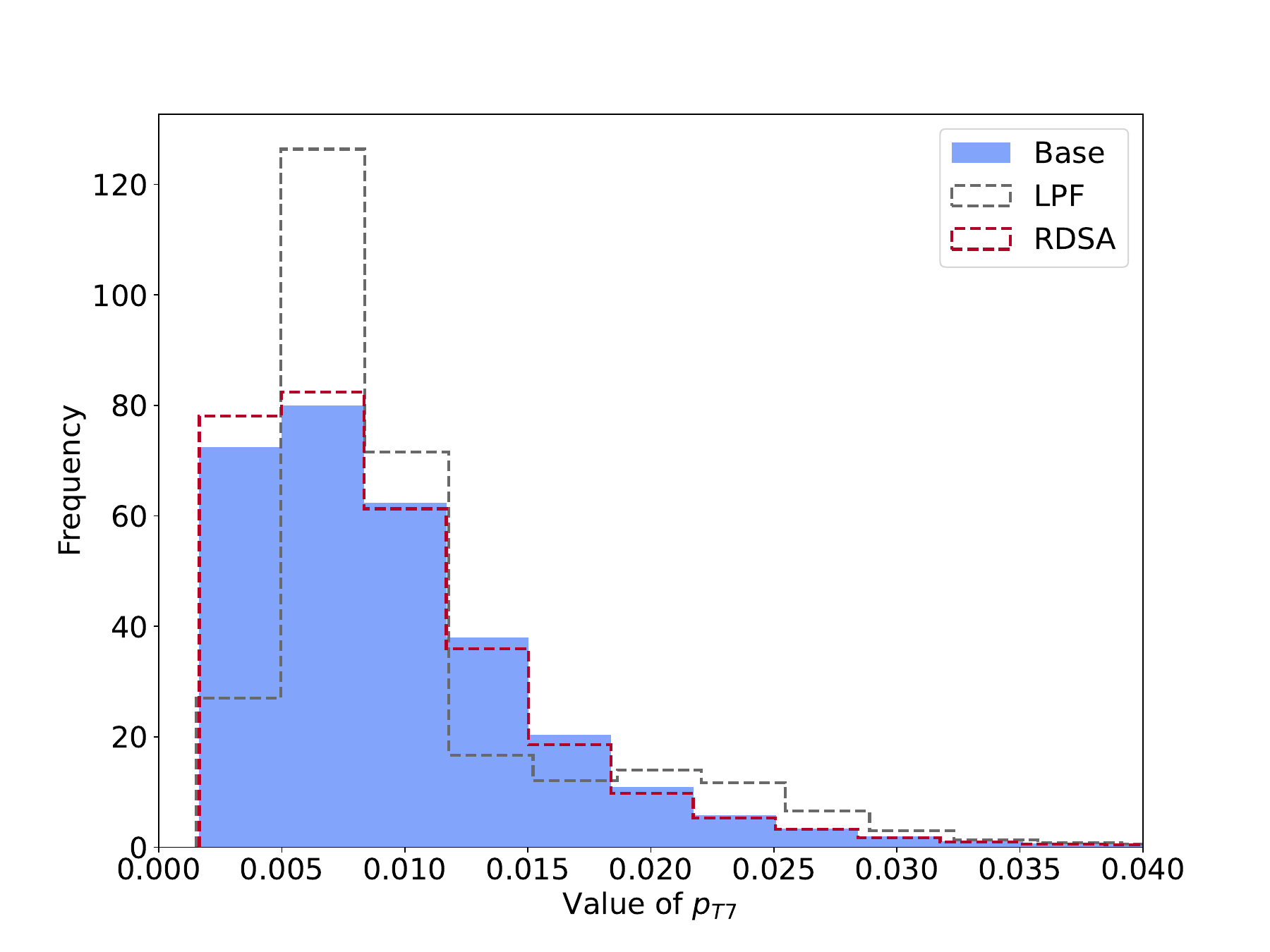}
    \newline 
    \caption{Frequency distribution of one input feature after pre-processing of the original training data (blue), of the adversarial sets generated with LPF (grey), and adversarial sets generated using RDSA (red).}
    \label{fig:DistributionsCompare}
  \end{minipage}
  \hfill
  \begin{minipage}[b]{0.45\textwidth}
    \includegraphics[width=\textwidth]{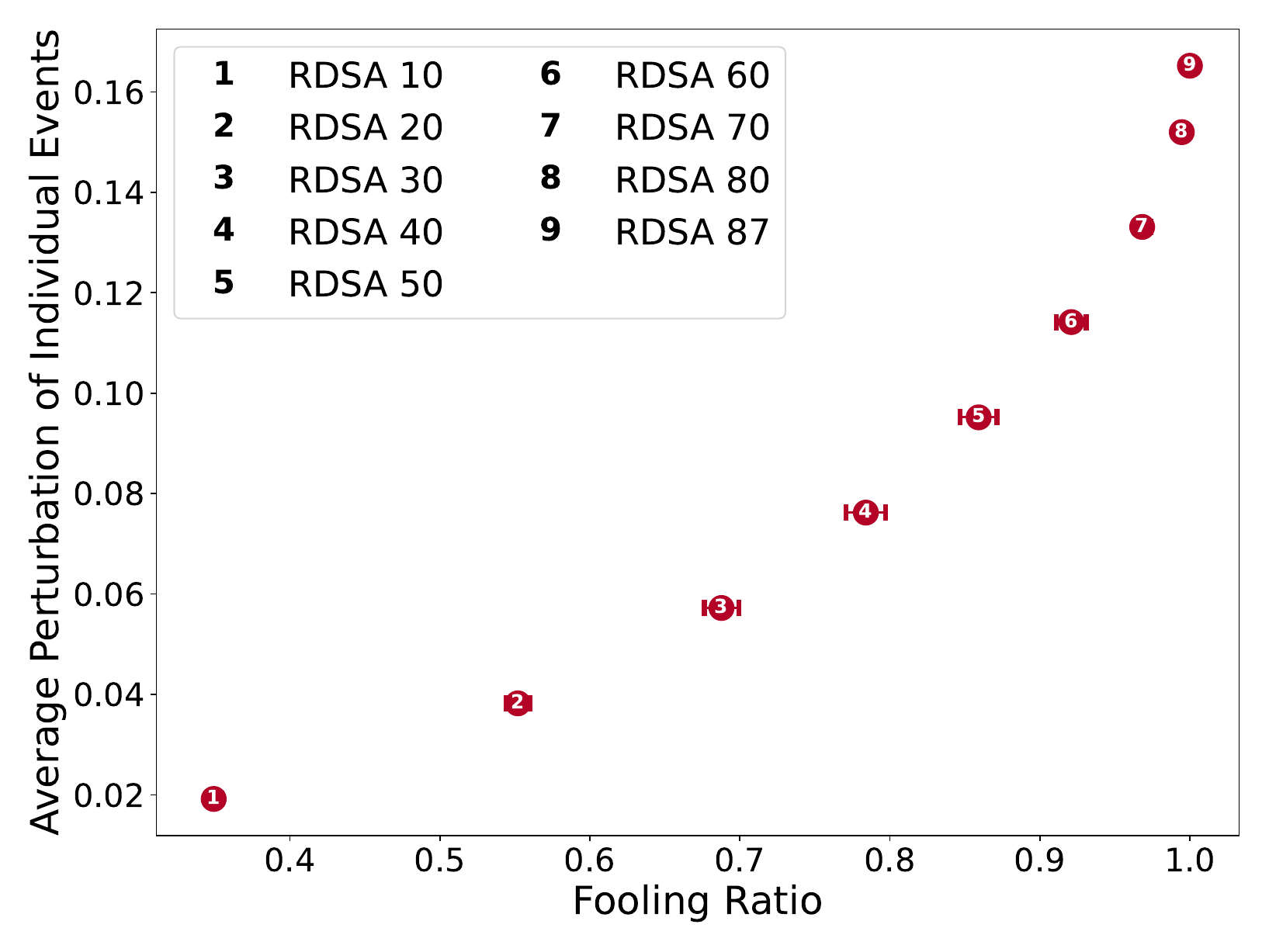}
    \newline 
    \caption{Average difference between individual clean inputs (test set) and corresponding adversaries for the TopoDNN example model for varying amounts of variables perturbed (10, 20, ...). }
    \label{fig:Event_Diff_FR_RDSA}
  \end{minipage}
\end{figure}

To understand the structure of data sets in the context of high energy physics, it is illustrative to discuss some basic aspects of classification tasks in this domain. Typical data analyses in the context of fundamental physics rely on multiple real-valued input parameters, each with unique physical interpretation, such as a particle's energy or flight direction. In this context, a collision event can be viewed as an instance in a machine learning setting, where the observables measured from the collision (e.g., energy, momentum, transverse momentum) serve as the input features that describe the instance. For example, Figure \ref{fig:DistributionsCompare} shows the transverse momentum ($p_T$) distribution of a particle, recorded by the CMS detector \cite{CMS:2008xjf} at the Large Hadron Collider \cite{Evans:2008zzb}. Deep learning methods offer a powerful way to analyze such data, where neural networks process these particle collisions (instances) by using the values of several observables (input features). A single instance is thus constructed by gathering values of several input features (observables) from a single particle collision. The distribution of each input feature (e.g., the transverse momentum $p_T$) can then be studied across the entire dataset, such as the training data. Adversarial algorithms typically introduce noticeable changes to the distributions of these features and can affect classification results. Figure \ref{fig:DistributionsCompare} shows as example the transverse momentum distribution in adversarial samples generated using the LowProFool (LPF) algorithm.

While adversarial methods often target and perturb individual features - resulting in significant changes to the underlying feature distributions - the correlations between features are equally (or even more) important. For instance, the relationship between a particle’s mass and the momentum of its decay products is governed by energy conservation. These correlations are fixed by physical principles and hold independently from the actuall mass value. Similarly, correlations are critical in other domains, such as medicine (e.g., between symptoms and diagnoses) or predicting outcomes in weather forecasting. However, most adversarial algorithms overlook these relationships, focusing on altering independent feature distributions, rather than their fundamental correlations.

To address this, we hypothesize that neural network classifiers can gain more robustness by focusing on the correlations between input variables rather than individual feature distributions. To test this, we introduce the Random Distribution Shuffle Attack (RDSA), a novel adversarial attack that generates examples with minimal changes to one-dimensional feature distributions while significantly altering correlations between observables. Adding these adversarial examples to the training set thus minimally affects the one-dimensional distributions, but significantly alters the global correlations between input features, encouraging the model to focus more on these correlations.

We evaluate RDSA on two primary goals: (1) fooling fully trained networks and (2) leveraging adversarial examples generated by the attack as a data augmentation method. Our experiments include six benchmark neural network models across diverse domains: high-energy physics, weather forecasting, handwritten digit recognition, human activity recognition, and medicine. Results indicate that RDSA effectively deceives networks while maintaining minimal alterations to one-dimensional distributions. Moreover, applying RDSA as a data augmentation technique proves competitive with state-of-the-art methods for tabular data.

\section{Related Work}

Adversarial attack algorithms have been extensively studied, particularly in safety-critical sectors such as image recognition for self-driving cars \cite{kong2021physgan}.
Unlike the Random Distribution Shuffle Attack (RDSA) presented in this work, current state-of-the-art attack algorithms such as the Projected Gradient Descent (PGD) \cite{madry2019deep} or LowProFool (LPF) \cite{ballet2019imperceptibleadversarialattackstabular} typically generate adversaries by leveraging the gradient of the loss function with respect to a given input, applying malicious perturbations based on the sign of this gradient. PGD and LPF are optimized to minimize the perceived changes - typically measured using norms — of a single input to the classifier. In contrast, RDSA is designed to minimize changes to the underlying one-dimensional input feature distributions.

Data augmentation involves creating synthetic data to increase sample sizes for training deep neural networks. Among various methods, two state-of-the-art approaches for generating synthetic tabular data are TVAE and CTGAN \cite{xu2019modeling}. CTGAN, a Conditional Tabular Generative Adversarial Network, generates data using mode-specific normalization and addresses data imbalance through a conditional generator and training-by-sampling. TVAE, a Tabular Variational Auto-Encoder, directly leverages the data to build the final generator. Both methods rely solely on the provided data. In contrast, our proposed data augmentation method using the Random Distribution Shuffle Attack (RDSA) incorporates both the data and the deep learning model. Traditional synthesizers, like TVAE and CTGAN, aim to create samples that closely resemble the initial dataset. Using adversaries for augmentation introduces worst-case samples into the training process with corrected labels, enhancing robustness. Additionally, while TVAE and CTGAN potentially alter the one-dimensional input feature distributions noticeably, the RDSA approach does not.

\section{Method: Random Distribution Shuffle Attack}

The goal of the new adversarial attack algorithm is to generate training examples that leave the one-dimensional distributions of all input parameters unchanged but alter the correlations among them to maximize changes in the network output, resulting in incorrect classifications. Using adversaries generated with this approach in order to adversarially train a neural network forces the network to focus on the differing correlations between classification classes since the one-dimensional distributions remain consistent.

The \textit{Random Distribution Shuffle Attack} (RDSA) method involves the following steps: First, represent the one-dimensional distributions of each variable as finely binned histograms, where the y-axis indicates the frequency of values within each bin. The bin size is chosen such that the statistical uncertainties in each bin is of roughly a similar size\footnote{This choice is not possible for non-continuous distributions.}. These histograms are calculated over a complete dataset, such as the training or testing dataset. Then, for each input to be perturbed, the values for each feature are resampled based on these distributions, using the frequencies as probabilities. This shuffling process preserves the original distributions but reduces the correlations among variables. The algorithm attempts this reshuffling up to a predefined maximum number of tries, $N_s$, stopping early if an adversarial example is found. After each shuffling step, where every relevant feature of a single input is shuffled according to the approach defined above, we query the model with the resulting shuffled input. If this query returns a class that is different from the initial class, we can conclude that the shuffled input is an adversary to our model.

RDSA can be extended to multiple input variables by specifying the number of variables to shuffle ($n_{Vars}$) globally. For example, with $n_{Vars}$ set to three, only three variables will be reshuffled for each input. To avoid biases, these variables are randomly selected. If a model has eight input variables and $n_{Vars}$ is set to three, three out of the eight variables are randomly shuffled for each perturbed input. The pseudo-code for RDSA is provided in the technical appendix.

After applying this attack, the changes to the one-dimensional variable distributions remain minimal, but the correlations among shuffled variables decrease, approaching zero. This occurs because random shuffling introduces randomness into the relationships between input features, reducing their absolute correlations.

\section{Data Sets, Tasks and Classification Models Used for the Performance Evaluation}

We first evaluate the RDSA algorithm on two common particle physics classification tasks using publicly available data from the CERN Open Data initiative \cite{CERNOpenData, VBFSignal, VBFBGLplus, VBFBGWpWm, VBFBGWpWm2, VBFBGZJet, TTJets, WWJets}, specifically from the CMS experiment. A table summarizing the input features used, how many are continuous, the trainable parameters found in the applied deep learning networks, as well as the output dimensions of said models can be seen in Table \ref{tab:model_summary}.

\begin{table}[htb!]
\centering
\begin{tabular}{|l|c|c|c|c|}
\hline
\textbf{Task/Model} & \textbf{Input Features} & \textbf{Continuous Features} & \textbf{Trainable Parameters} & \textbf{Output} \\
\hline
VBF Model & 8 & 8 & 210 & 2 (binary) \\
\hline
TopoDNN Model & 87 & 87 & 59,263 & 2 (binary) \\
\hline
Weather Forecasting & 21 & 9 & 1,421 & 2 (binary) \\
\hline
Hand-Written Recognition & 784 & 560 & 111,514 & 10 (digits 0-9) \\
\hline
Human Activity Recognition & 561 & 548 & 82,902 & 6 (activities) \\
\hline
ICU Mortality Prediction & 153 & 33 & 65,093 & 2 (binary) \\
\hline
\end{tabular}
\caption{Summary of models, their input features, continuous features, trainable parameters, and output dimensions.}
\label{tab:model_summary}
\end{table}

The first model distinguishes between two physics processes at the LHC, identifying invisible Higgs boson decays via vector boson fusion. In this case, the models input features are variables derived from the kinematics of the two jets with the highest momentum - called dijet - as well as variables derived from the collisions missing transverse energy. More details of the underlying physics and network architecture are summarized in Ref. \cite{Ngairangbam_2020}. This model, referred to as the \textsc{VBF} model, is a simple feedforward neural network with 210 trainable parameters, taking eight continuous float values as input and producing a binary output to distinguish signal from background processes.

The second model, the \textsc{TopoDNN} model, is more complex and classifies different types of particle jets in proton-proton collisions. Specifically, here we try to classify between two types of particle jets, namely Top-Top-Jets (TTJets) and Jets originating from two W-Bosons, called WWJets. More details about the underlying physics and data are described in Ref. \cite{Kasieczka_2019}. It uses 87 input features consisting of continouous variables pertaining to the underlying jet particles kinematics and has 59,263 trainable parameters.

To illustrate how correlations between input features are common in other research fields, we additionally tested RDSA on a weather forecasting, hand-written digit recognition, human activity recognition, and an ICU mortality prediction tasks. For the weather forecasting model, a standard feed-forward neural network, predicts whether it will rain in Australia the next day. Trained on the \textit{Rain in Australia} Kaggle dataset \cite{RainKaggle}, it has 21 input features - of which 9 are continuous - and 1,421 trainable parameters. For this network, the input features contain meteorological information, such as the minimal and maximal temperature of the current day.

The hand-written recognition model is tested on a modified version of the MNIST dataset \cite{deng2012mnist}, called MNIST784. This dataset contains flattened versions of the 28x28 images found in MNIST. As the name implies, the data consists of a total of 784 features per input, of which we consider 560 to be continuous (the remaining values are either always 0, or almost always 0). The model applied to recognize these flattened hand-written digits has 111,514 trainable parameters.

For the human activity recognition task, the HAR dataset \cite{s20082200} is used. This dataset contains 561 features, of which 548 are continuous. These features contain time and frequency domain variables derived from sensor signals of both the accelerometer and the gyroscope of smartphones carried on the waist of experimental subjects performing six distinct activities (walking, walking upstairs, walking downstairs, sitting, standing, and laying). The model used to recognize the activity of the experiment participants contains 82,902 trainable parameters.

The medical model, based on data from the MIMIC-IV database \cite{MIMICIV, Johnson2023, PhysioNet}, predicts ICU patient mortality. The MIMIC-IV data pipeline \cite{gupta2022extensive} was used to preprocess the version 2.0 dataset, resulting in 153 input features - of which 33 are continuous - from 57,712 samples with 65,093 trainable parameters. This model uses the mean values of 24-hour time-series data - containing information such as the medication given to the patient, or some diagnoses made - in 2-hour steps, simplifying compatibility with the RDSA approach and the reference LPF attack. For all of the previously mentioned models, a full list of input features and a brief description of the architectures and hyperparameters are provided in the appendix.

\section{Impact of the Random Distribution Shuffle Attack}

The impact of RDSA on all six models is evaluated using four metrics. First, the Fooling Ratio (FR), which measures the ratio of previously correctly classified inputs that are misclassified after perturbation, is used to assess the effectiveness of the adversarial attacks. The uncertainty of the FR is estimated as the standard deviation over ten attack runs for each model. 

Second, in order to quantify the perturbation applied by the adversarial algorithms, we define the \textit{mean change of the input features} as the average change applied to each event:

\[\langle c_f \rangle := \dfrac{1}{N} \sum_{i=1}^{N} \Biggl[ \dfrac{1}{F} \sum_{j=1}^{F} |C_{ij} - A_{ij}| \Biggr] , \]

where $N$ is the number of samples, $F$ is the number of input features, $C$ represents clean inputs, and $A$ represents adversarial inputs. This metric calculates the average absolute difference between clean and adversarial inputs across all features and samples \footnote{Where applicable - e.g. for the VBF model - we map all input features to their respective z-scores as a form of normalization.}. While RDSA does not specifically optimize this metric, including it enables a comparison with other attack algorithms.

Given that RDSA specifically targets one-dimensional distributions, a third metric, the Jensen-Shannon Distance (JSD), is introduced to measure the similarity between original and attacked distributions. The Jensen-Shannon Distance is the square root of the Jensen-Shannon Divergence \cite{MENENDEZ1997307}, commonly used in machine learning to quantify the similarity between two distributions. It is defined as:

\[JSD := \sqrt{\dfrac{D(\Vec{p} \Vert \Vec{m}) + D(\Vec{q} \Vert \Vec{m})}{2}}, \]

where $\Vec{m}$ is the pointwise mean of $\Vec{p}$ and $\Vec{q}$, and $D$ is the Kullback-Leibler divergence. In this work, the SciPy implementation of JSD \cite{2020SciPy-NMeth} is used.

RDSA aims to change the correlations between input parameters to generate adversarial examples. Thus, the difference in the correlations between the clean and adversarial dataset is used as a final metric. This is calculated using the correlation matrices of both datasets:

\[\langle c_c \rangle := \dfrac{1}{F^2} \sum_{i=1}^{F} \sum_{j=1}^{F} |\boldsymbol{C_C} - \boldsymbol{C_A}|_{ij} ,\] 
where $F$ is the number of input features, $C_C$ is the correlation matrix of the clean dataset, and $C_A$ is the correlation matrix of the adversarial dataset. Subsequently, the element-wise absolute differences of these matrices are averaged. Small values indicate minor changes in correlations, while large values indicate significant changes.

\begin{figure}[!htb]
    \centering
    \includegraphics[width=\textwidth]{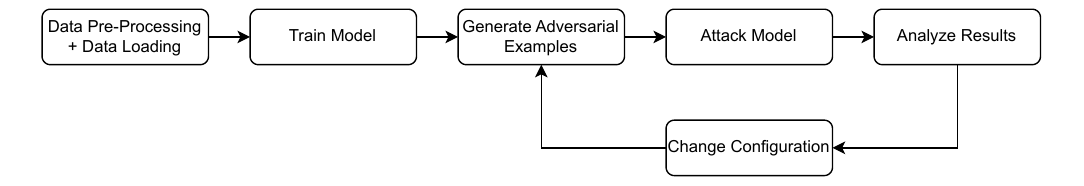}
    \caption{Flow chart visualizing the general attack pipeline.}
    \label{fig:AttackFlowChart}
\end{figure}

The general attack pipeline (visualized in Figure \ref{fig:AttackFlowChart}) is structured as follows: First, data pre-processing is performed, and the processed datasets — training, validation, and test — are fully loaded. Next, the model is trained using the complete training and validation datasets. After training, we generate adversarial examples using the full test dataset for all the attacks and their respective configurations. These adversarial examples are then used to evaluate the trained model's performance against each attack. Finally, the results are analyzed and compared across the different attack methods and configurations to gain insights into their effectiveness.

We run the Random Distribution Shuffle Attack (RDSA) on all six models with varying numbers of perturbed variables ($n_{Vars}$) per input, each for up to 100 shuffle attempts. The number of perturbed variables differed by model. For the \textsc{VBF} model, 1 to 8 variables were perturbed; for the \textsc{TopoDNN} model, 10 to 87 variables were perturbed in increments of 10 (and 7 for the last step). The \textsc{Rain in Australia} model had 1 to 9 variables perturbed, while the \textsc{MIMIC-IV} medical model perturbed 1 to 33 variables in increments of four. For MNIST784, 1 to 560 variables were perturbed (in increments of 99/100/160) and for HAR 1 to 548 were perturbed (increments of 99/100/148). For the \textsc{Rain in Australia}, \textsc{MIMIC-IV}, \textsc{MNIST784}, and \textsc{HAR} models, only a subset of variables (9, 33, 560, and 548 respectively) was perturbed, as other variables were either strictly categorical or had limited unique values. While RDSA can technically perturb categorical variables, doing so risks introducing heavy biases into the distributions.

Figure \ref{fig:Event_Diff_FR_RDSA} displays the mean change of the input features of individual inputs ($\langle c_f \rangle$) for the TopoDNN model and different values of the perturbation dimension (the results for the other models can be seen in Figure \ref{fig:Appendix_Event_Diff_FR} in the appendix. Consistently, both $\langle c_f \rangle$ and the achieved Fooling Ratio increase with the number of shuffled variables in RDSA-generated adversaries across all models.

Figure \ref{fig:DistributionsCompare} illustrates the changes in one-dimensional distributions of selected input variables for the TopoDNN model (again, the results for all models can be found in the appendix, in Figure \ref{fig:DistributionsCompare_App}. As expected, minimal changes are observed between clean and adversarial examples generated by RDSA. In contrast, the impact of a LPF attack on the same distribution is shown for comparison, revealing drastic changes. Although both attacks achieve comparable Fooling Ratios, the LPF attacks more noticeably alters some of the one-dimensional distributions.

\begin{figure}[!htb]
     \centering
     \begin{subfigure}[b]{0.45\textwidth}
         \centering
         \includegraphics[width=\textwidth]{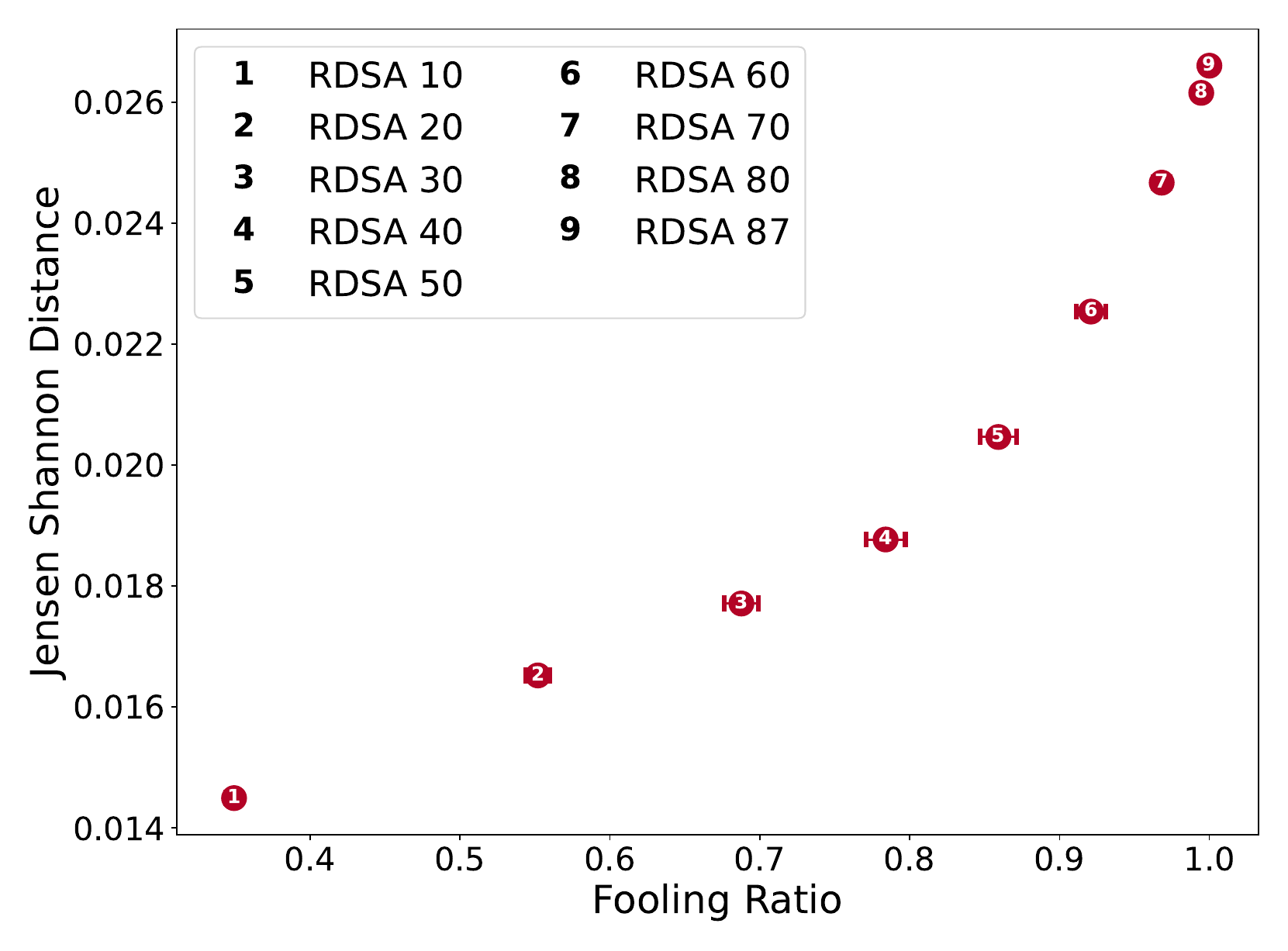}
         \caption{TopoDNN}
         \label{fig:JSD_RDSA_Topo}
     \end{subfigure}
     \hfill
     \begin{subfigure}[b]{0.45\textwidth}
         \centering
         \includegraphics[width=\textwidth]{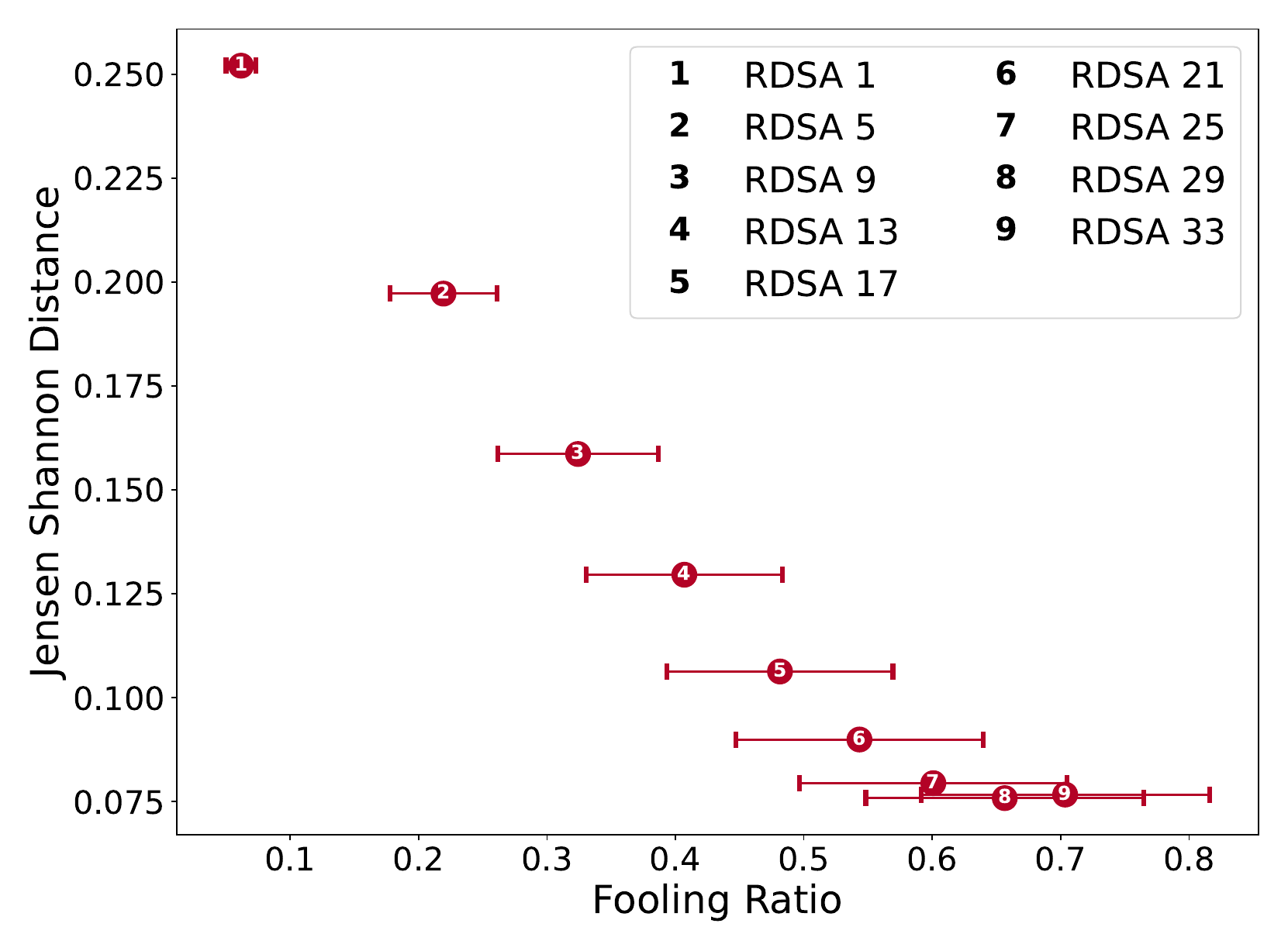}
         \caption{MIMIC-IV Mortality Model}
         \label{fig:JSD_RDSA_MIMICIV}
     \end{subfigure}
     \newline
        \caption{Average Jensen-Shannon Distances between the initial distributions and the adversarial distributions for different attacks applied.}
        \label{fig:JSD_RDSA}
\end{figure}

Figure \ref{fig:JSD_RDSA} illustrates the compatibility between clean and adversarial examples using the Jensen-Shannon Distance (JSD) averaged over runs for the TopoDNN and the MIMIC-IV Mortality Model. The results for the other models can be found in the appendix, in Figure \ref{fig:JSD_RDSA_App}. For the high-energy physics models, the average JSD is on the order of $10^{-2}$, indicating minor differences between initial and perturbed distributions. In the weather forecast example, JSD ranges from $10^{-1}$ to $10^{-2}$, increasing with the number of shuffled variables. For this model the increase in JSD occurs possibly due to pseudo-continuous features having only a limited amount of unique values, and class imbalance. Conversely, for the MIMIC-IV model, the JSD decreases with increasing the amount of variables shuffled, reaching a desirable order of $10^{-2}$ for high Fooling Ratio attacks, indicative of a good match between initial and adversarial distributions. The observed trend of decreasing JSD with an increasing number of shuffled features for the MIMIC-IV model, accompanied by a corresponding increase in the Fooling Ratio, may be attributed to the attack requiring - on average - less extreme shuffling of individual features to achieve misclassification. For this model, the effect appears to culminate in a 'sweet spot' when approximately half of the continuous features are shuffled.

Comparatively, the average JSD for LPF adversarial distributions is typically larger by between a factor of 2 and an order of magnitude, suggesting significantly greater discrepancies, as demonstrated in Figure \ref{fig:DistributionsCompare}. Detailed JSD results for LPF attacks are provided in the appendix.

\begin{figure}[htb!]
    \centering
    \begin{subfigure}[b]{0.4\textwidth}
         \centering
         \includegraphics[width=\textwidth]{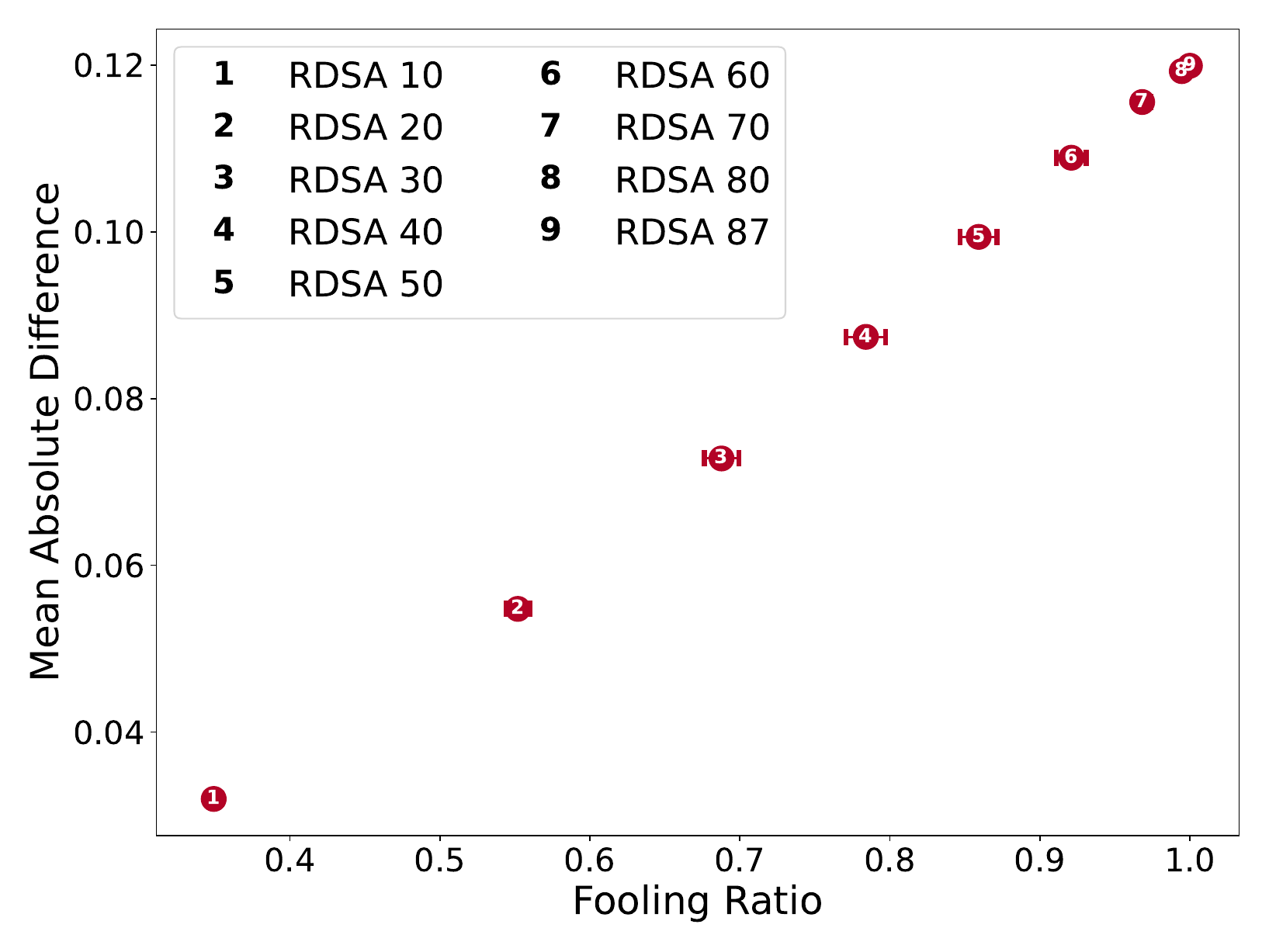}
         \caption{TopoDNN}
         \label{fig:Mean_Difference_Correlation_RDSA_Topo}
     \end{subfigure}
     \hfill
     \begin{subfigure}[b]{0.4\textwidth}
         \centering
         \includegraphics[width=\textwidth]{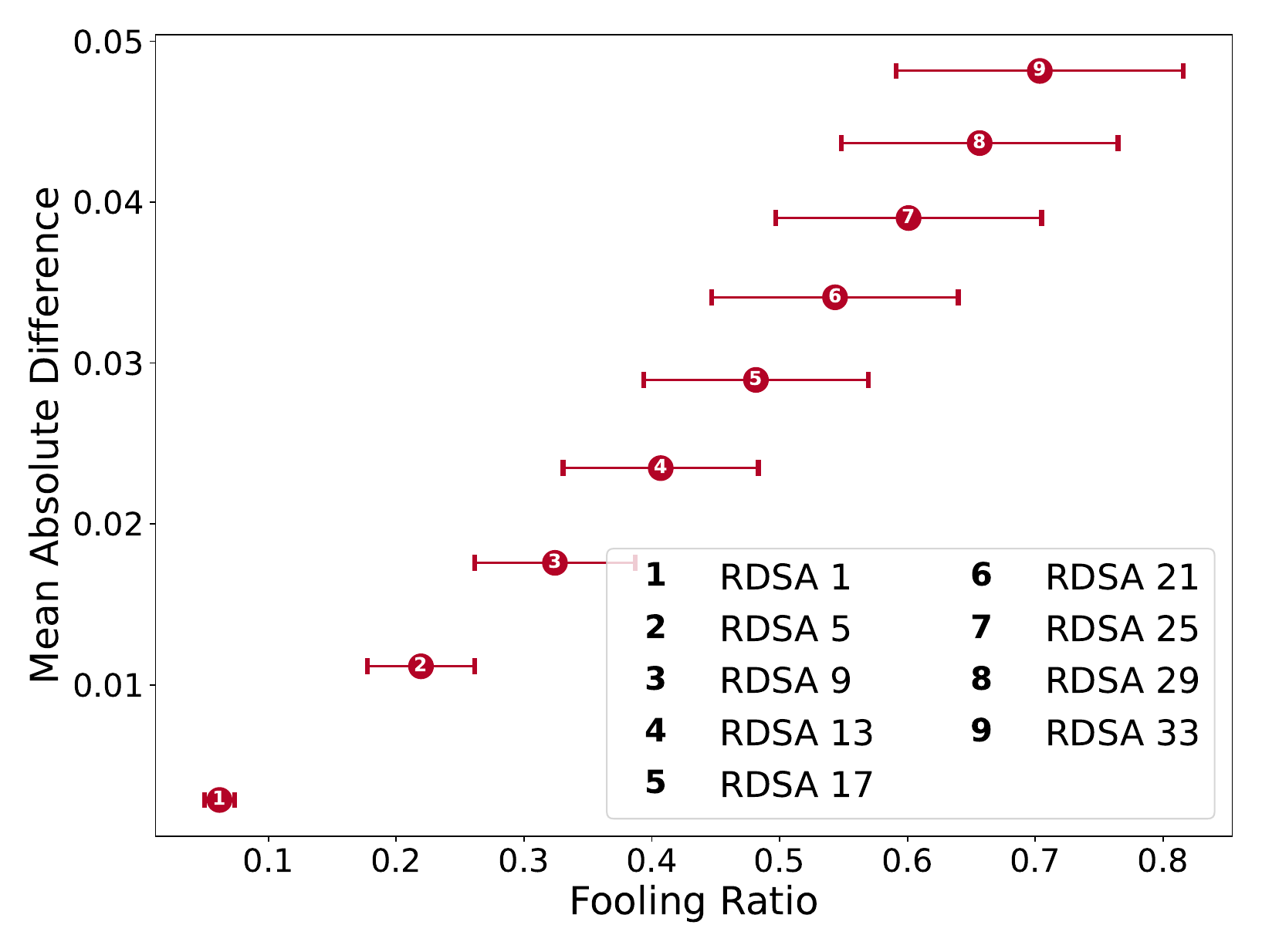}
         \caption{MIMIC-IV Mortality Model}
         \label{fig:Mean_Difference_Correlation_RDSA_MIMICIV}
     \end{subfigure}
     \newline
    \caption{Average absolute difference between the clean correlation matrices and the adversarial correlation matrices for different attacks applied on the TopoDNN (a) and the MIMIC-IV model (b).}
    \label{fig:Mean_Difference_Correlation_RDSA}
\end{figure}

As illustrated in the previous studies, RDSA yields minimal changes in one-dimensional input feature distributions, hence the significant fooling ratios are achieved by the altering the correlations between those features. To quantify this effect, we show the average change in correlation, $\langle s \rangle$, for the TopoDNN and the MIMIC-IV model in Figure \ref{fig:Mean_Difference_Correlation_RDSA}.

\begin{figure}[htb!]
     \centering
     \begin{subfigure}[b]{0.45\textwidth}
         \centering
         \includegraphics[width=\textwidth]{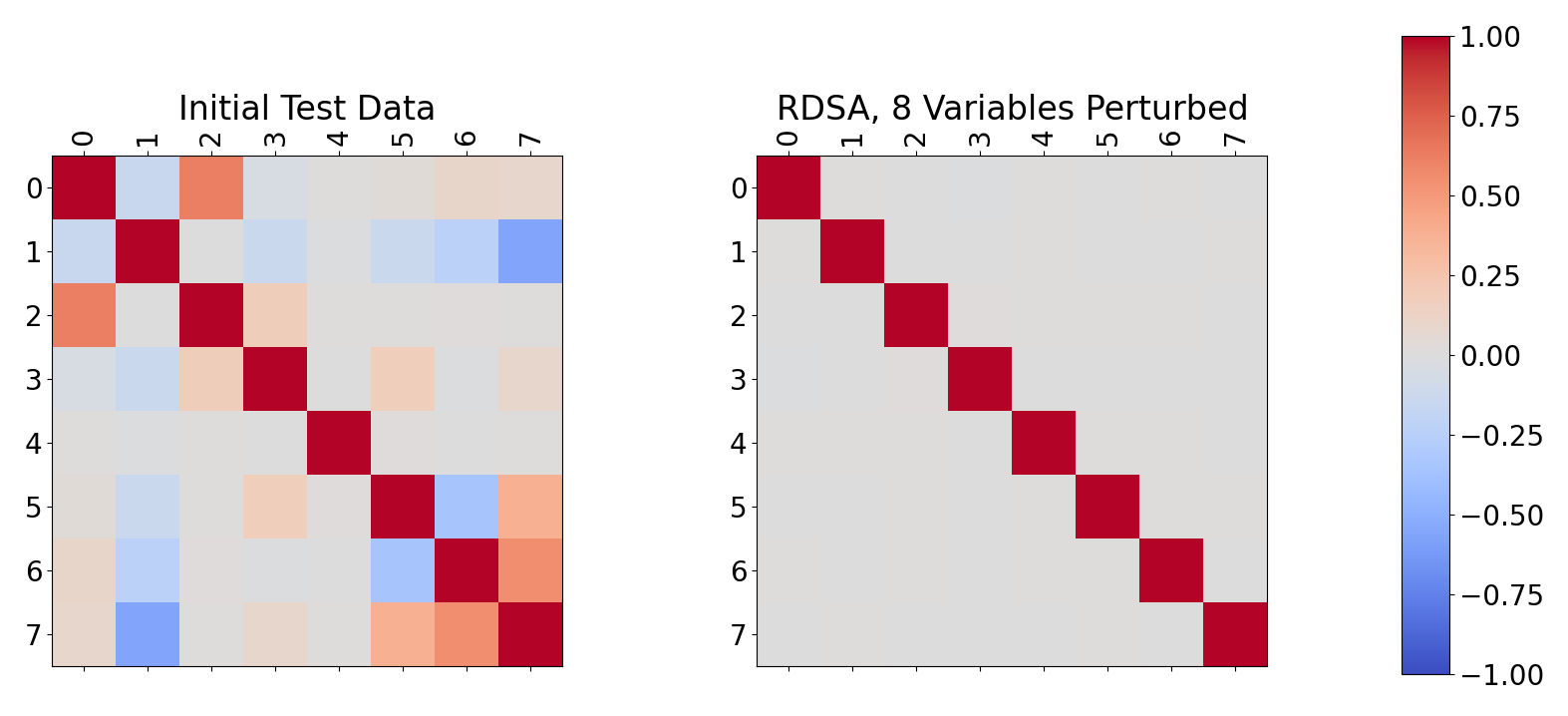}
         \caption{VBF Model}
         \label{fig:Correlation_Matrices_CA_8}
     \end{subfigure}
     \hfill
     \begin{subfigure}[b]{0.45\textwidth}
         \centering
         \includegraphics[width=\textwidth]{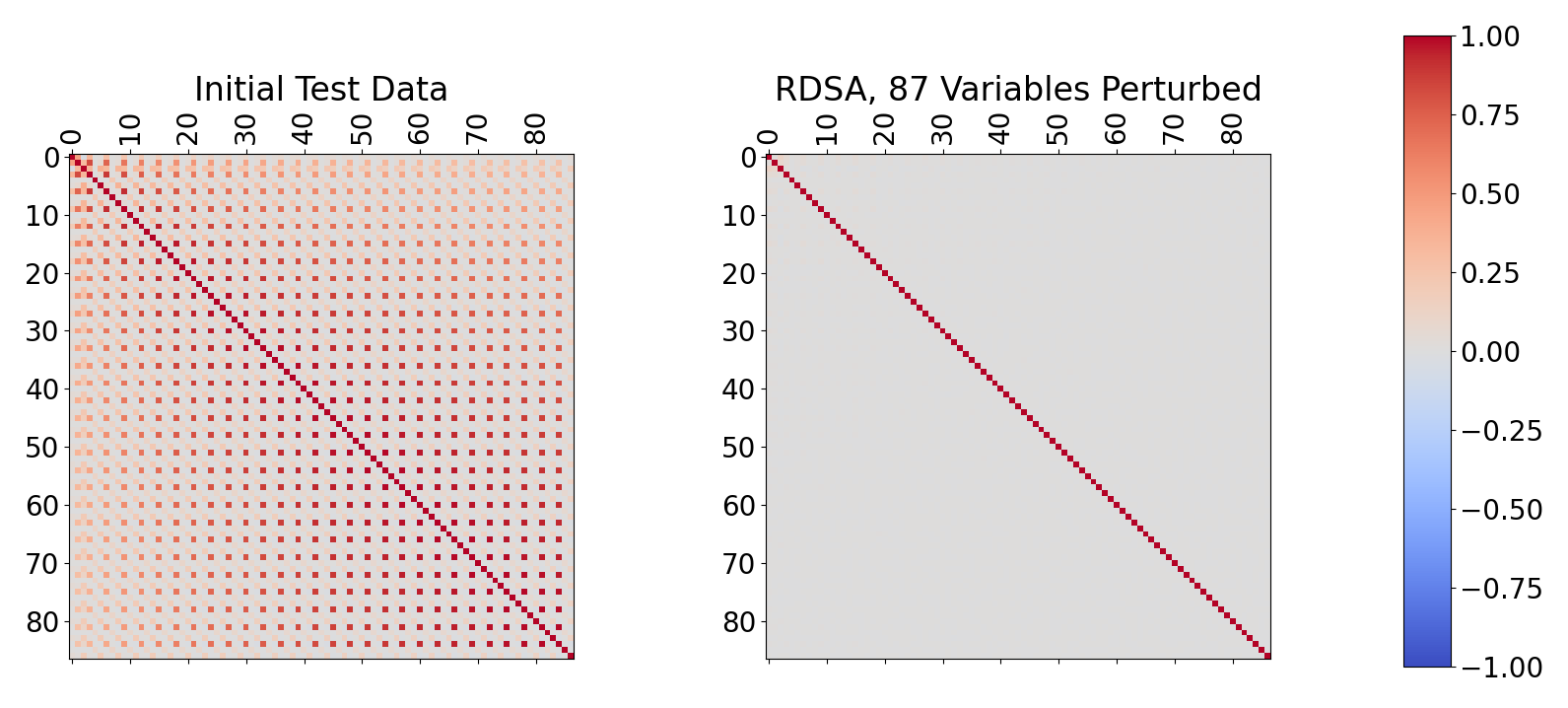}
         \caption{TopoDNN}
         \label{fig:Correlation_Matrices_CA_87}
     \end{subfigure}
     \hfill
     \centering   
     \begin{subfigure}[b]{0.45\textwidth}
         \centering
         \includegraphics[width=\textwidth]{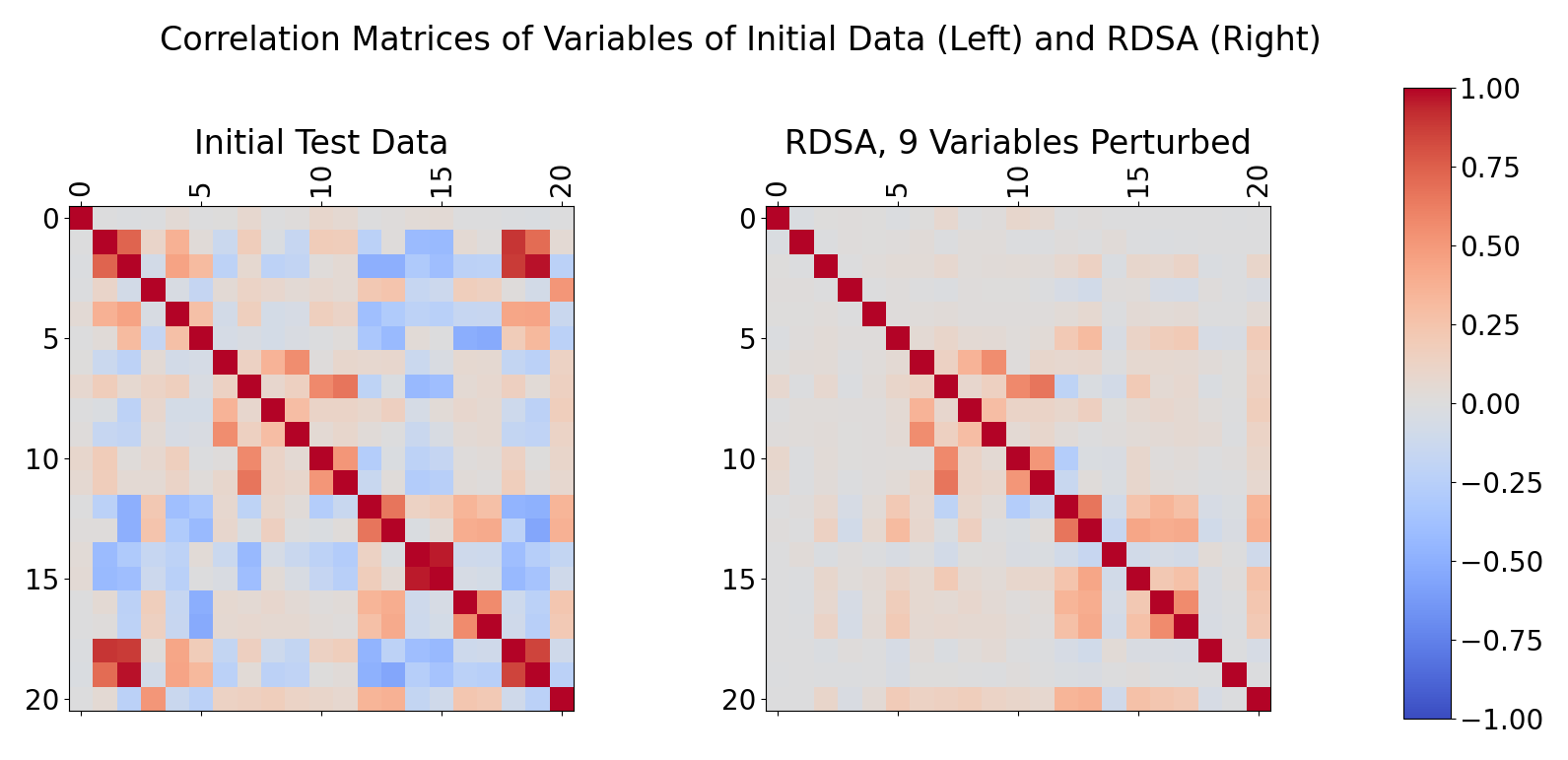}
         \caption{Rain in Australia Model}
         \label{fig:Correlation_Matrices_CA_9}
     \end{subfigure}
     \hfill
     \begin{subfigure}[b]{0.45\textwidth}
         \centering
         \includegraphics[width=\textwidth]{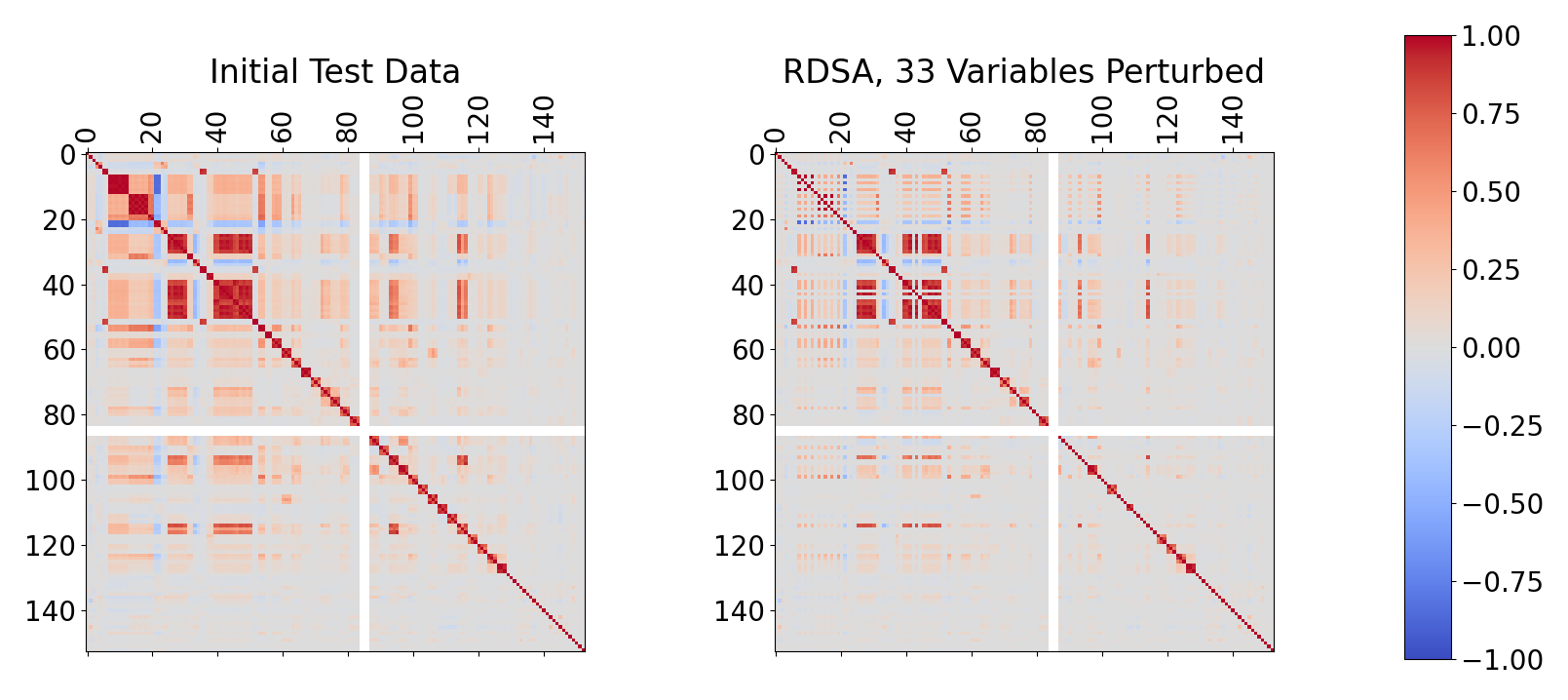}
         \caption{MIMIC-IV Mortality Model}
         \label{fig:Correlation_Matrices_CA_33}
     \end{subfigure}
    \centering
     \begin{subfigure}[b]{0.45\textwidth}
         \centering
         \includegraphics[width=\textwidth]{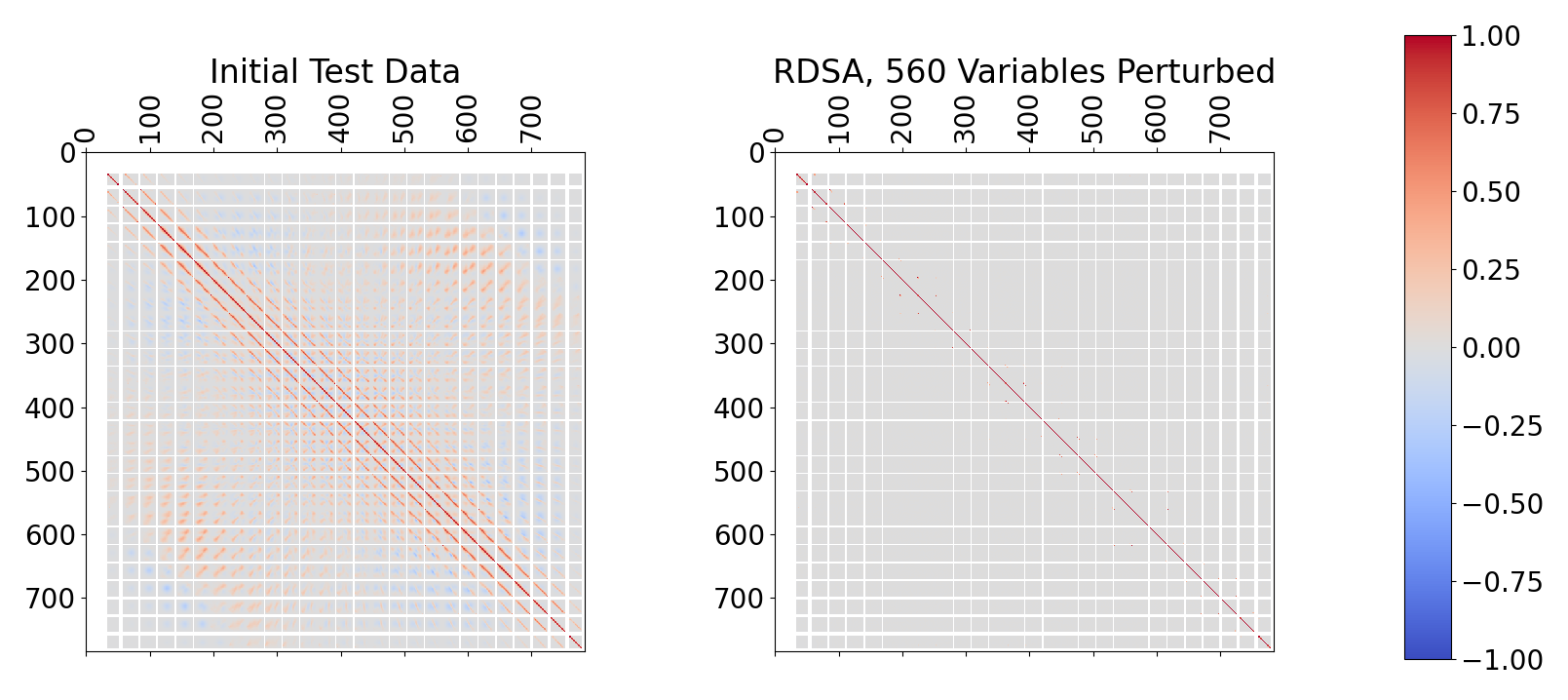}
         \caption{MNIST784 Model}
         \label{fig:Correlation_Matrices_CA_560}
     \end{subfigure}
     \hfill
     \begin{subfigure}[b]{0.49\textwidth}
         \centering
         \includegraphics[width=\textwidth]{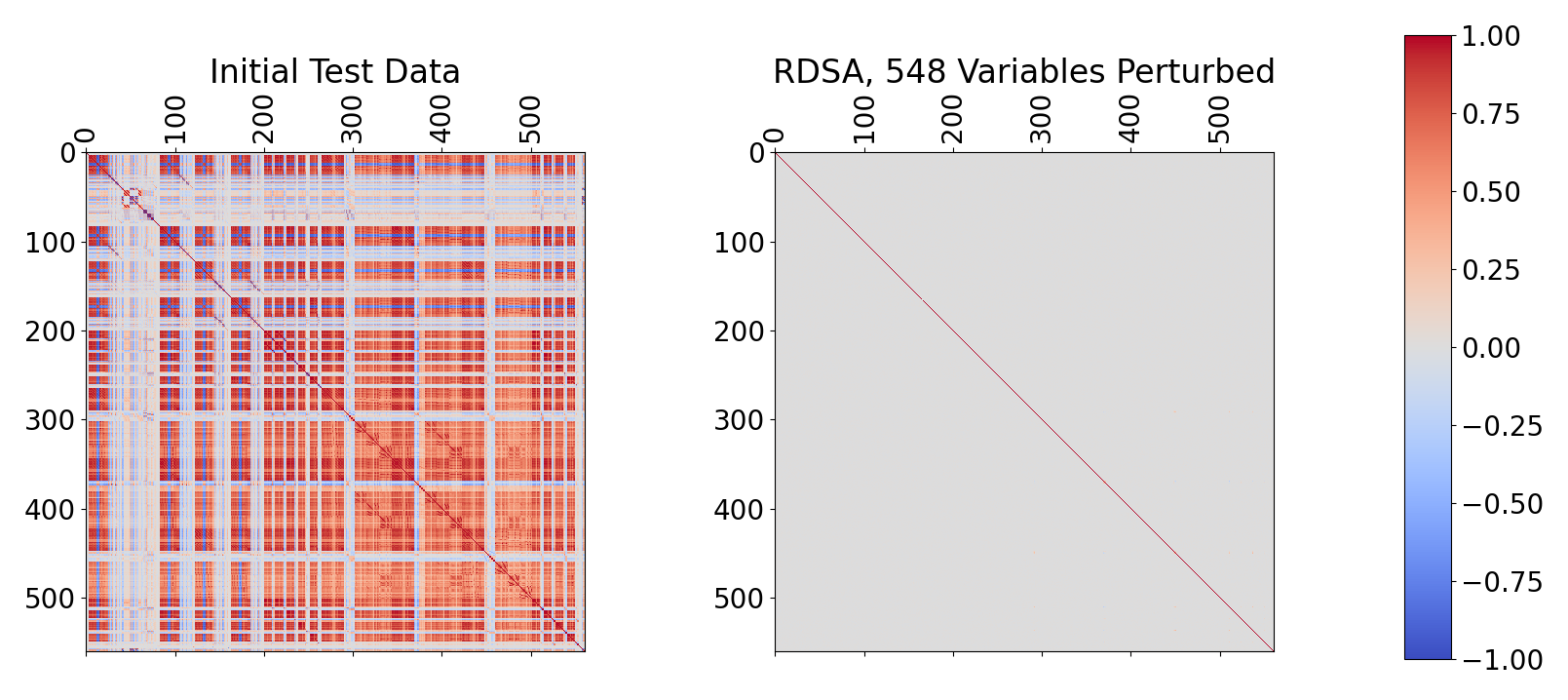}
         \caption{HAR Model}
         \label{fig:Correlation_Matrices_CA_548}
     \end{subfigure}
     \newline
        \caption{Correlation matrices for the clean test data (left) and the adversarial examples (right), generated using the Random Distribution Shuffle Attack on all 8 (VBF) / 87 (TopoDNN) variables (a, b), and on all continuous variables for Rain in Australia, MIMIC-IV, MNIST784 and HAR (c, d, e, f).}
        \label{fig:Correlation_Matrices}
\end{figure}

As expected, RDSA results in noticeable changes in the correlation matrices between clean and perturbed data. For all models, the correlation difference increases with the number of shuffled variables. When RDSA shuffles every variable, correlations vanish completely, reducing the correlation among input variables. This is illustrated in Figure \ref{fig:Correlation_Matrices} (a, b) for the \textsc{VBF} and \textsc{TopoDNN} models.

For the \textsc{Weather Forecast}, \textsc{MNIST784}, \textsc{HAR}, and \textsc{MIMIC-IV Mortality} models, only continuous variables are shuffled, affecting the resulting correlation matrices accordingly. The resulting correlation matrices are shown in Figure \ref{fig:Correlation_Matrices} (c, d, e, and f).

\section{Adversarial Training using the Random Distribution Shuffle Attack}

To test the classifier's robustness, training data-sets sizes were artificially reduced, then augmented using various techniques, doubling their reduced size. Classifiers were retrained with these augmented data-sets, and performance uncertainty was estimated by repeating this process 100 times, leveraging their RMS values as uncertainty.

Besides using RDSA for augmentation, we additionally tested CTGAN, TVAE \cite{xu2019modeling}, LPF attacks, and combinations of CTGAN, RDSA, and LPF. LPF was applied with varying $\alpha$ values, with a fixed amount of 100 steps.

\begin{figure}[htb!]
    \centering
    \includegraphics[width=\textwidth]{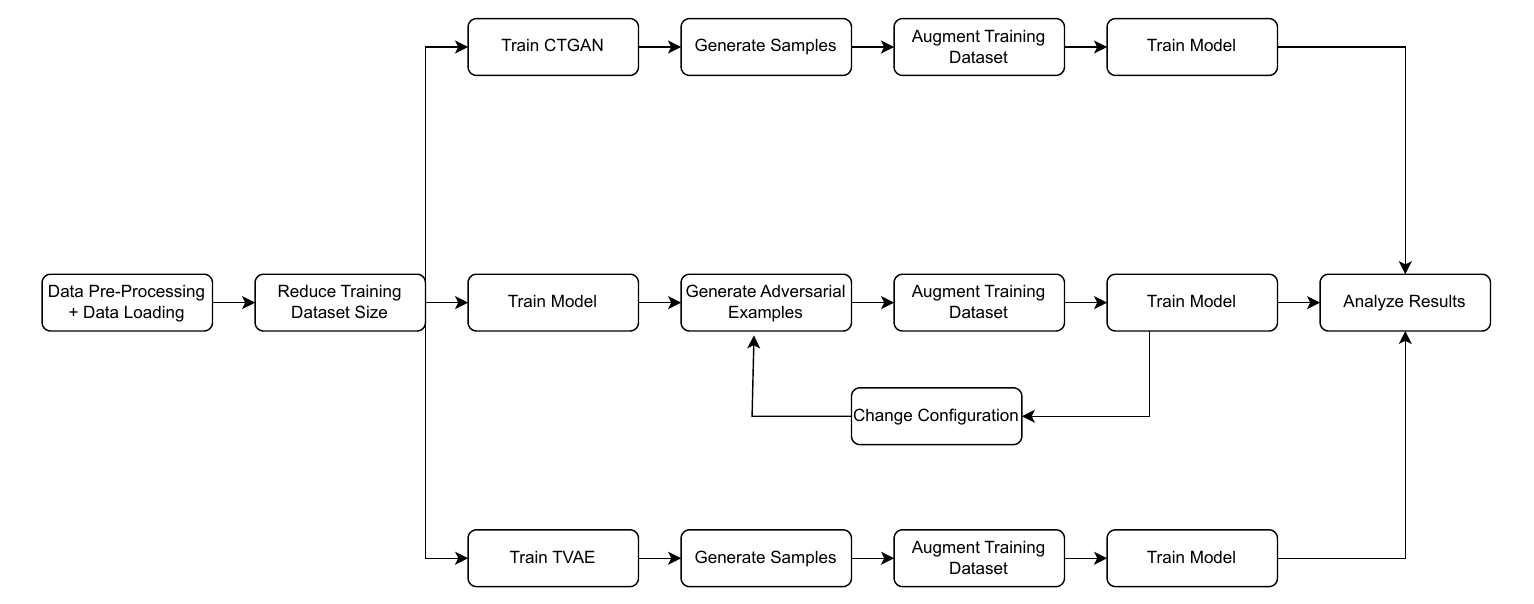}
    \caption{Flow chart visualizing the general data augmentation pipeline.}
    \label{fig:AugmentationFlowChart}
\end{figure}

The pipeline for data augmentation (visualized in Figure \ref{fig:AugmentationFlowChart}) is more involved than that for the attacks. First, the fully pre-processed datasets for training, testing, and validation are loaded again. These datasets are then artificially reduced in size until the model reaches data-starved regions. This reduction is achieved by randomly sub-sampling from the original training dataset, using a fixed random seed of 42 to ensure consistent results across 50 iterations.

Once the reduced training set is obtained, the respective models — as well as a CTGAN and a TVAE — are trained using this data, for 20 (CTGAN) and 40 (TVAE) epochs respectively, otherwise using the default parameters. Next, adversarial examples using RDSA and LPF, each with various configurations, are generated based on the reduced training data using the data-starved model. Additionally, the trained CTGAN and TVAE synthesizers are used to generate new training samples equal in number to the samples in the reduced training set.

The generated adversarial examples and synthesized samples are then combined with the reduced clean training data to form augmented training datasets. These augmented datasets are used to re-train the initial model, with its weights and biases reset. Finally, the performance of these augmented models is tested using the full initial test dataset, that has not been altered in any form.

\begin{figure}[htb!]
     \centering
     \begin{subfigure}[b]{0.4\textwidth}
         \centering
         \includegraphics[width=\textwidth]{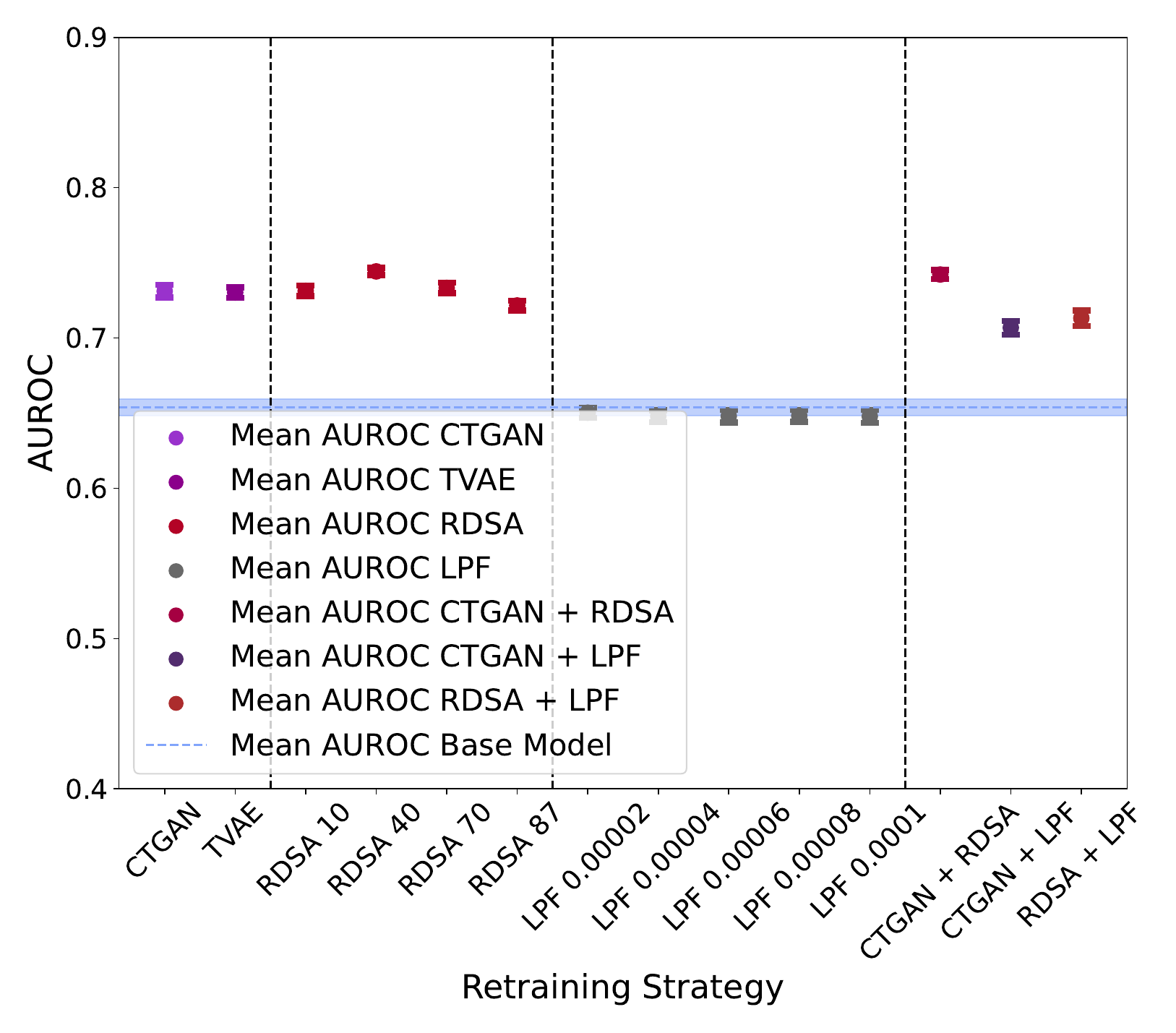}
         \caption{TopoDNN}
         \label{fig:AUROC_Retr_Topo}
     \end{subfigure}
     \hfill
     \begin{subfigure}[b]{0.4\textwidth}
         \centering
         \includegraphics[width=\textwidth]{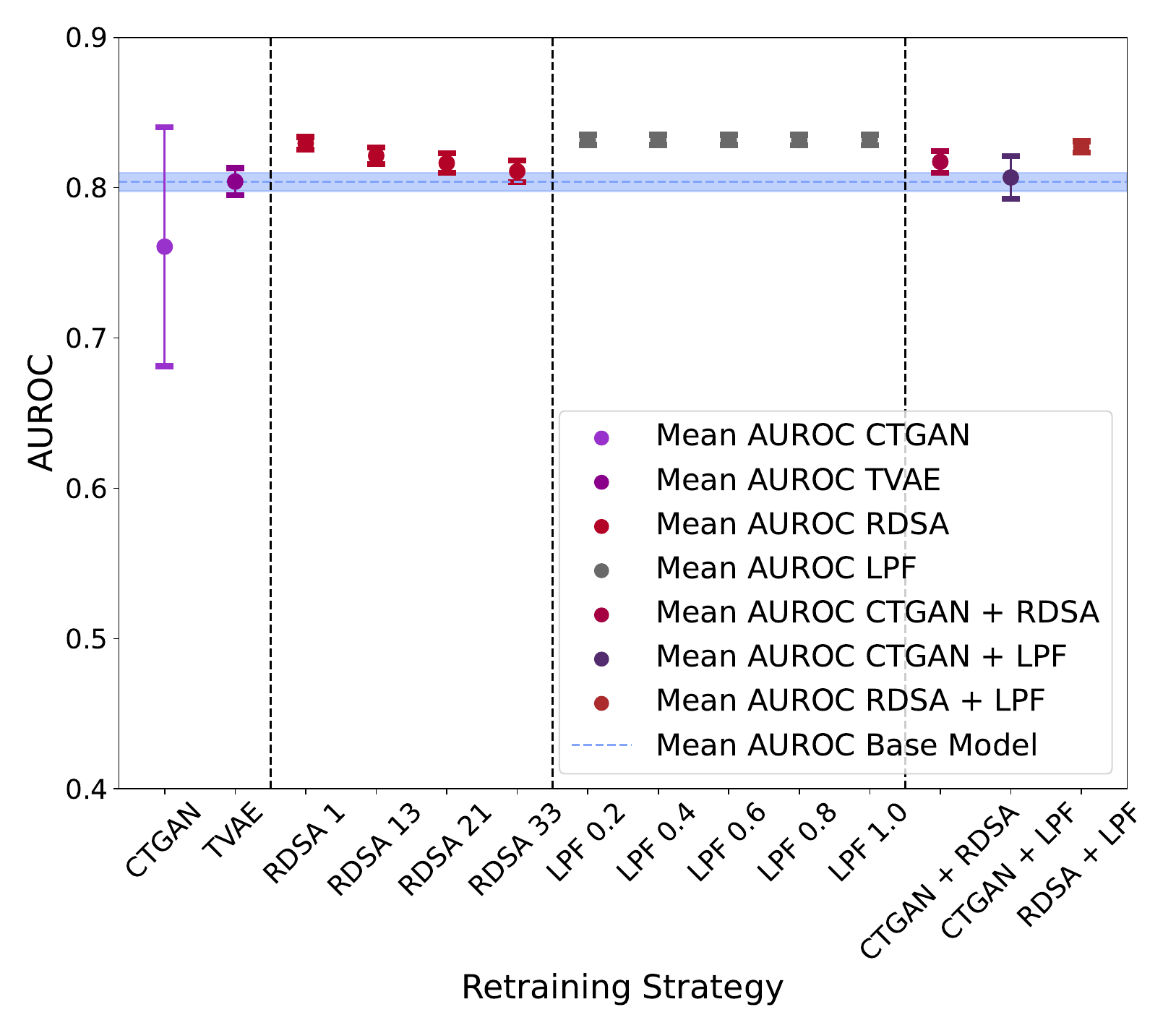}
         \caption{MIMIC-IV Mortality Model}
         \label{fig:AUROC_Retr_MIMICIV}
     \end{subfigure}
     \hfill
     \begin{subfigure}[b]{0.4\textwidth}
         \centering
         \includegraphics[width=\textwidth]{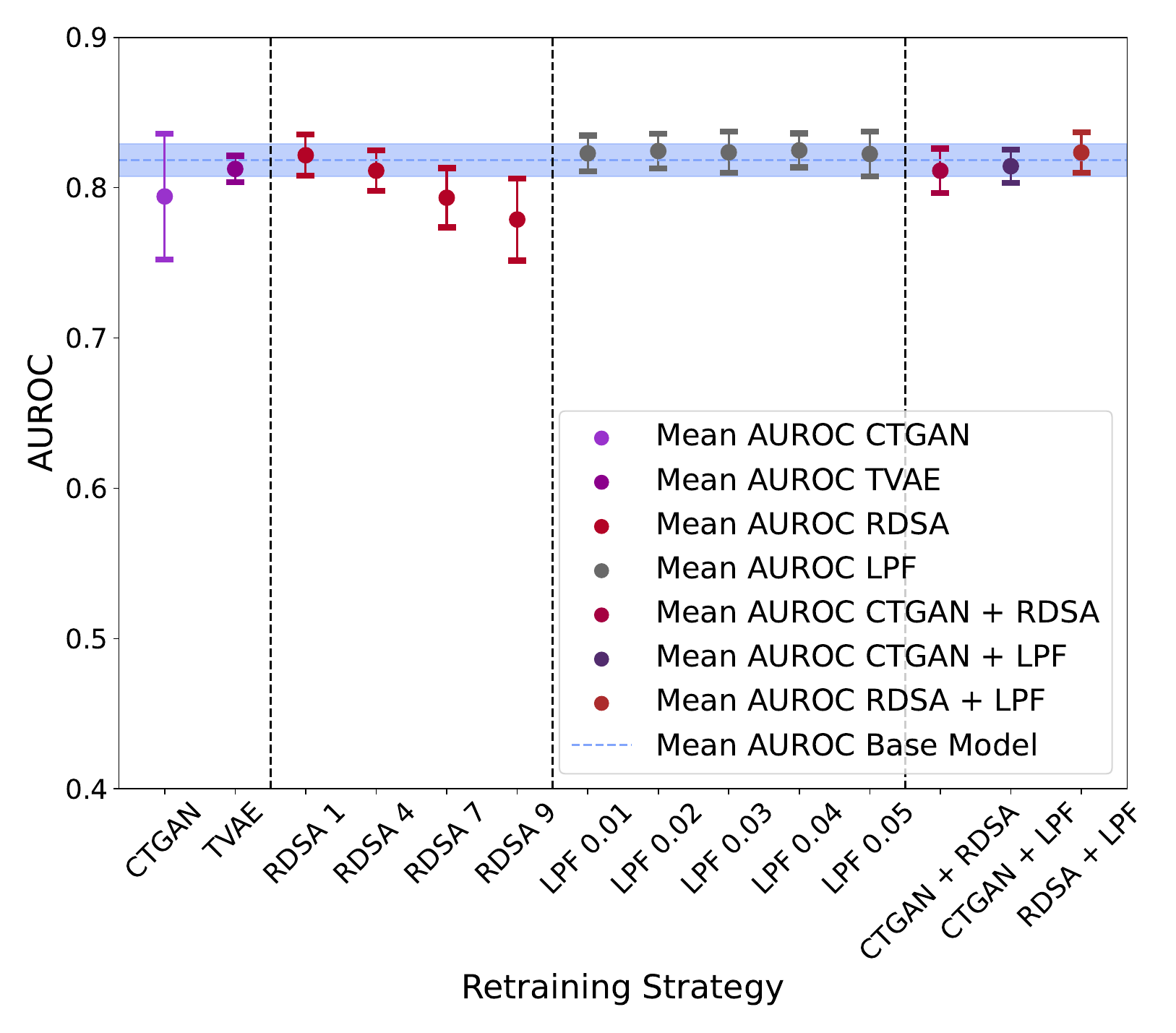}
         \caption{Rain in Australia Model}
         \label{fig:AUROC_Retr_Rain}
     \end{subfigure}
     \hfill
     \begin{subfigure}[b]{0.4\textwidth}
         \centering
         \includegraphics[width=\textwidth]{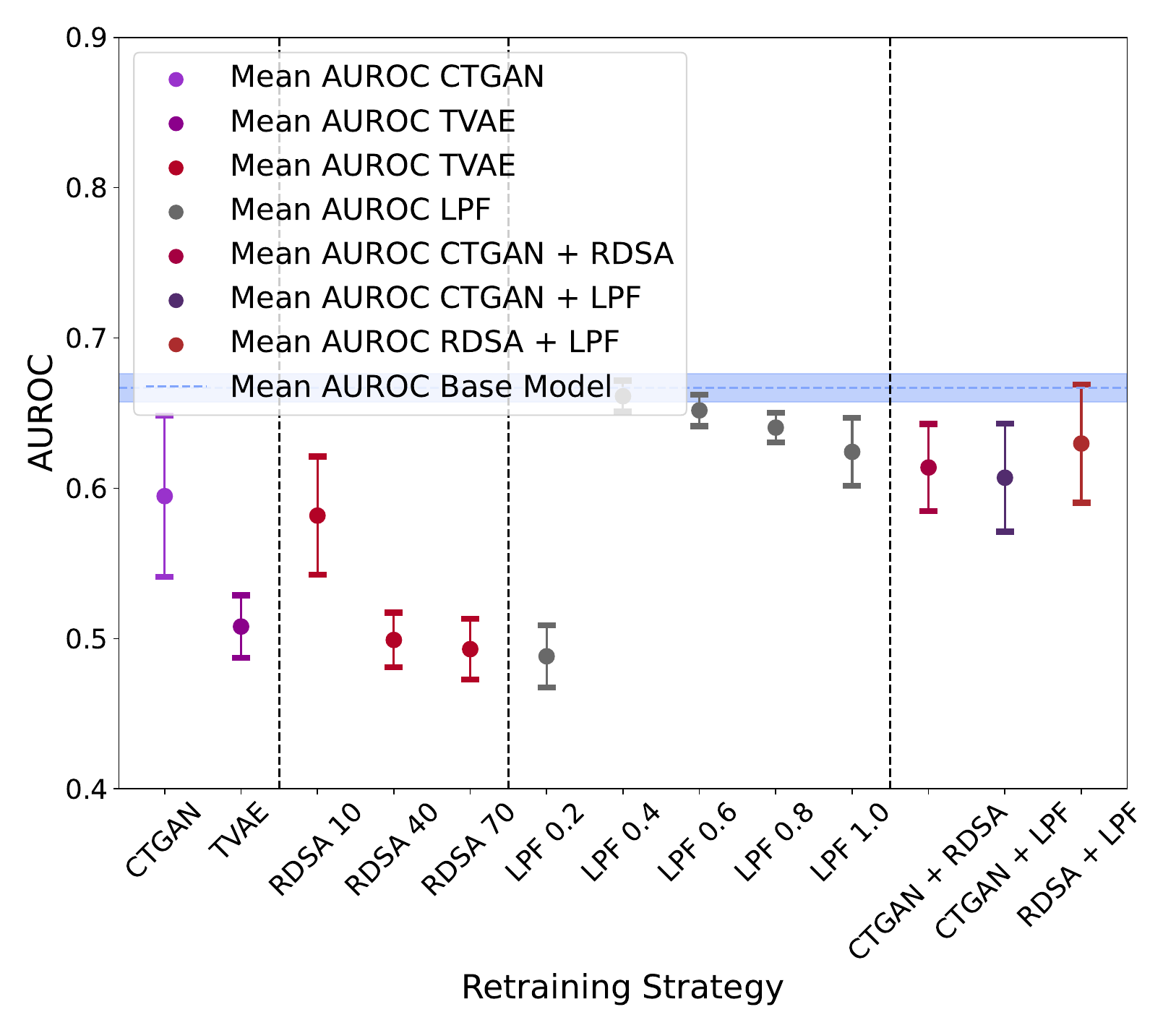}
         \caption{VBF Model}
         \label{fig:AUROC_Retr_VBF}
     \end{subfigure}
     \hfill
     \begin{subfigure}[b]{0.4\textwidth}
         \centering
         \includegraphics[width=\textwidth]{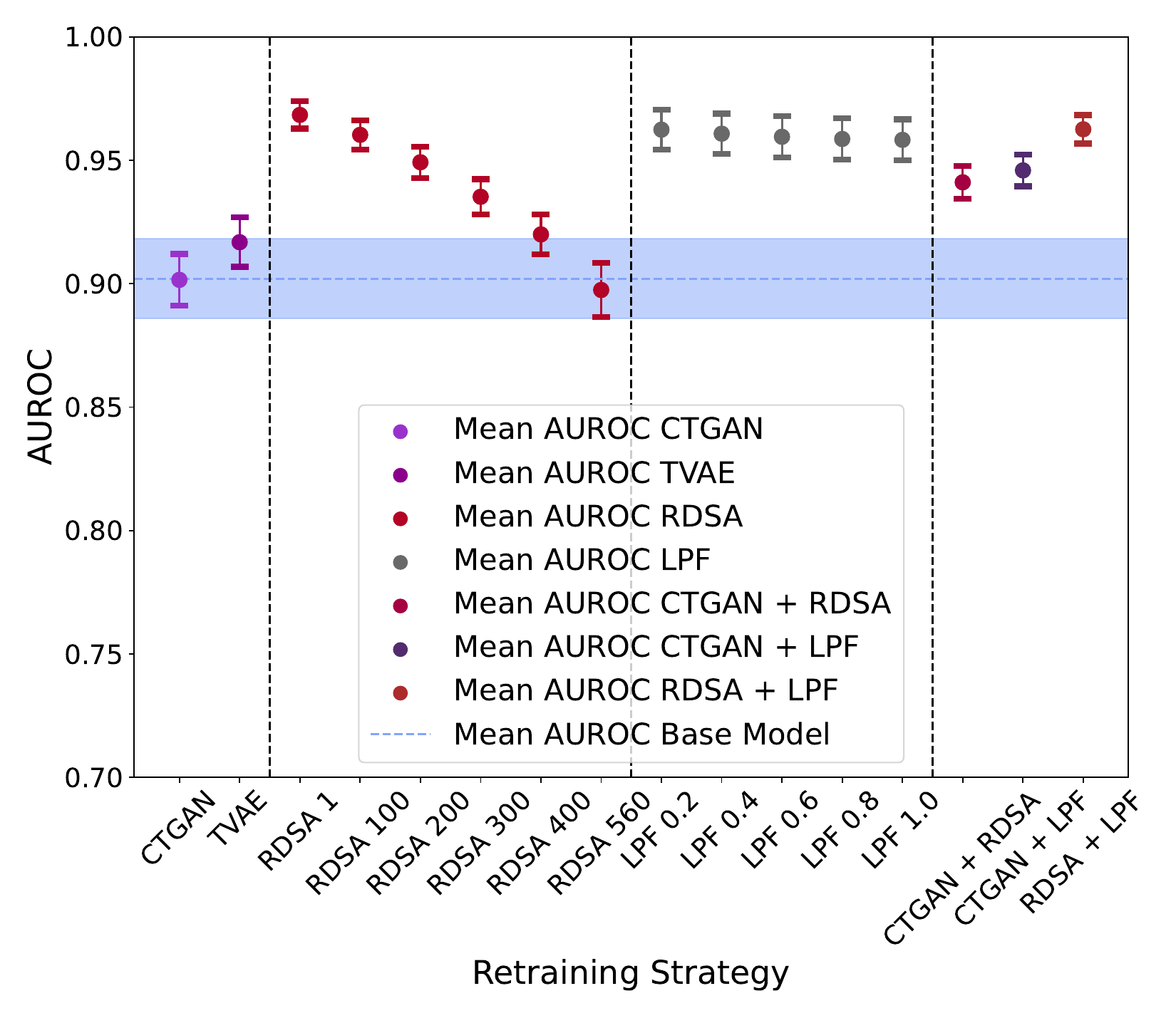}
         \caption{MNIST784 Model}
         \label{fig:AUROC_Retr_MNIST784}
     \end{subfigure}
     \hfill
     \begin{subfigure}[b]{0.4\textwidth}
         \centering
         \includegraphics[width=\textwidth]{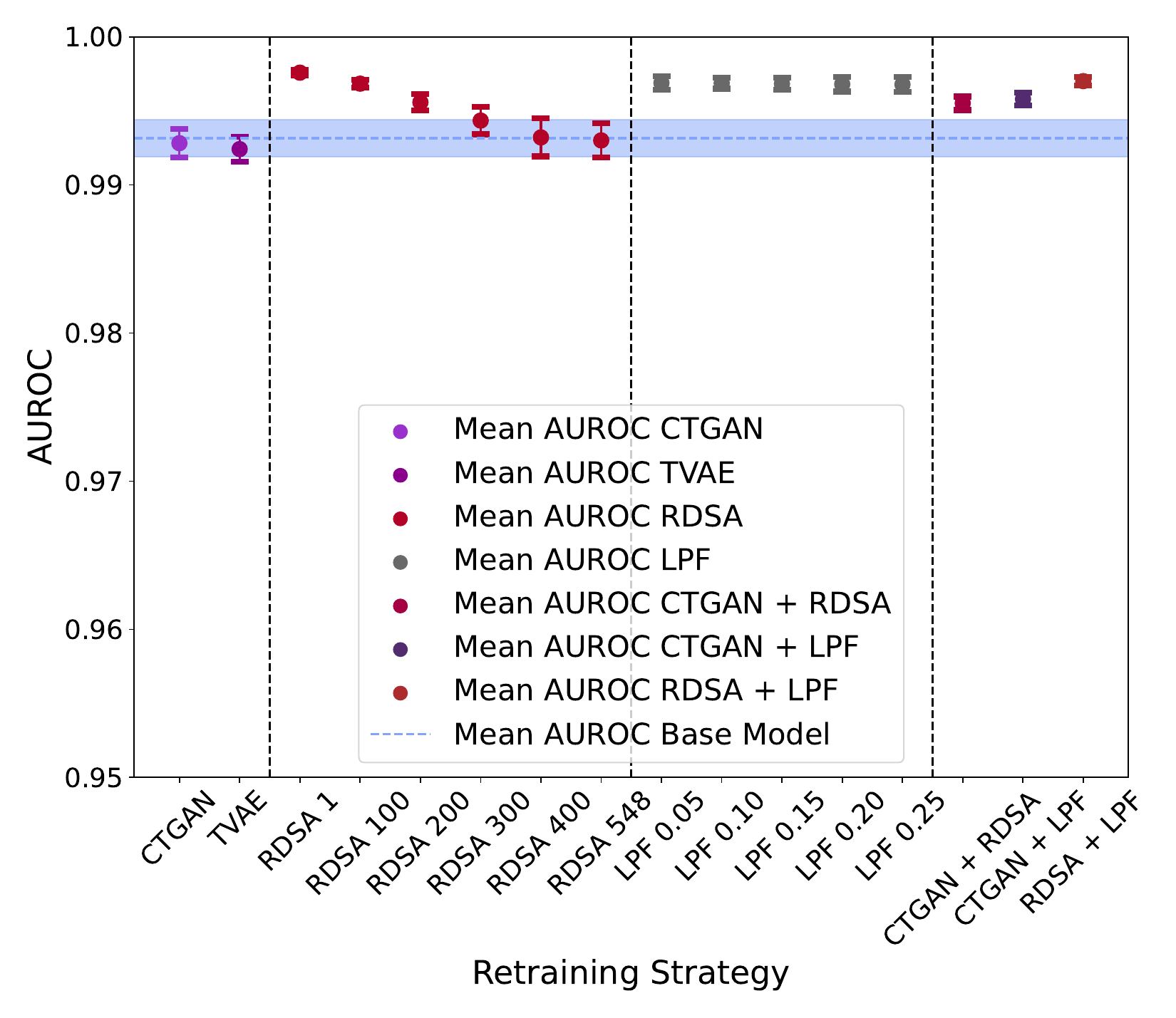}
         \caption{HAR Model}
         \label{fig:AUROC_Retr_HAR}
    \end{subfigure}
        \caption{Mean AUROC of varying data augmentation strategies, where the error bars are given by standard deviations of the AUROC values encountered during the runs. The blue line represents the average base model performances without any data augmentation, and the colored region its standard deviation.}
        \label{fig:AUROC_Retr}
\end{figure}

Given that some example models under study have a non-equal target class distribution for the training samples, the effectiveness of data augmentation was assessed by comparing the AUROC of each augmented model to the initial data-starved model. The mean AUROC performance is shown in Figure \ref{fig:AUROC_Retr}.

\textsc{TopoDNN} model results show that all augmentation techniques - except for LPF - display significant improvements over the baseline. RDSA always outperforms LPF and is competitive with CTGAN. 

A weaker but still significant improvement is also observed for the MIMIC-IV Mortality Model, where the RDSA outperforms most of the alternative approaches, with the exception of LPF retraining. This supports our hypothesis that neural network robustness can be improved by focusing on input feature correlations. 

The results on MNIST784 and HAR show similar results, where specific configurations of RDSA improve the networks performance significantly over the data-starved baseline. For both of these data sets, RDSA and LPF achieve fairly similar and competitive results, where both approaches noticeably outperform the state-of-the-art augmentation methods of CTGAN and TVAE. When considering the variability of the results found on the AUROC using the standard deviations, it appears that RDSA in general (very noticeable for the MNIST784 model) produced more stable results. However, for the Rain in Australia Model as well as the VBF model, almost all augmentation methods negatively impact the performance (Figure \ref{fig:AUROC_Retr}). 

\section{Limitations}

While the results for both leveraging RDSA as an attack and as a form of data augmentation show great potential, there are several limitations to this approach. First, the approach relies heavily on the (one-dimensional) feature distributions in the given dataset, i.e. in the test, validation or training data. Since this is a highly statistical method, its efficiency is strongly influenced by the amount of data available. In low data regimes, this attack can become volatile.

Additionally, the attack success depends on the availability of a sufficiently large dataset of continuous input features. While our approach can theoretically be applied to categorical features, doing so can introduce significant biases, thereby altering the distributions of these categorical variables.

\section{Conclusions}

This paper introduces the Random Distribution Shuffle Attack (RDSA), a novel method for generating adversarial examples in deep neural networks by altering the correlations between input features while preserving their distributions.

RDSA was tested on six diverse classification tasks, spanning high energy physics, meteorology, human activity, image recognition, and medical domain, demonstrating its effectiveness in generating adversarial examples with high Fooling Ratios and minimal changes to feature distributions. Retraining classifiers with these adversarial samples showed the capability to significantly improve their performance and robustness, often outperforming standard data augmentation methods like CTGAN, TVAE, and LPF.

Future research can explore the potential of RDSA to model intrinsic uncertainties and uncover more realistic vulnerabilities in neural networks, particularly in high energy physics. Additionally, using RDSA for data augmentation may enhance the focus on higher-order statistical moments, improving the interpretability of neural network classification results.

While initially motivated by particle physics, RDSA is equally applicable for other domains, being beneficial for a wide range of challenges, where a preservation of realistic relations between input observables in simulated data is of high importance.

\section*{Acknowledgement}
This work has been conducted in the context of the AISafety Project, funded by the BMBF under the grant proposal 05D23UM1. 

\bibliographystyle{elsarticle-num}

\appendix

\section{Input Data}
\label{App:Data}

The fully pre-processed and complete datasets (for all except the MIMIC-IV model), as well as a code snippet describing the models architecture, its hyperparameters, as well as some additional information such as model checkpoints can be found here: Currently not in here, as it is not anonymous. We will put the links back in here once the anonymous phase is over.

\subsection{VBF Model}
\label{App:VBF}
This model was trained for a total of 200 epochs, using a batch size of 300. The optimizer used is Nadam using a learning rate of 0.001. The original training data contains 200000 samples, and the reduced training set for data augmentation 1000 samples.

\begin{table}[h!]
\centering
\begin{tabular}{ |p{3cm}|p{3cm}|p{5cm}|  }
 \hline
 \multicolumn{3}{|c|}{\textbf{VBF Model Architecture}} \\
 \hline
  \textbf{Layer} & \textbf{Nodes} & \textbf{Activation Function}\\
 \hline \hline
 Dense (Input) &    8   &   ReLU\\
 \cline{1-3}
 Dense  &    8   &   ReLU\\
 \cline{1-3}
 Dense  &   4   &   ReLU\\
 \cline{1-3}
 Dense  &    4   &   ReLU\\
 \cline{1-3}
 Dense (Output)  &    2   &   Softmax\\
 \cline{1-3}
 \hline
\end{tabular}
\caption{Brief description of the architecture of the VBF Model.}
\label{table:VBFArchitecture}
\end{table}

Used as input for this model are the following variables: $\Delta\eta_{JJ}$, $\Delta\phi_{JJ}$, $m_{JJ}$, $MET$, $MET_{\phi}$, $\Delta MET_{\phi_{J1}}$, $\Delta MET_{\phi_{J2}}$, $\Delta MET_{\phi_{J1J2}}$, and $HT_1$ to $HT_{16}$. A detailed description of these variables can be found in the original paper \cite{Ngairangbam_2020}.
\subsection{TopoDNN}
\label{App:Topo}
This model was trained for a total of 100 epochs, using a batch size of 200. The optimizer used is Adam using a learning rate of 0.00005. The original training data contains 276720 samples, and the reduced training set for data augmentation 60878 samples.

\begin{table}[h!]
\centering
\begin{tabular}{ |p{3cm}|p{3cm}|p{5cm}|  }
 \hline
 \multicolumn{3}{|c|}{\textbf{TopoDNN Architecture}} \\
 \hline
  \textbf{Layer} & \textbf{Nodes} & \textbf{Activation Function}\\
 \hline \hline
 Dense (Input) &    300   &   ReLU\\
 \cline{1-3}
 Batch Normalization &    -   &   -\\
 \cline{1-3}
 Dense  &    102   &   ReLU\\
 \cline{1-3}
 Batch Normalization &    -   &   -\\
 \cline{1-3}
 Dense  &   12   &   ReLU\\
 \cline{1-3}
 Batch Normalization &    -   &   -\\
 \cline{1-3}
 Dense  &    6   &   ReLU\\
 \cline{1-3}
 Batch Normalization &    -   &   -\\
 \cline{1-3}
 Dense (Output)  &    1   &   Sigmoid\\
 \cline{1-3}
 \hline
\end{tabular}
\caption{Brief description of the architecture of the TopoDNN Model.}
\label{table:TopoDNNArchitecture}
\end{table}

As input for this model, we take $p_T$, $\eta$, and $\phi$ of the first 30 jet constituents, sorted by their momentum. However, we leave out $\eta$ and $\phi$ of the first constituent, as well as $\eta$ of the second constituent, as these take always the same value due to the pre-processing applied. Again, a more detailed overview of the variables can be found in the original paper \cite{Kasieczka_2019}.

\subsection{Rain in Australia Model}
\label{App:Rain}
This model was trained for a total of 150 epochs, using a batch size of 200. The optimizer used is Adam using a learning rate of 0.0001. The original training data contains 93094 samples, and the reduced training set for data augmentation 1396 samples.

\begin{table}[h!]
\centering
\begin{tabular}{ |p{3cm}|p{3cm}|p{5cm}|  }
 \hline
 \multicolumn{3}{|c|}{\textbf{Rain in Australia Model Architecture}} \\
 \hline
  \textbf{Layer} & \textbf{Nodes} & \textbf{Activation Function}\\
 \hline \hline
 Dense (Input) &    21   &   ReLU\\
 \cline{1-3}
 Dense  &    21   &   ReLU\\
 \cline{1-3}
 Dense  &   12   &   ReLU\\
 \cline{1-3}
 Dense  &    12   &   ReLU\\
 \cline{1-3}
 Dense  &    4   &   ReLU\\
 \cline{1-3}
 Dense  &    4    &   ReLU\\
 \cline{1-3}
 Dense (Output)  &    1   &   Sigmoid\\
 \cline{1-3}
 \hline
\end{tabular}
\caption{Brief description of the architecture of the Rain in Australia model.}
\label{table:RainArchitecture}
\end{table}

For this model, the input variables are the following: The location where the weather data was measured (Location), the minimal temperature of the day (MinTemp), the maximum encountered temperature of the day (MaxTemp), the total rainfall of the day (Rainfall), the total evaporation of the day (Evaporation), the numbers of hours of sunshine during the day (Sunshine), the direction of the strongest wind gust of the day (WindGustDir), the speed of the strongest wind gust of the day (WindGustSpeed), the direction of the wind at 9am (WindDir9am), the direction of the wind at 3pm (WindDir3pm), the humidity at 9am (Humidity9am), the humidity at 3pm (Humidity3pm), the atmospheric pressure at 3pm (Pressure3pm), the fraction of the sky obscured by clouds at 9am (Cloud9am), the fraction of the sky obscured by clouds at 3pm (Cloud3pm), the temperature at 9am (Temp9am), the temperature at 3pm (Temp3pm), as well as a boolean flag indicating whether there was any rain during this day (RainToday).

\subsection{MIMIC-IV Mortality Model}
\label{App:MIMICIV}
This model was trained for a total of 100 epochs, using a batch size of 200. The optimizer used is Adam using a learning rate of 0.000003. The original training data contains 36791 samples, and the reduced training set for data augmentation 9198 samples.

\begin{table}[h!]
\centering
\begin{tabular}{ |p{3cm}|p{3cm}|p{5cm}|  }
 \hline
 \multicolumn{3}{|c|}{\textbf{MIMIC-IV Mortality Model Architecture}} \\
 \hline
  \textbf{Layer} & \textbf{Nodes} & \textbf{Activation Function}\\
 \hline \hline
 Dense (Input) &    153   &   ReLU\\
 \cline{1-3}
 Dense  &    153   &   ReLU\\
 \cline{1-3}
 Dense  &   64   &   ReLU\\
 \cline{1-3}
 Dense  &    64   &   ReLU\\
 \cline{1-3}
 Dense  &    32   &   ReLU\\
 \cline{1-3}
 Dense  &    32    &   ReLU\\
 \cline{1-3}
 Dense  &    16    &   ReLU\\
 \cline{1-3}
 Dense  &    16    &   ReLU\\
 \cline{1-3}
 Dense (Output)  &    1   &   Sigmoid\\
 \cline{1-3}
 \hline
\end{tabular}
\caption{Brief description of the architecture of the MIMIC-IV Mortality Model.}
\label{table:MIMICIVArchitecture}
\end{table}

The input variables for this model can be separated into four broad categories. The first category entails only three variables pertaining to general information about the subject, namely their ethnicity, age, and gender. 

The second category - called chart events - contains vital signs and values of the patient, such as the measured heart rate or blood pressure. This category is for each measurement divided into two parts. The signal, which indicates whether this vital sign was measured during the stay, as well as the measured value if it was indeed measured. The vital measurements used in this model are as follows: Heart Rate (220045), Heart Rate Alarm - Low (220047), Arterial Blood Pressure systolic (220050), Arterial Blood Pressure diastolic (220051), Arterial Blood Pressure mean (220052), Pulmonary Artery Pressure systolic (220059), Pulmonary Artery Pressure diastolic (220060), Pulmonary Artery Pressure mean (220061), Central Venous Pressure (220074), Non Invasive Blood Pressure mean (220181), Respiratory Rate (220210), Minute Volume Alarm - Low (220292), Minute Volume Alarm - High (220293), PEEP set (220339), Epinephrine (221289), Dopamine (221662), Midazolam (Versed) (221668), Fentanyl (221744), Phenylephrine (221749), Propofol (222168), Vasopressin (222315), Temperature Celsius (223762), O2 Flow (223834), Resp Alarm - High (224161), Daily Weight (224639), Tidal Volume (set) (224684), Tidal Volume (observed) (224685), Minute Volume (224687), Respiratory Rate (Set) (224688), Peak Insp. Pressure (224695), Mean Airway Pressure (224697), SpO2 Desat Limit (226253), Glucose (whole blood) (226537), and Height (226707).

The third category contains all of the medications applied to the patient. This category is further split into three values for each measurement, namely the signal - indicating whether this medication was used during treatment - the rate - indicating the rate at which this medication was applied - as well as the amount - indicating the amount of the medication given to the patient. The medications considered as input for this model are as follows: Albumin 5\% (220864), Fresh Frozen Plasma (220970), Lorazepam (Ativan) (221385), Furosemide (Lasix) (221794), Hydralazine (221828), Norepinephrine (221906), Nitroglycerin (222056), Insulin - Regular (223258), Morphine Sulfate (225154), Packed Red Blood Cells (225168), D5 1/2NS (225823), LR (225828), Solution (225943), Sterile Water (225944), Piggyback (226089), KCL (Bolus) (227522), and Magnesium Sulfate (Bolus) (227523). The numbers given in brackets after each element for the previous two categories is the ID given to these values within the MIMIC-IV database.

The final category contains the medical conditions/diagnoses of the patients. The following diagnoses were considered for the model input: C41, C79, C83, D50, D68, E03, E66, E87, F10, F19, F29, F32, G83, I27, I28, I49, I50, I73, I97, K27, K76, M08, N19, and R63. The given IDs here represent the ICD-10 codes of the respective conditions.

\subsection{MNIST784 Model}
\label{App:MNIST784}
This model was trained for a total of 100 epochs, using a batch size of 200. The optimizer used is Adam using a learning rate of 0.000003. The original training data contains 56000 samples, and the reduced training set for data augmentation 11200 samples.

\begin{table}[h!]
\centering
\begin{tabular}{ |p{3cm}|p{3cm}|p{5cm}|  }
 \hline
 \multicolumn{3}{|c|}{\textbf{MNIST784 Architecture}} \\
 \hline
  \textbf{Layer} & \textbf{Nodes} & \textbf{Activation Function}\\
 \hline \hline
 Dense (Input) &    128   &   ReLU\\
 \cline{1-3}
 Dense  &    64   &   ReLU\\
 \cline{1-3}
 Dense  &   32   &   ReLU\\
 \cline{1-3}
 Dense  &    16   &   ReLU\\
 \cline{1-3}
 Dense  &    8   &   ReLU\\
 \cline{1-3}
 Dense (Output)  &    10   &   Softmax\\
 \cline{1-3}
 \hline
\end{tabular}
\caption{Brief description of the architecture of the MNIST784 model.}
\label{table:RainArchitecture}
\end{table}

For this model, the input variables are the following are corresponding to a single pixel in the original 28x28 hand-written digit greyscale images.

\subsection{HAR Model}
\label{App:HAR}
This model was trained for a total of 100 epochs, using a batch size of 200. The optimizer used is Adam using a learning rate of 0.00003. The original training data contains 8239 samples, and the reduced training set for data augmentation 1647 samples.

\begin{table}[h!]
\centering
\begin{tabular}{ |p{3cm}|p{3cm}|p{5cm}|  }
 \hline
 \multicolumn{3}{|c|}{\textbf{HAR Model Architecture}} \\
 \hline
  \textbf{Layer} & \textbf{Nodes} & \textbf{Activation Function}\\
 \hline \hline
 Dense (Input) &    128   &   ReLU\\
 \cline{1-3}
 Dense  &    64   &   ReLU\\
 \cline{1-3}
 Dense  &   32   &   ReLU\\
 \cline{1-3}
 Dense  &    16   &   ReLU\\
 \cline{1-3}
 Dense (Output)  &    6   &   Softmax\\
 \cline{1-3}
 \hline
\end{tabular}
\caption{Brief description of the architecture of the HAR model.}
\label{table:RainArchitecture}
\end{table}

For this model, the input variables are the following correspond to 561 time and frequency domain variables derived from measurements conducted by the accelerometer and gyroscope present in smartphones that were carried on the participants' waists.

\section{Full Results (Plots)}

\subsection{Attack}

\begin{figure}[H]
     \centering
     \begin{subfigure}[b]{0.45\textwidth}
         \centering
         \includegraphics[width=\textwidth]{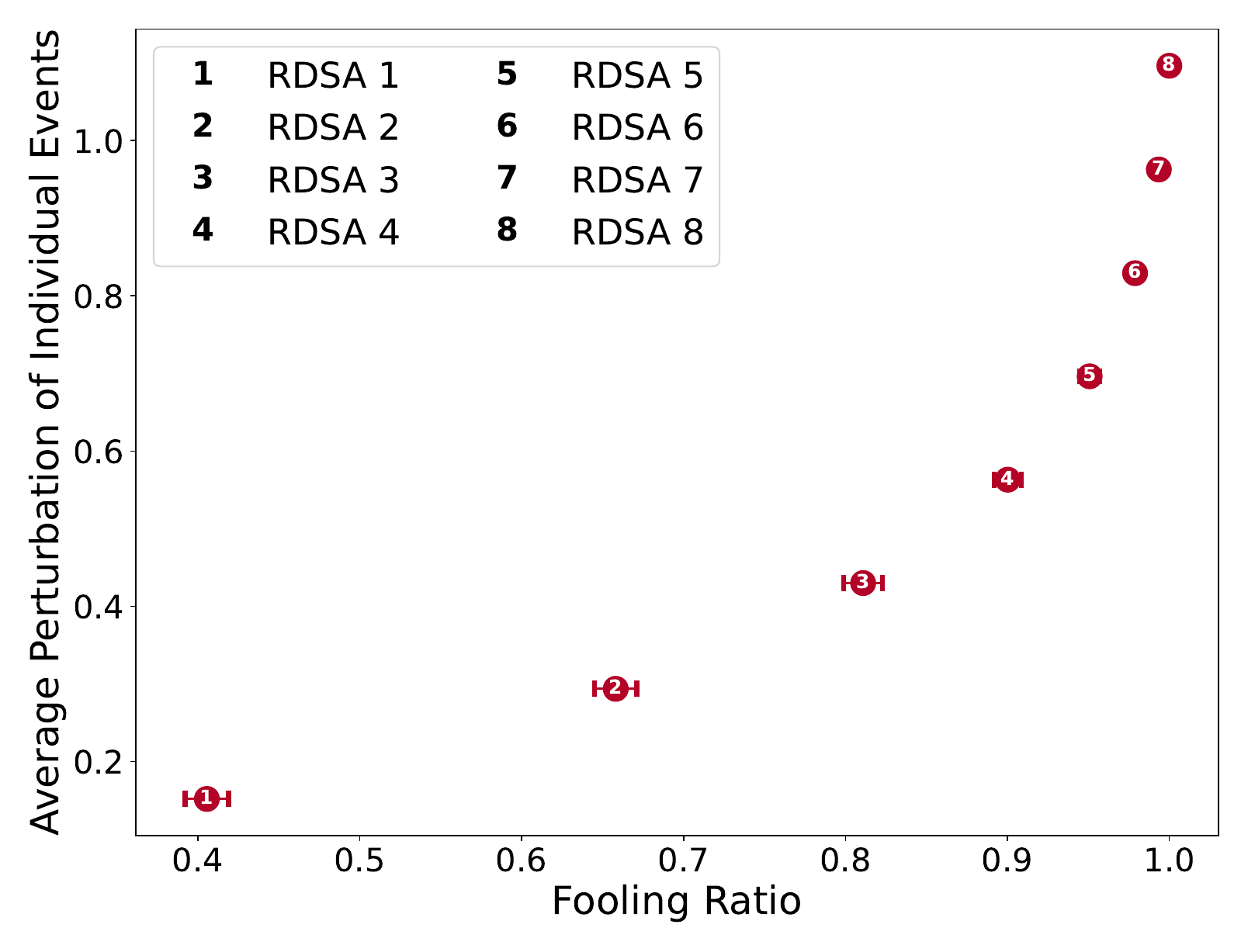}
         \caption{VBF Model}
         \label{fig:Event_Diff_FR_RDSA_VBF}
     \end{subfigure}
     \hfill
     \begin{subfigure}[b]{0.45\textwidth}
         \centering
         \includegraphics[width=\textwidth]{Event_Diff_FR_RDSA_Topo.pdf}
         \caption{TopoDNN}
         \label{fig:Event_Diff_FR_RDSA_Topo}
     \end{subfigure}
     \hfill
     \begin{subfigure}[b]{0.45\textwidth}
         \centering
         \includegraphics[width=\textwidth]{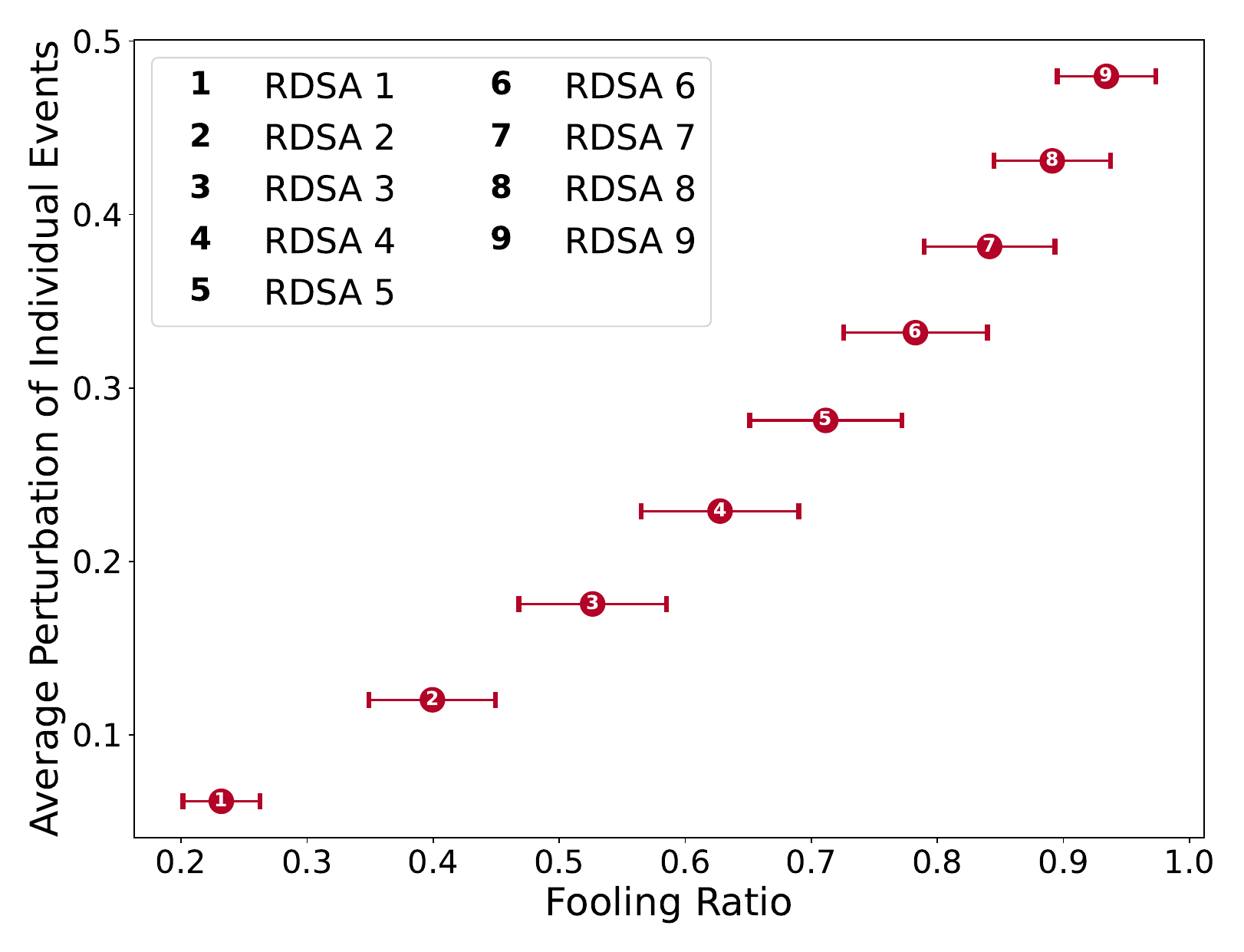}
         \caption{Rain in Australia Model}
         \label{fig:Event_Diff_FR_RDSA_Rain}
     \end{subfigure}
     \hfill
     \begin{subfigure}[b]{0.45\textwidth}
         \centering
         \includegraphics[width=\textwidth]{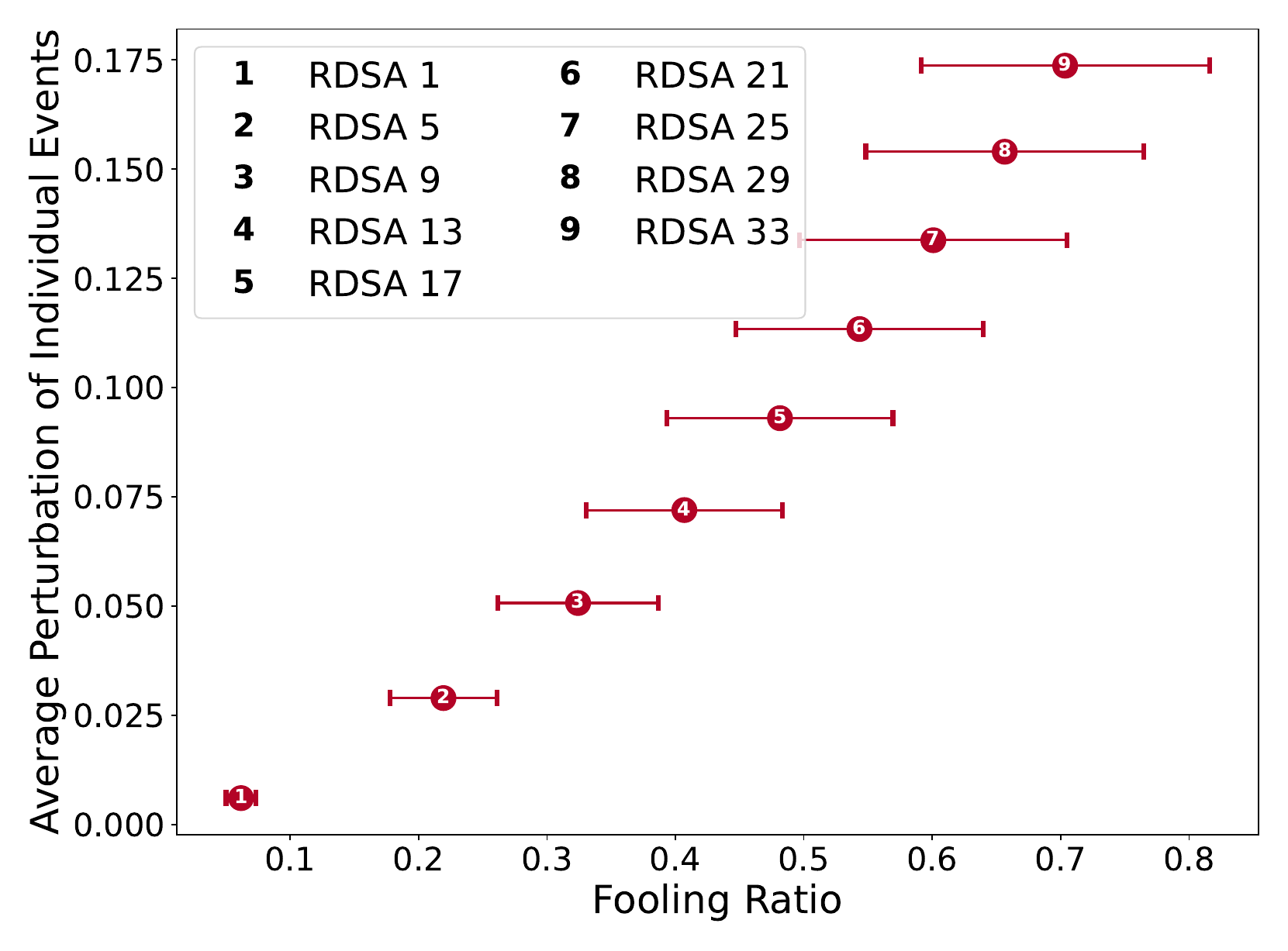}
         \caption{MIMIC-IV Mortality Model}
         \label{fig:Event_Diff_FR_RDSA_MIMICIV}
     \end{subfigure}
     \hfill
     
     \begin{subfigure}[b]{0.45\textwidth}
         \centering
         \includegraphics[width=\textwidth]{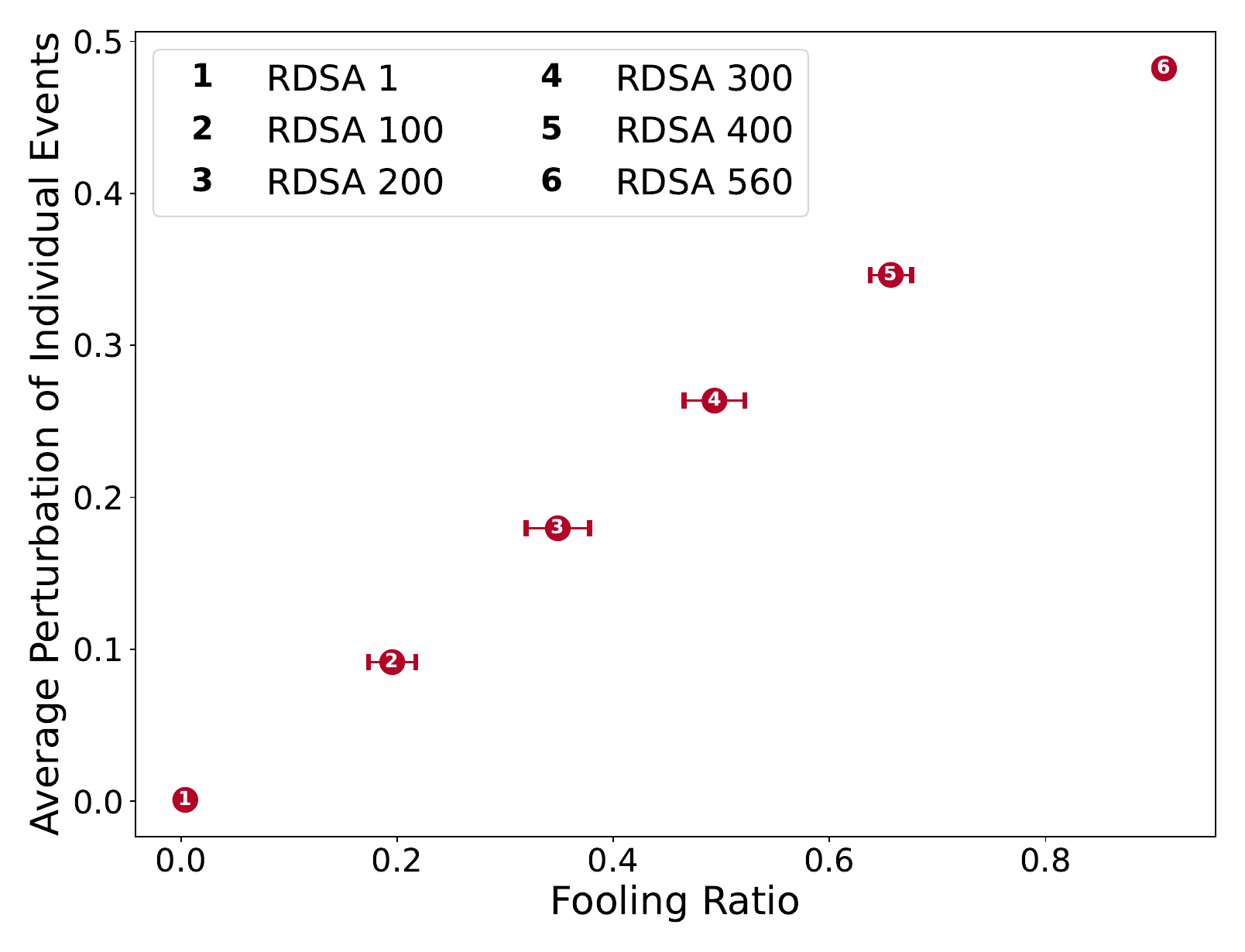}
         \caption{MNIST784 Model}
         \label{fig:Event_Diff_FR_RDSA_MNIST784}
     \end{subfigure}
     \hfill
     \begin{subfigure}[b]{0.45\textwidth}
         \centering
         \includegraphics[width=\textwidth]{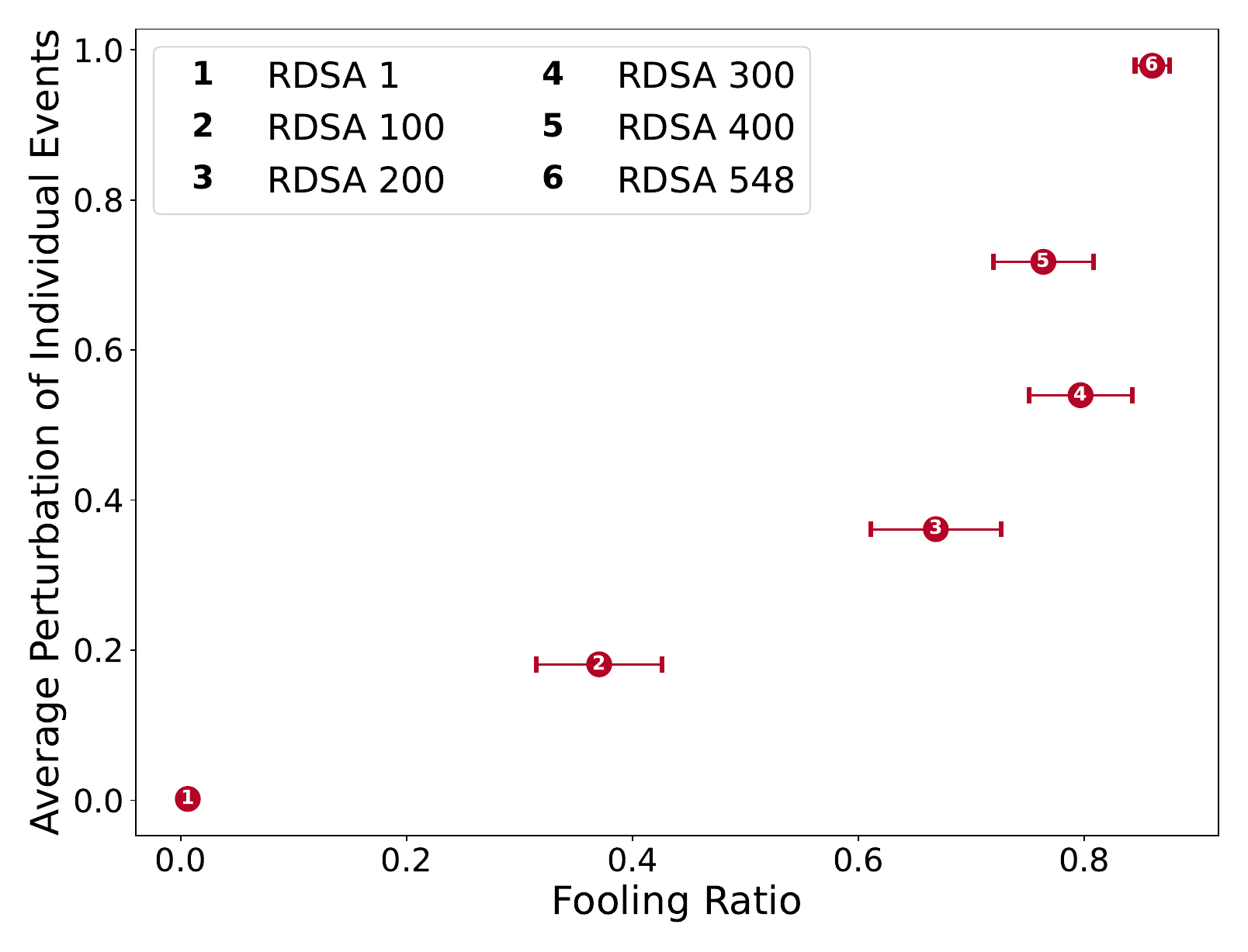}
         \caption{HAR Model}
         \label{fig:Event_Diff_FR_RDSA_HAR}
     \end{subfigure}
        \caption{Average difference / perturbation between individual clean inputs (test set) and corresponding adversaries.}
        \label{fig:Event_Diff_FR_RDSA_App}
    \label{fig:Appendix_Event_Diff_FR}
\end{figure}

\begin{figure}[H]
     \centering
     \begin{subfigure}[b]{0.45\textwidth}
         \centering
         \includegraphics[width=\textwidth]{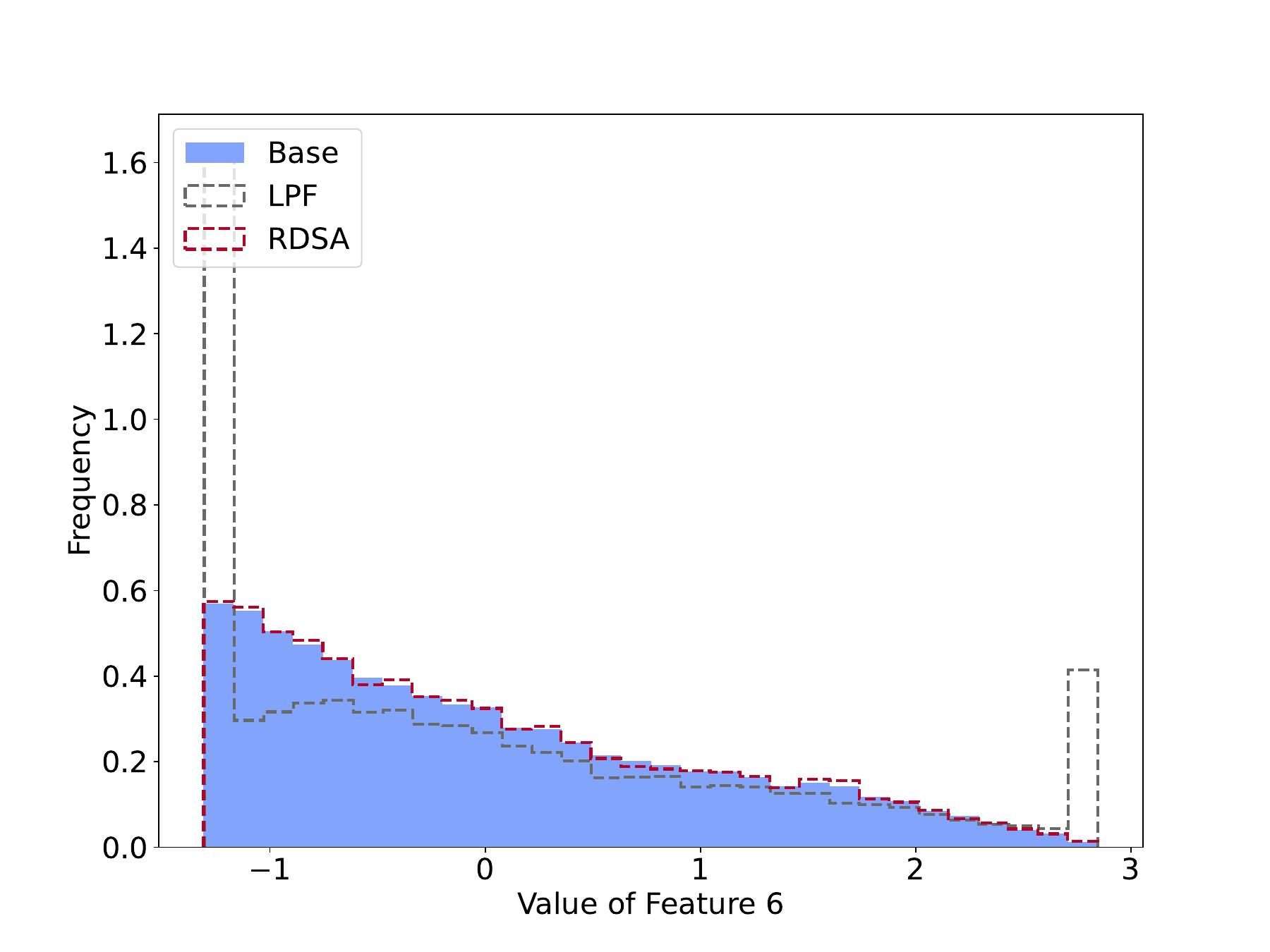}
         \caption{VBF Model}
         \label{fig:DistributionsCompare_VBF}
     \end{subfigure}
     \hfill
     \begin{subfigure}[b]{0.45\textwidth}
         \centering
         \includegraphics[width=\textwidth]{DistributionsCompare_Topo.pdf}
         \caption{TopoDNN}
         \label{fig:DistributionsCompare_Topo}
     \end{subfigure}
     \hfill
     \begin{subfigure}[b]{0.45\textwidth}
         \centering
         \includegraphics[width=\textwidth]{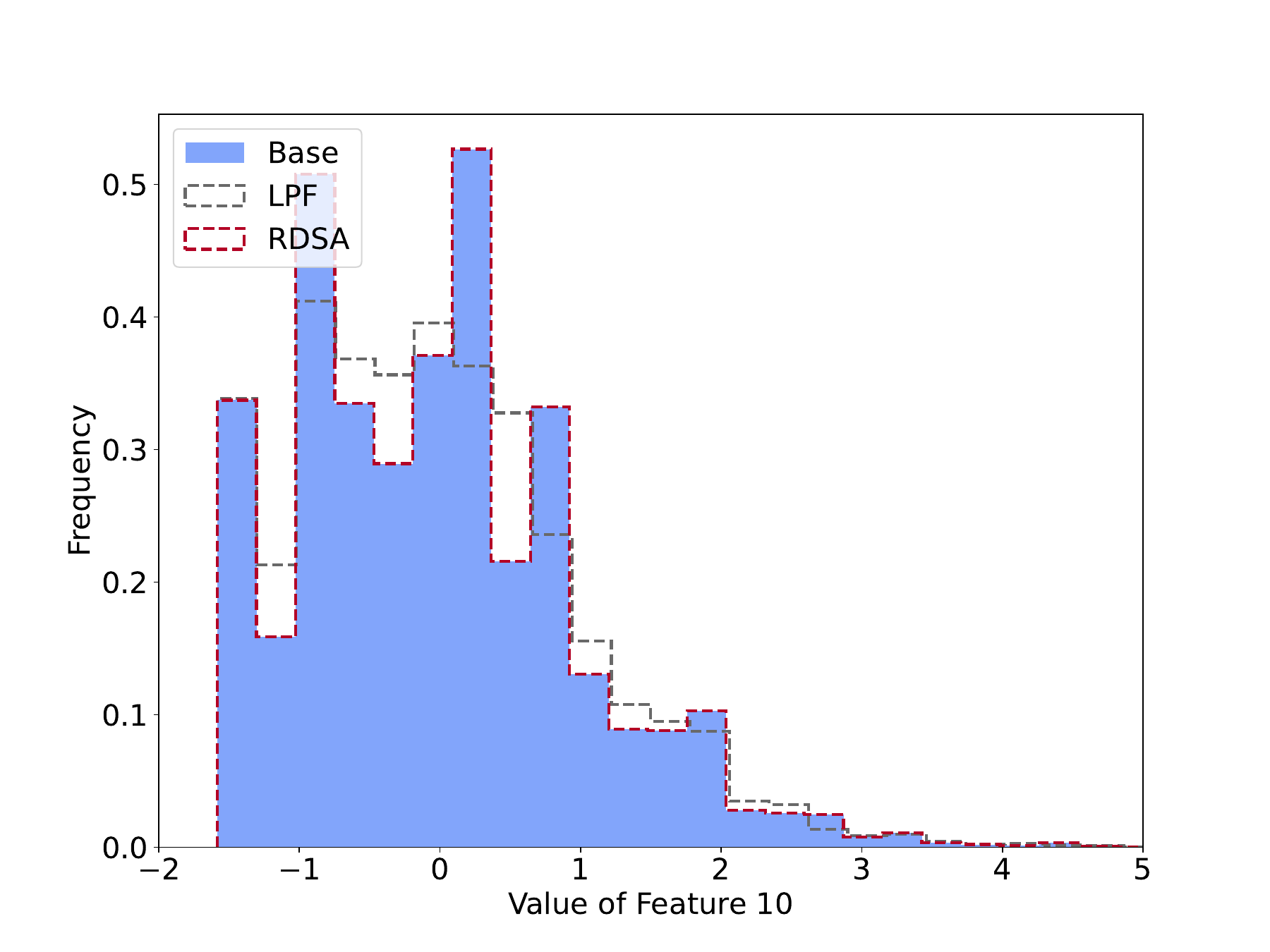}
         \caption{Rain in Australia Model}
         \label{fig:DistributionsCompare_Rain}
     \end{subfigure}
     \hfill
     \begin{subfigure}[b]{0.45\textwidth}
         \centering
         \includegraphics[width=\textwidth]{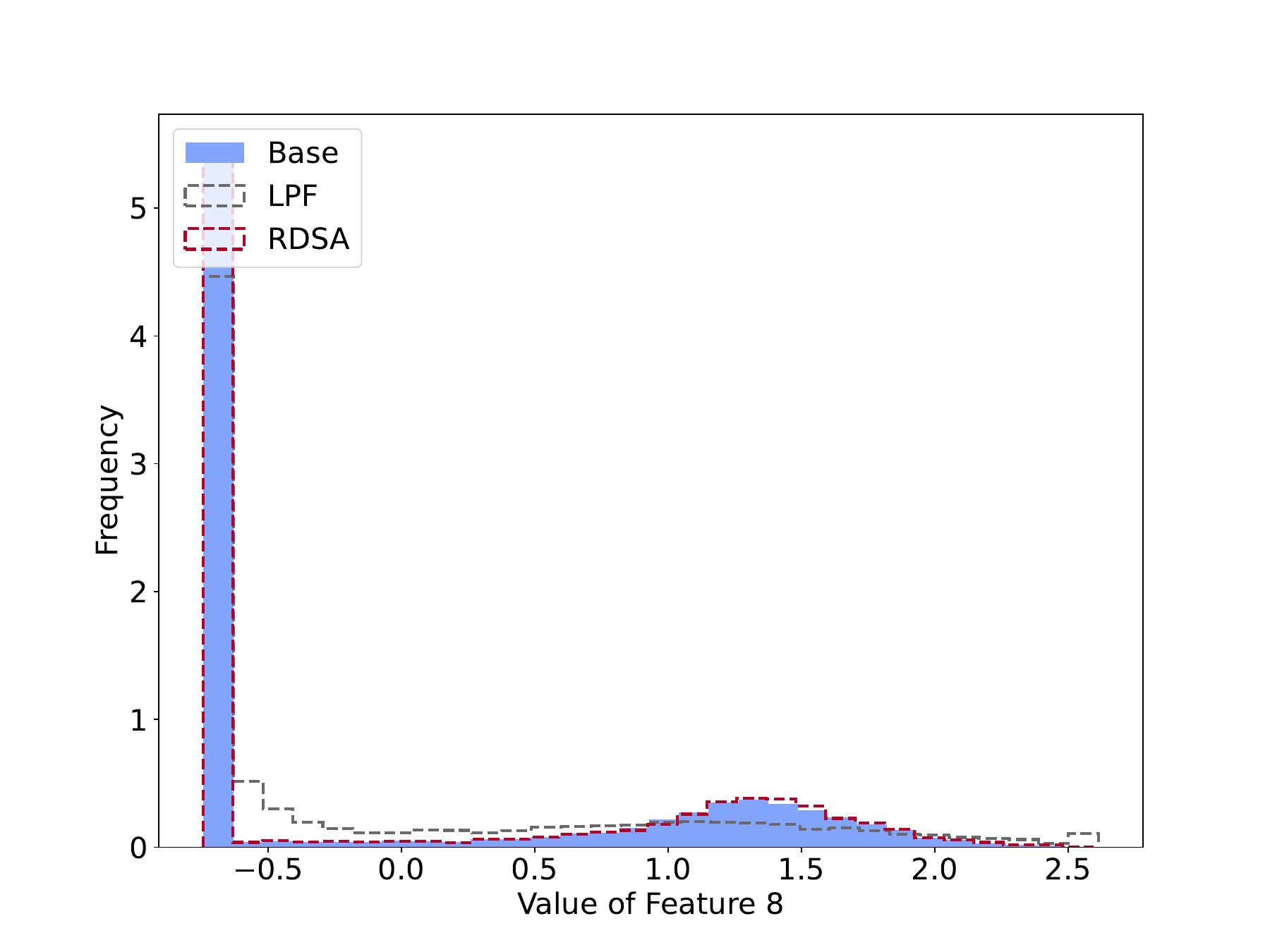}
         \caption{MIMIC-IV Mortality Model}
         \label{fig:DistributionsCompare_MIMICIV}
     \end{subfigure}

     \begin{subfigure}[b]{0.45\textwidth}
         \centering
         \includegraphics[width=\textwidth]{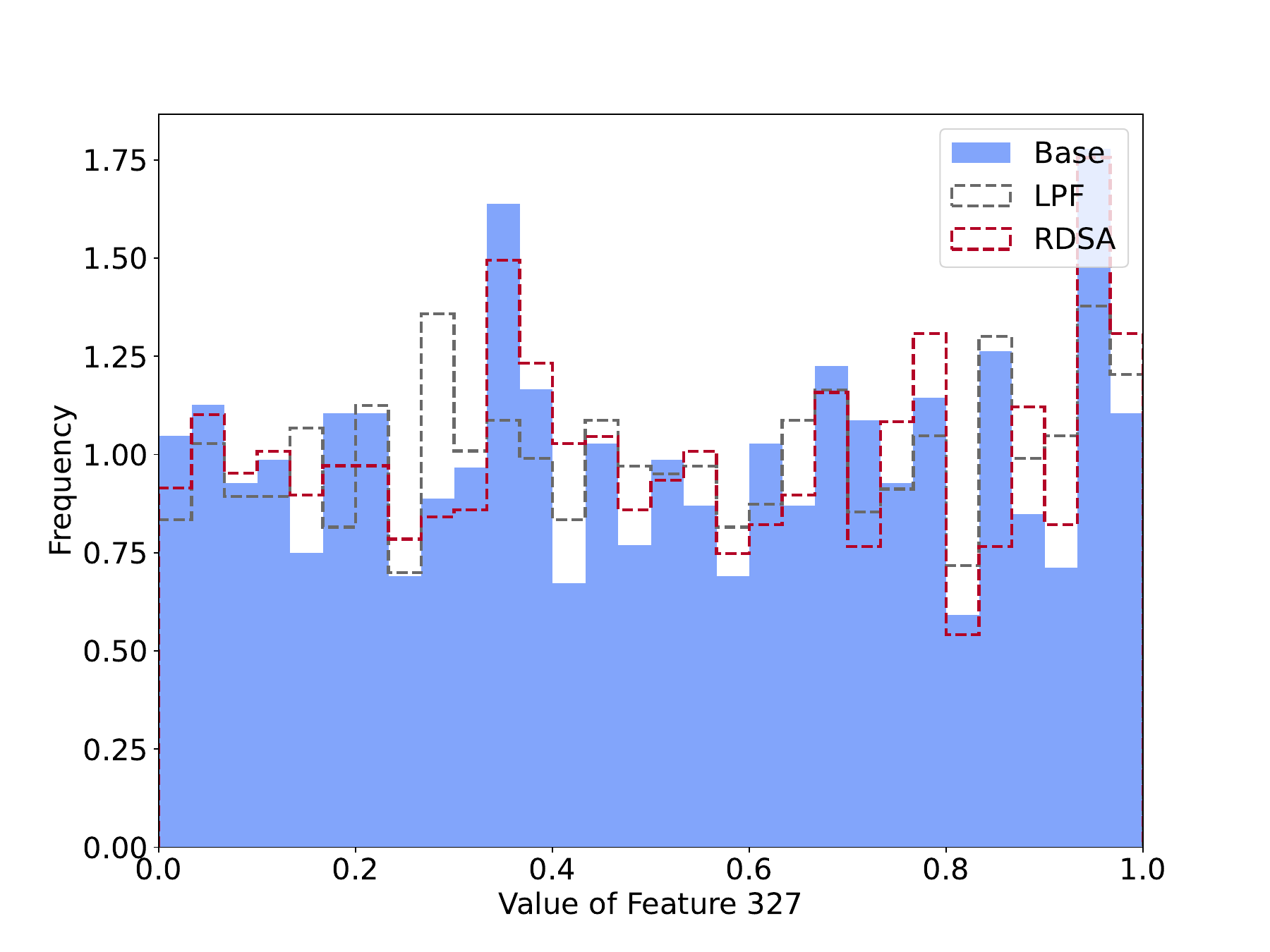}
         \caption{MNIST784 Model}
         \label{fig:DistributionsCompare_MNIST784}
     \end{subfigure}
     \hfill
     \begin{subfigure}[b]{0.45\textwidth}
         \centering
         \includegraphics[width=\textwidth]{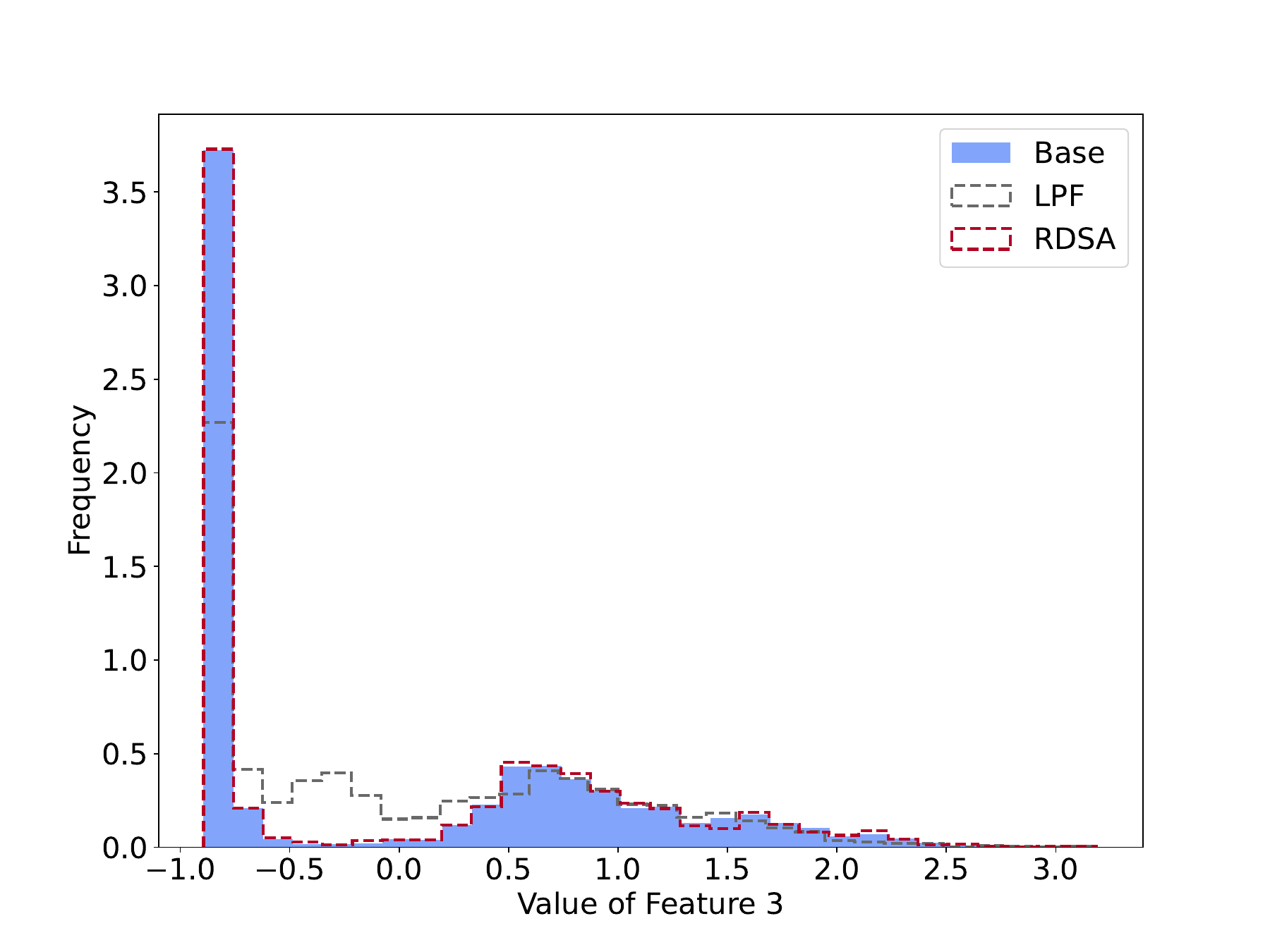}
         \caption{HAR Model}
         \label{fig:DistributionsCompare_HAR}
     \end{subfigure}
        \caption{Comparison of distributions of base test datasets, adversarial sets generated with LPF, and adversarial sets generated using RDSA.}
        \label{fig:DistributionsCompare_App}
\end{figure}

\begin{figure}[H]
     \centering
     \begin{subfigure}[b]{0.45\textwidth}
         \centering
         \includegraphics[width=\textwidth]{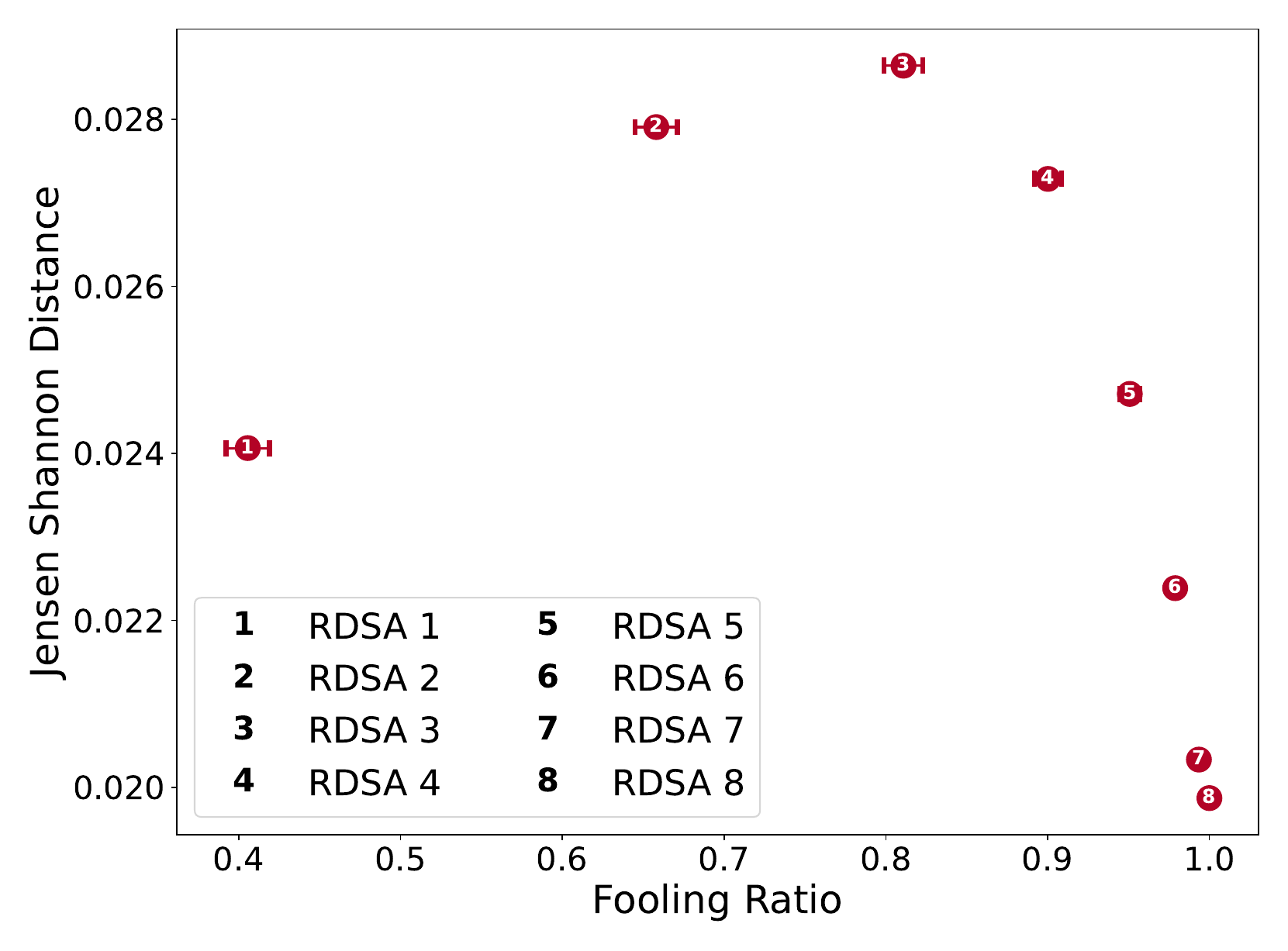}
         \caption{VBF Model}
         \label{fig:JSD_RDSA_VBF}
     \end{subfigure}
     \hfill
     \begin{subfigure}[b]{0.45\textwidth}
         \centering
         \includegraphics[width=\textwidth]{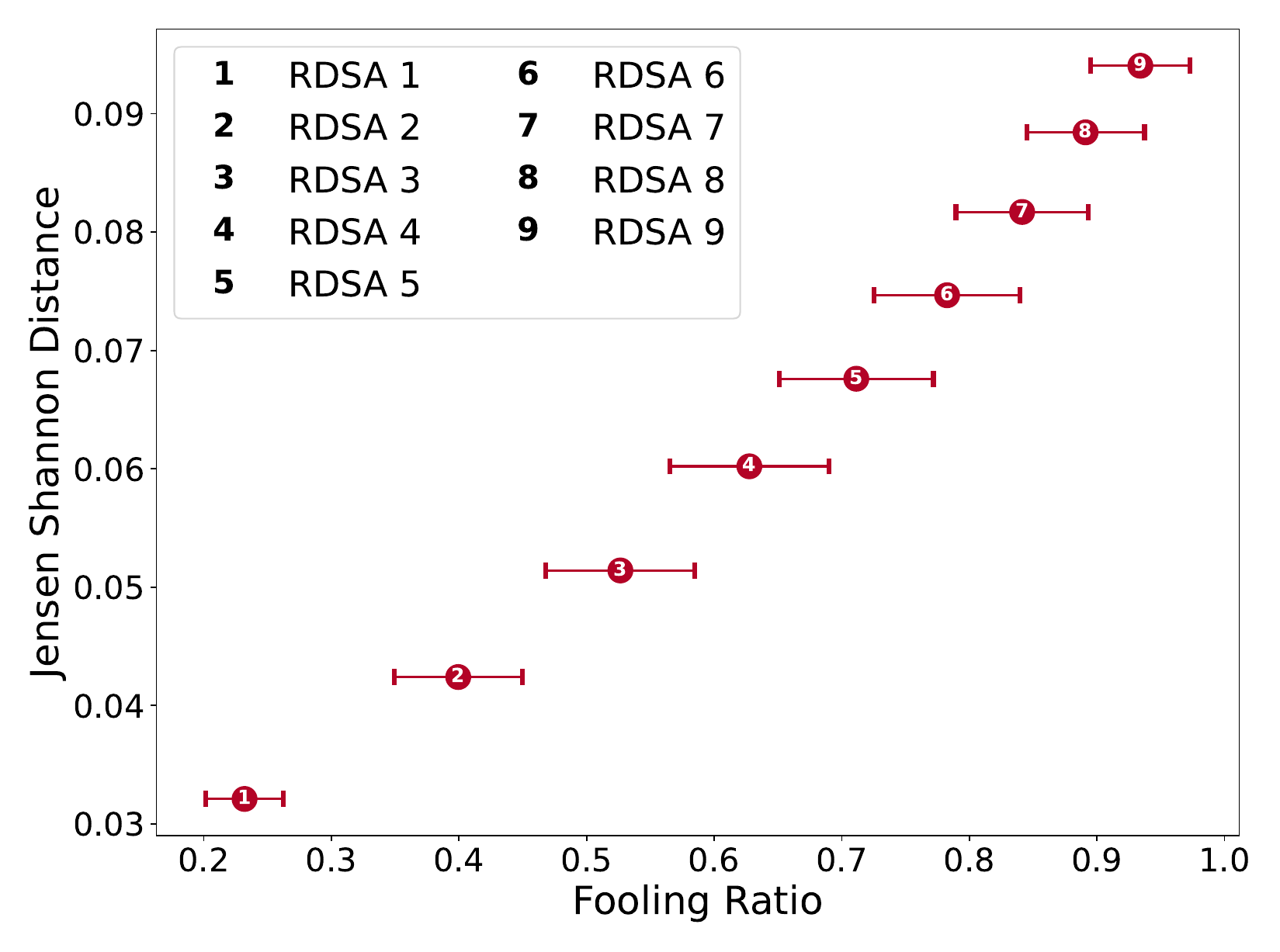}
         \caption{Rain in Australia Model}
         \label{fig:JSD_RDSA_Rain}
     \end{subfigure}
     \hfill
     \begin{subfigure}[b]{0.45\textwidth}
         \centering
         \includegraphics[width=\textwidth]{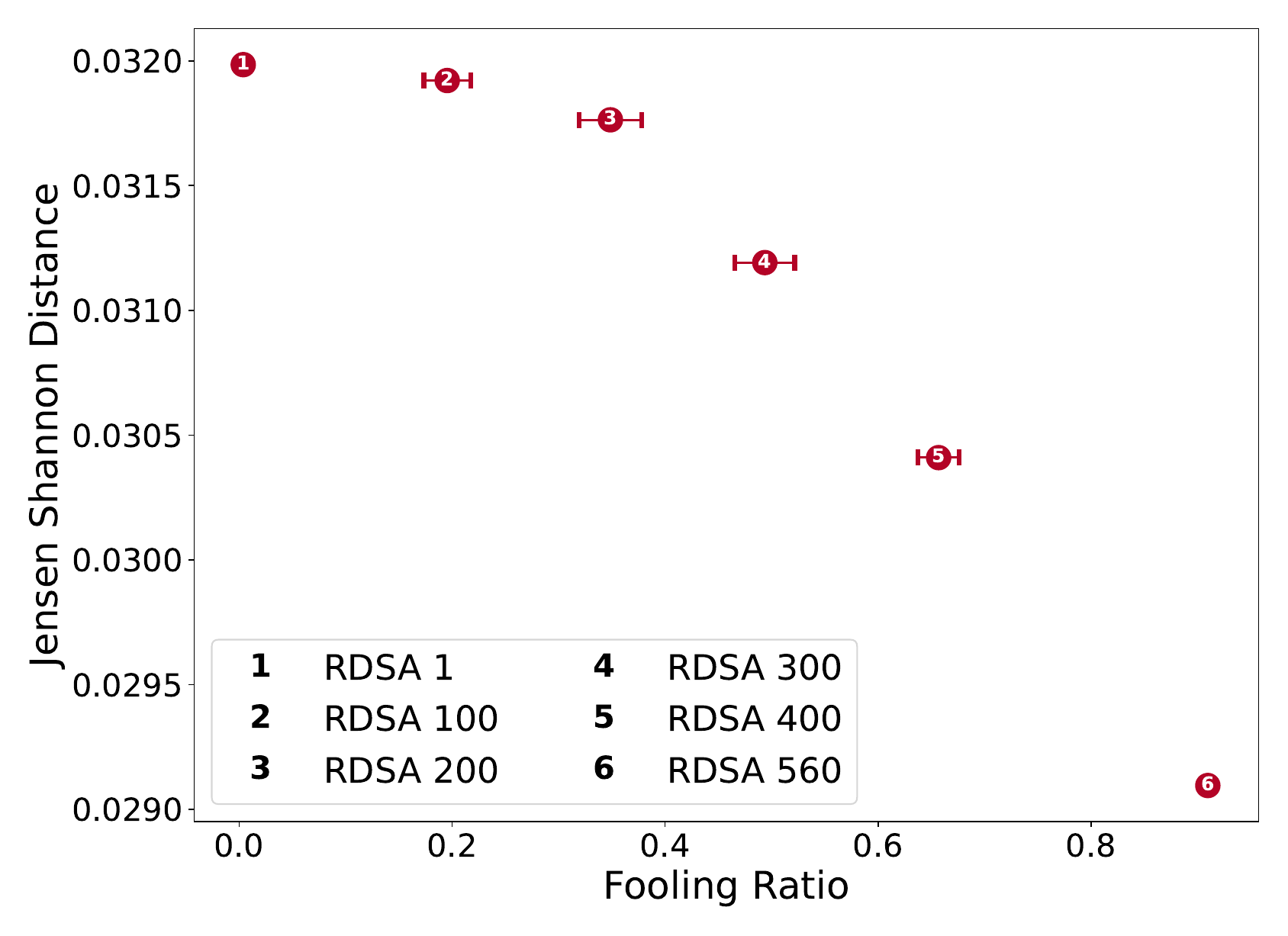}
         \caption{MNIST784 Model}
         \label{fig:JSD_RDSA_MNIST784}
     \end{subfigure}
     \hfill
     \begin{subfigure}[b]{0.45\textwidth}
         \centering
         \includegraphics[width=\textwidth]{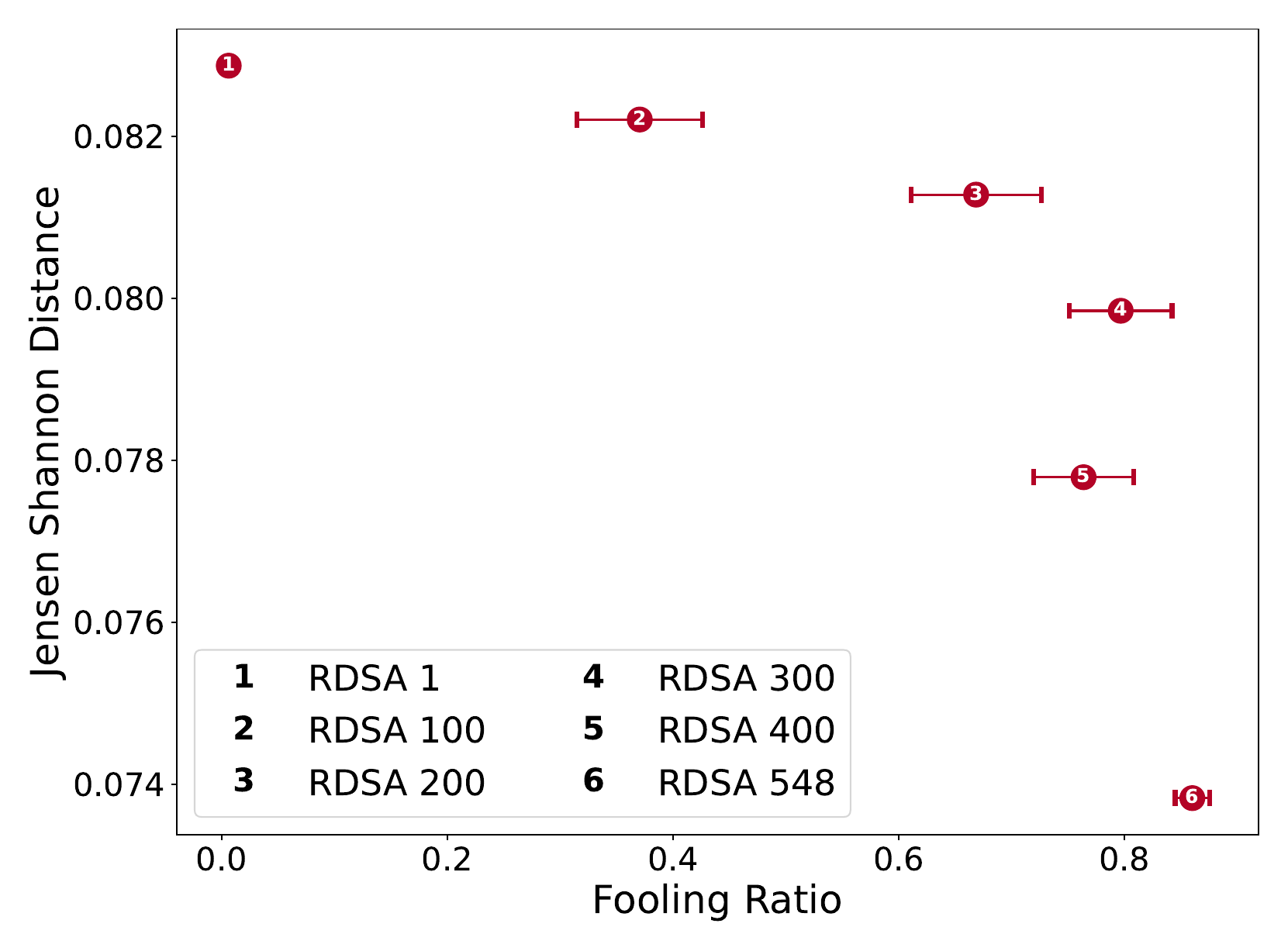}
         \caption{HAR Model}
         \label{fig:JSD_RDSA_HAR}
     \end{subfigure}
        \caption{Average Jensen-Shannon Distances between the initial distributions and the adversarial distributions for different attacks applied on the models.}
        \label{fig:JSD_RDSA_App}
\end{figure}

\begin{figure}[H]
     \centering
     \begin{subfigure}[b]{0.45\textwidth}
         \centering
         \includegraphics[width=\textwidth]{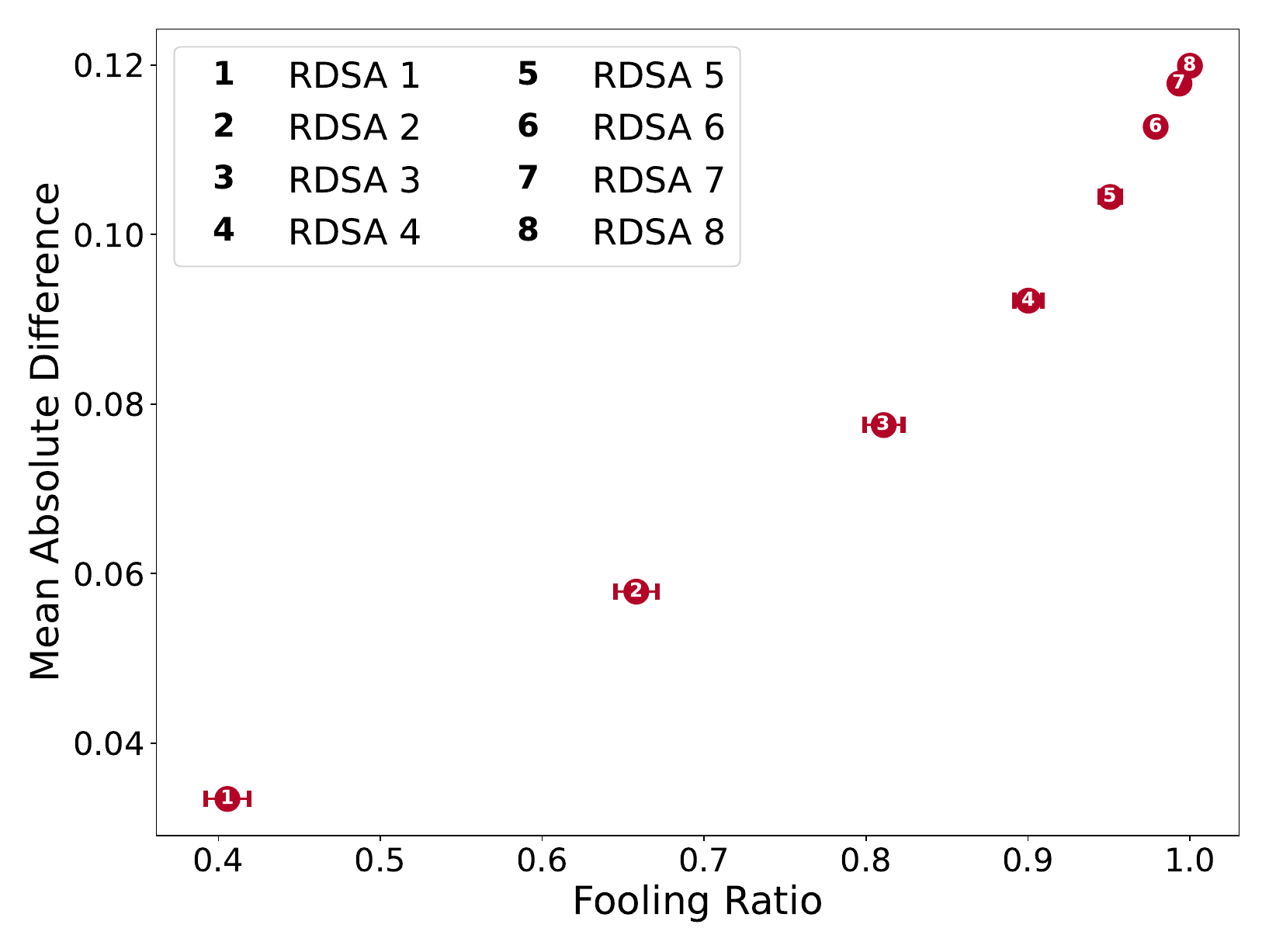}
         \caption{VBF Model}
         \label{fig:Mean_Difference_Correlation_RDSA_VBF}
     \end{subfigure}
     \hfill
     \begin{subfigure}[b]{0.45\textwidth}
         \centering
         \includegraphics[width=\textwidth]{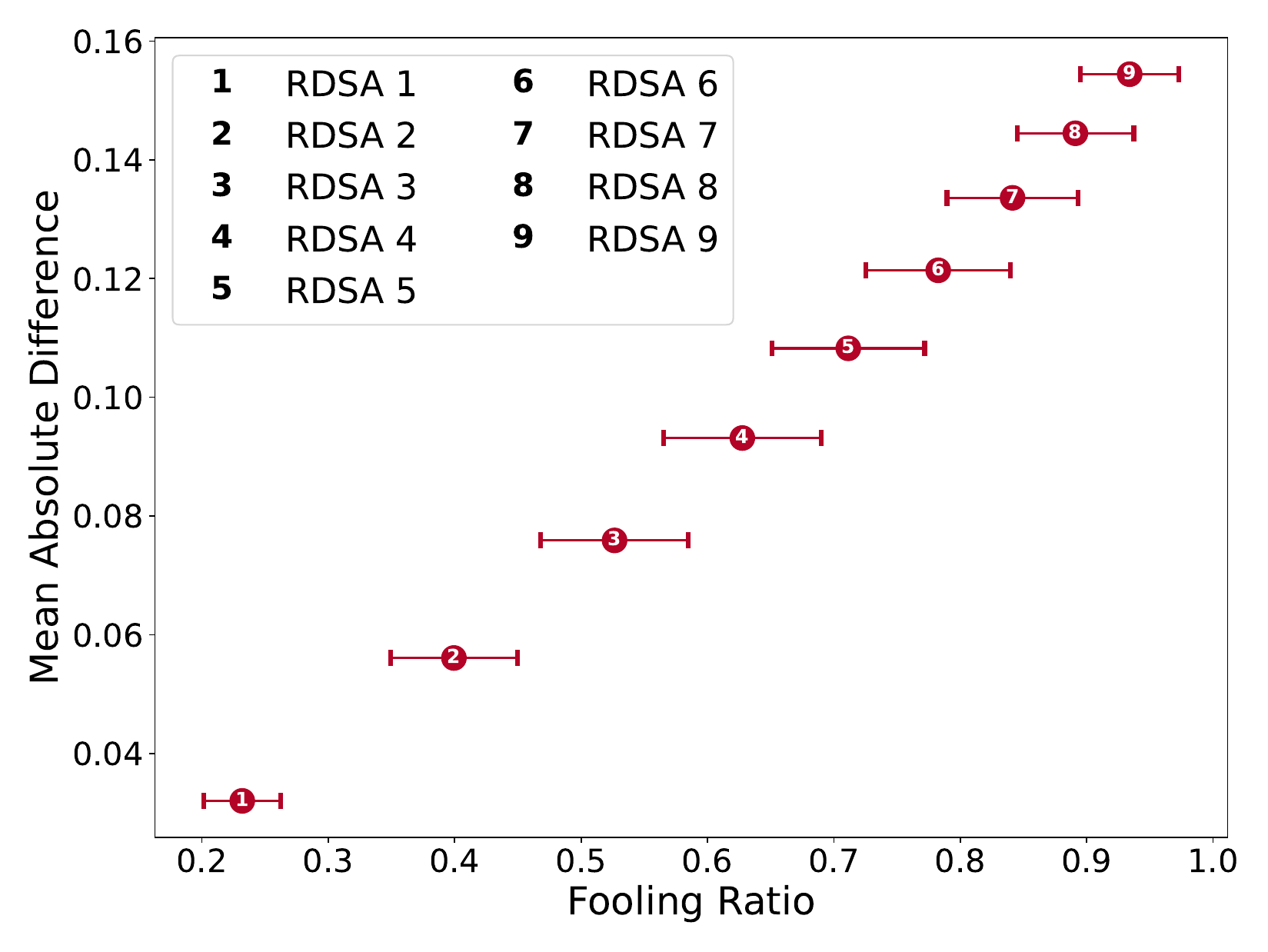}
         \caption{Rain in Australia Model}
         \label{fig:Mean_Difference_Correlation_RDSA_Rain}
     \end{subfigure}
     \hfill
     \hfill
     \begin{subfigure}[b]{0.45\textwidth}
         \centering
         \includegraphics[width=\textwidth]{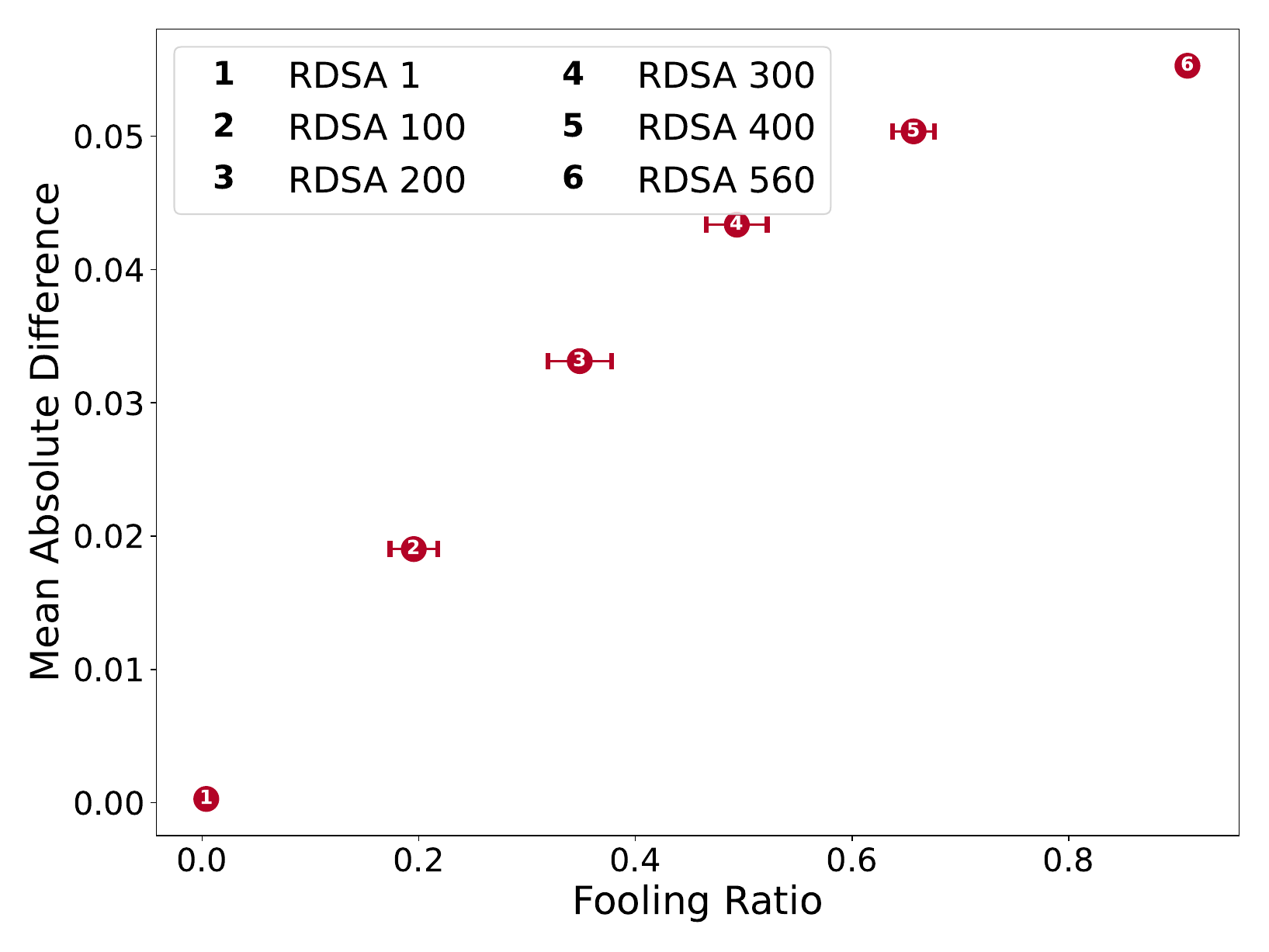}
         \caption{MNIST784 Model}
         \label{fig:Mean_Difference_Correlation_RDSA_MNIST784}
     \end{subfigure}
     \hfill
     \begin{subfigure}[b]{0.45\textwidth}
         \centering
         \includegraphics[width=\textwidth]{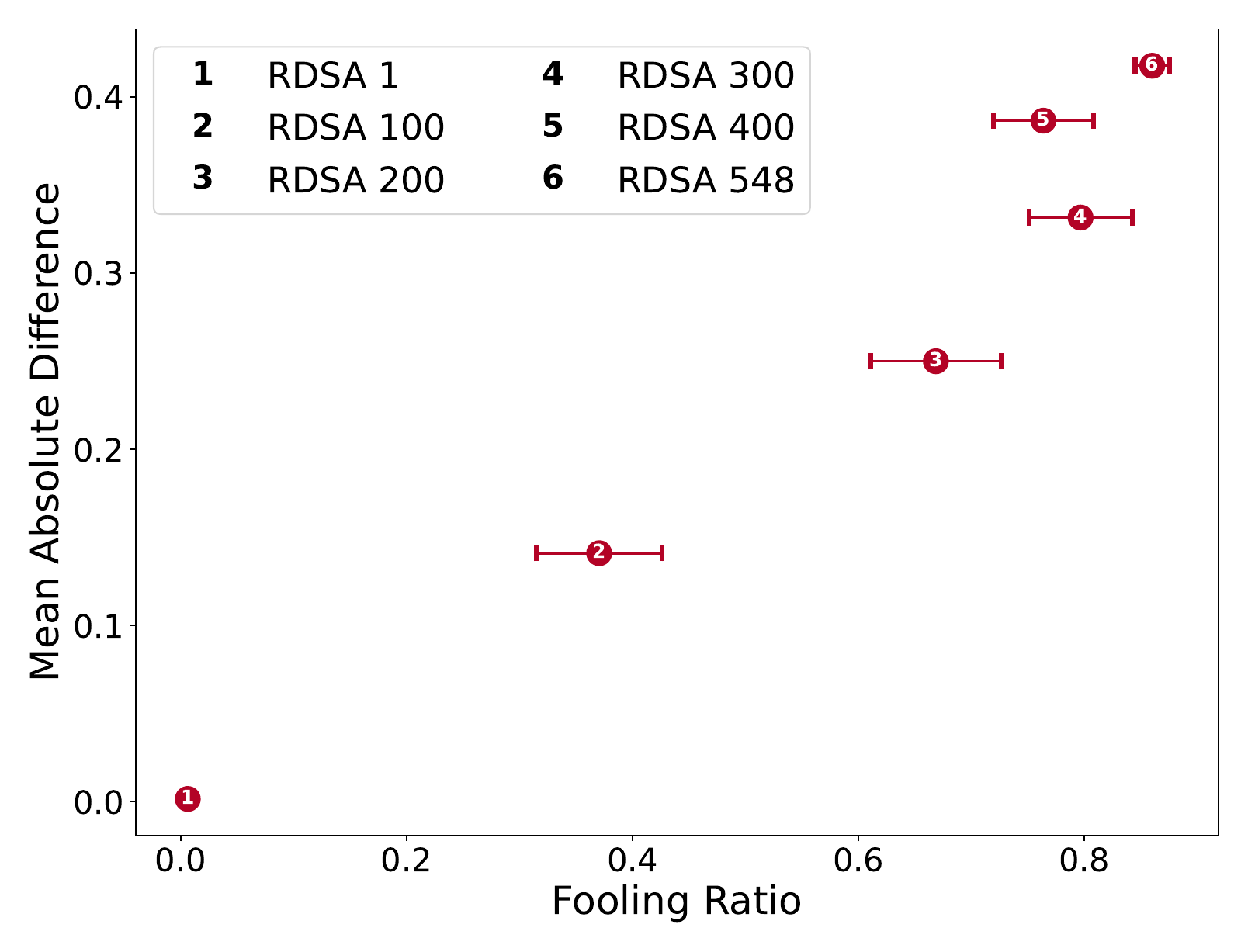}
         \caption{HAR Model}
         \label{fig:Mean_Difference_Correlation_RDSA_HAR}
     \end{subfigure}
     \hfill
        \caption{Average absolute difference between the clean correlation matrices and the adversarial correlation matrices for different attacks applied on the models.}
        \label{fig:Mean_Difference_Correlation_RDSA_App}
\end{figure}

\subsection{Data Augmentation}

\begin{figure}[H]
     \centering
     \begin{subfigure}[b]{0.4\textwidth}
         \centering
         \includegraphics[width=\textwidth]{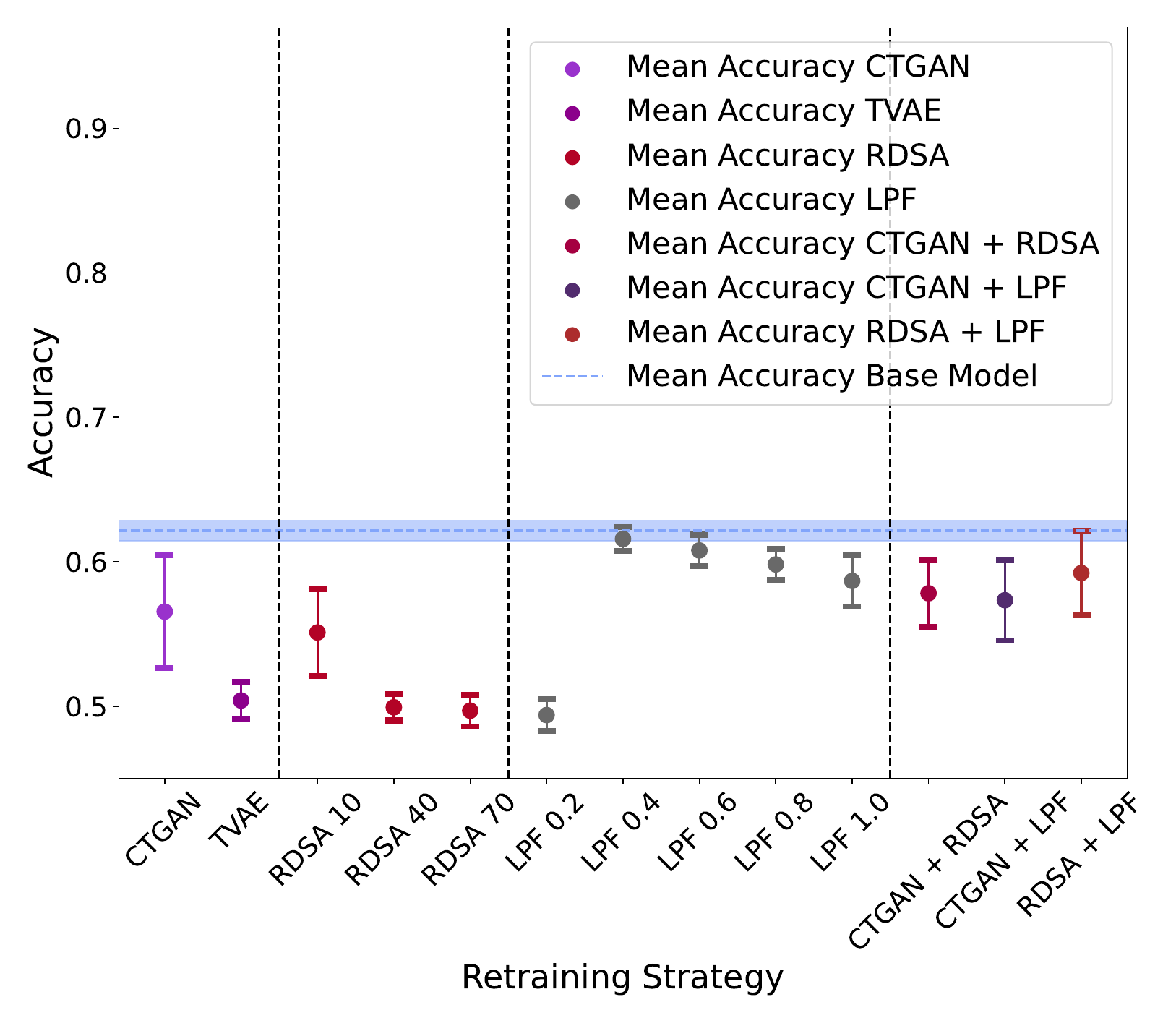}
         \caption{VBF Model}
         \label{fig:Accuracy_Retr_VBF}
     \end{subfigure}
     \begin{subfigure}[b]{0.4\textwidth}
         \centering
         \includegraphics[width=\textwidth]{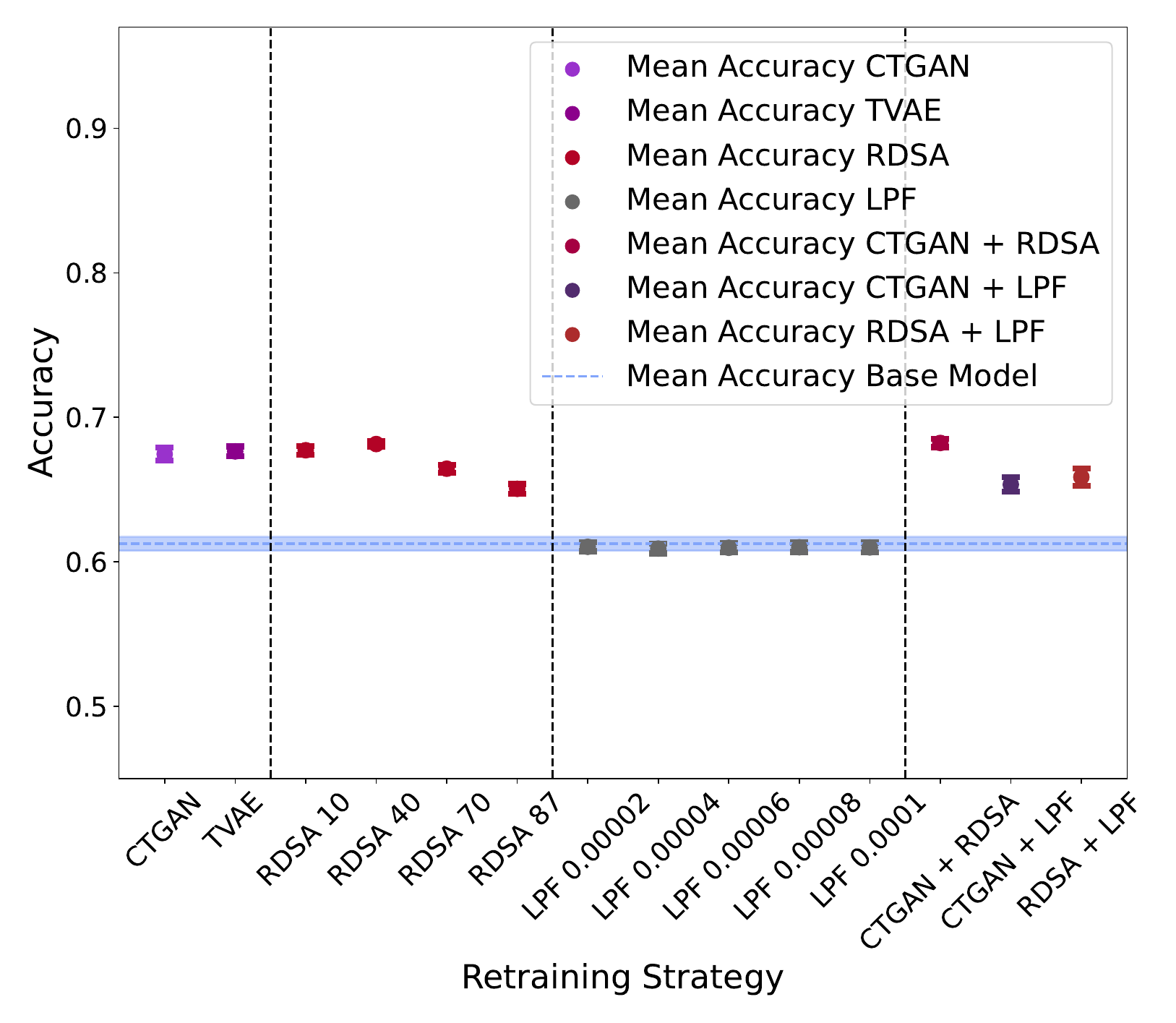}
         \caption{TopoDNN}
         \label{fig:Accuracy_Retr_Topo}
     \end{subfigure}
     \hfill
     \begin{subfigure}[b]{0.4\textwidth}
         \centering
         \includegraphics[width=\textwidth]{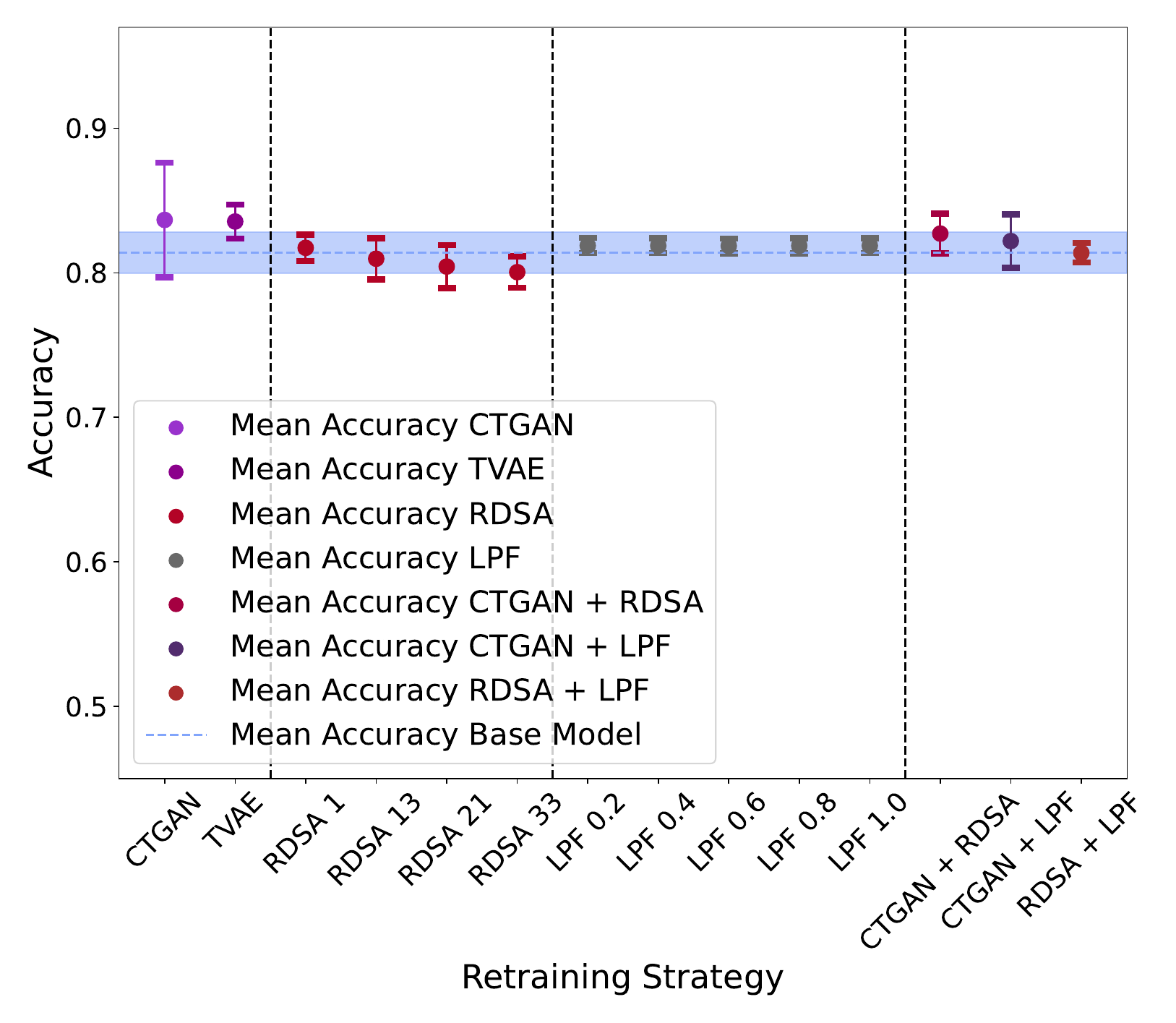}
         \caption{MIMIC-IV Mortality Model}
         \label{fig:Accuracy_Retr_MIMICIV}
     \end{subfigure}
     \begin{subfigure}[b]{0.4\textwidth}
         \centering
         \includegraphics[width=\textwidth]{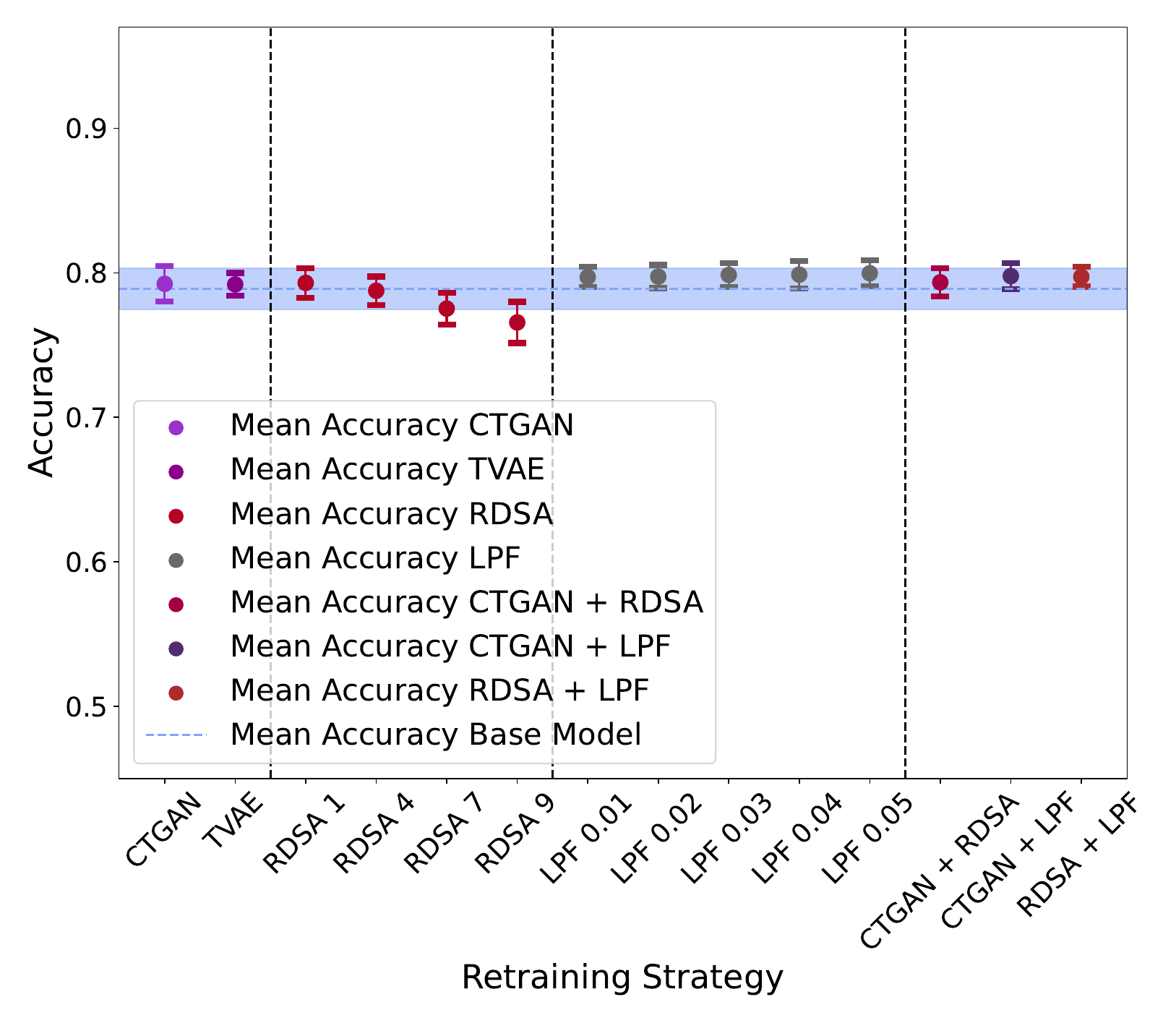}
         \caption{Rain in Australia Model}
         \label{fig:Accuracy_Retr_Rain}
     \end{subfigure}
     \hfill
     
     \begin{subfigure}[b]{0.4\textwidth}
         \centering
         \includegraphics[width=\textwidth]{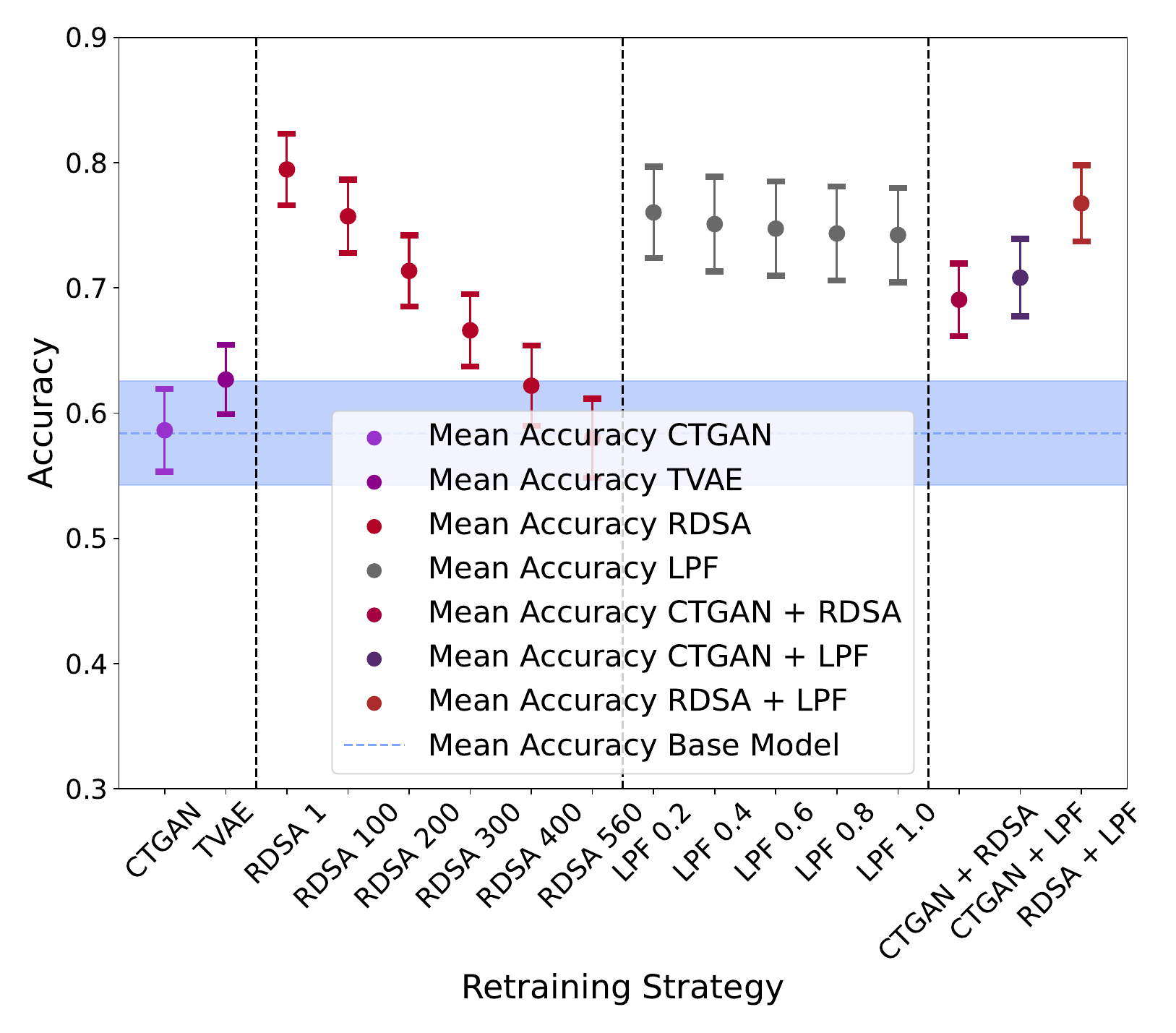}
         \caption{MNIST784 Model}
         \label{fig:Accuracy_Retr_MNIST784}
     \end{subfigure}
     \begin{subfigure}[b]{0.4\textwidth}
         \centering
         \includegraphics[width=\textwidth]{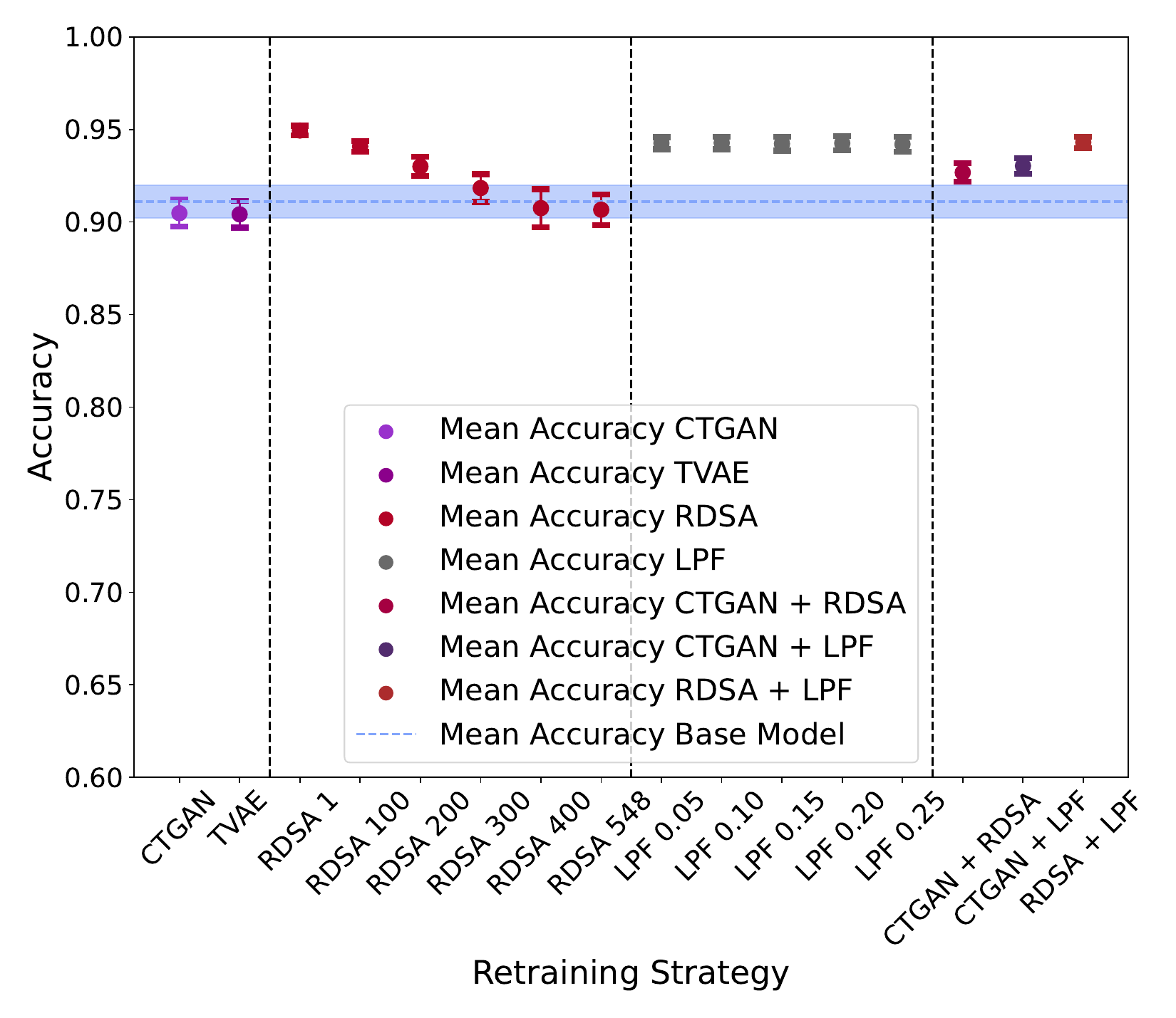}
         \caption{HAR Model}
         \label{fig:Accuracy_Retr_HAR}
     \end{subfigure}
        \caption{Mean accuracy of varying data augmentation strategies, where the error bars are given by standard deviations of the accuracy values encountered during the runs. }
        \label{fig:Accuracy_Retr_App}
\end{figure}

\section{LPF Attack Results}
\label{App:AttackResults}
\subsection{Difference per Event}

\begin{figure}[H]
     \centering
     \begin{subfigure}[b]{0.45\textwidth}
         \centering
         \includegraphics[width=\textwidth]{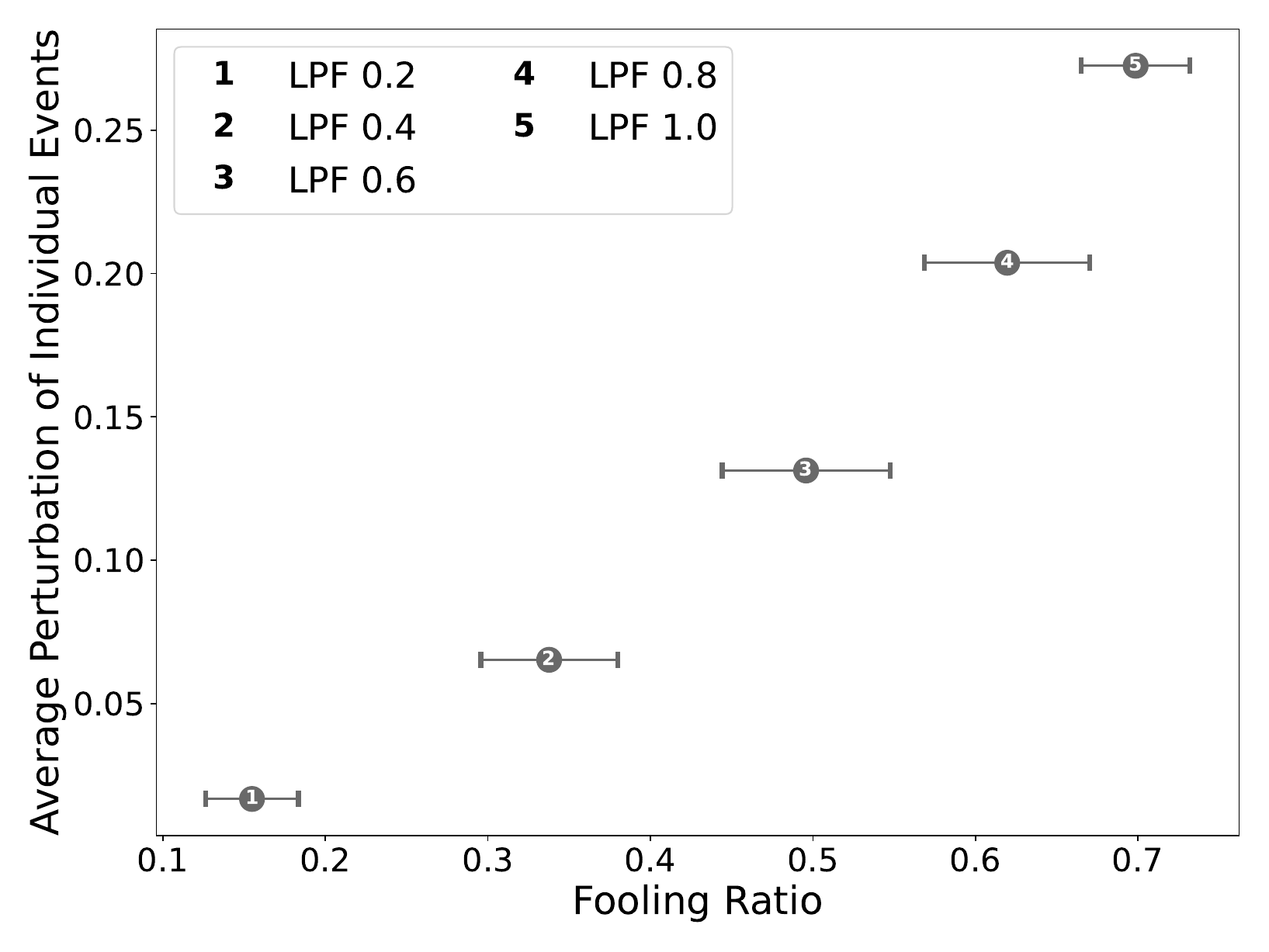}
         \caption{VBF Model}
         \label{fig:Event_Diff_FR_LPF_VBF}
     \end{subfigure}
     \hfill
     \begin{subfigure}[b]{0.45\textwidth}
         \centering
         \includegraphics[width=\textwidth]{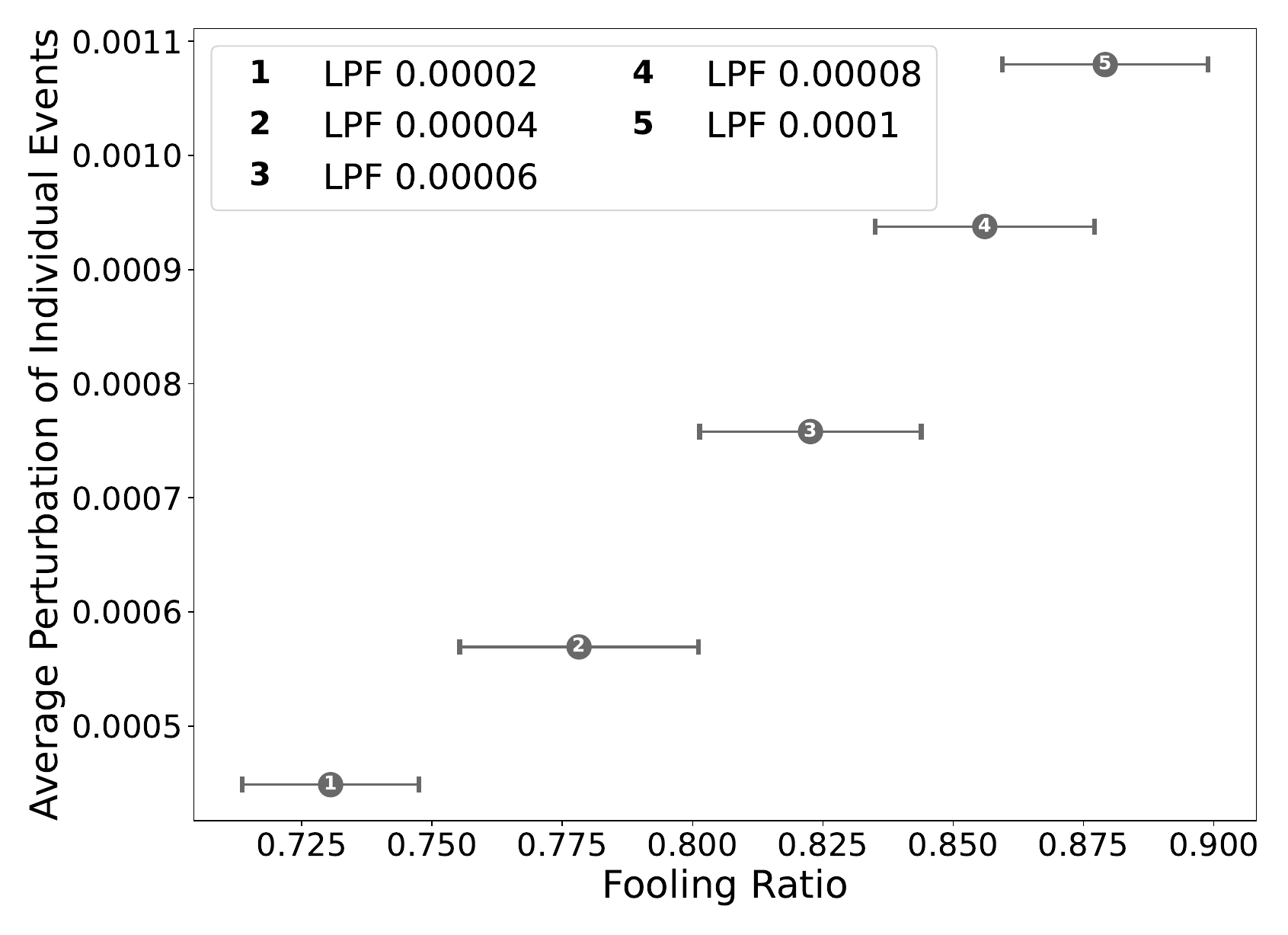}
         \caption{TopoDNN}
         \label{fig:Event_Diff_FR_LPF_Topo}
     \end{subfigure}
     \hfill
     \begin{subfigure}[b]{0.45\textwidth}
         \centering
         \includegraphics[width=\textwidth]{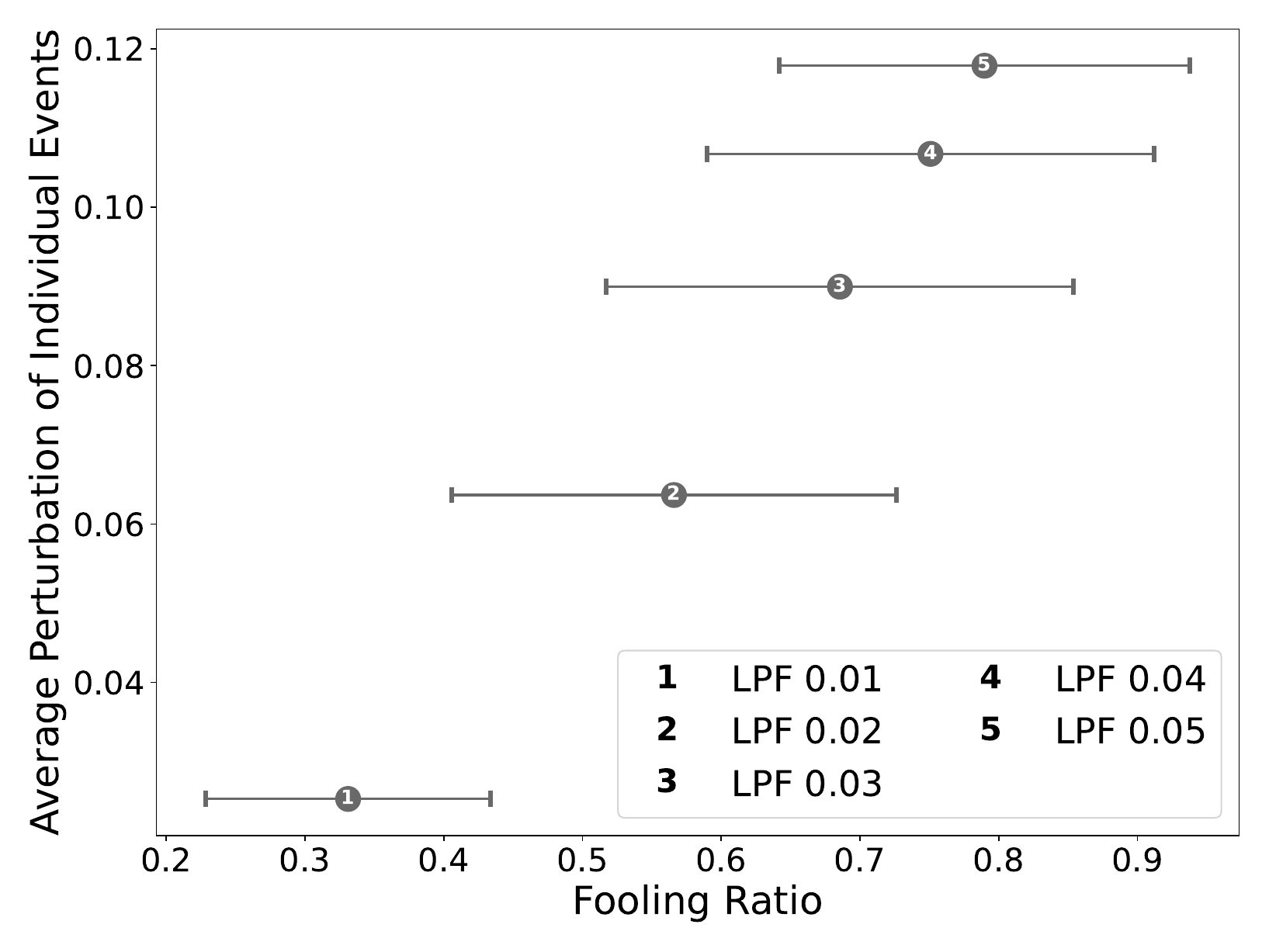}
         \caption{Rain in Australia Model}
         \label{fig:Event_Diff_FR_LPF_Rain}
     \end{subfigure}
     \hfill
     \begin{subfigure}[b]{0.45\textwidth}
         \centering
         \includegraphics[width=\textwidth]{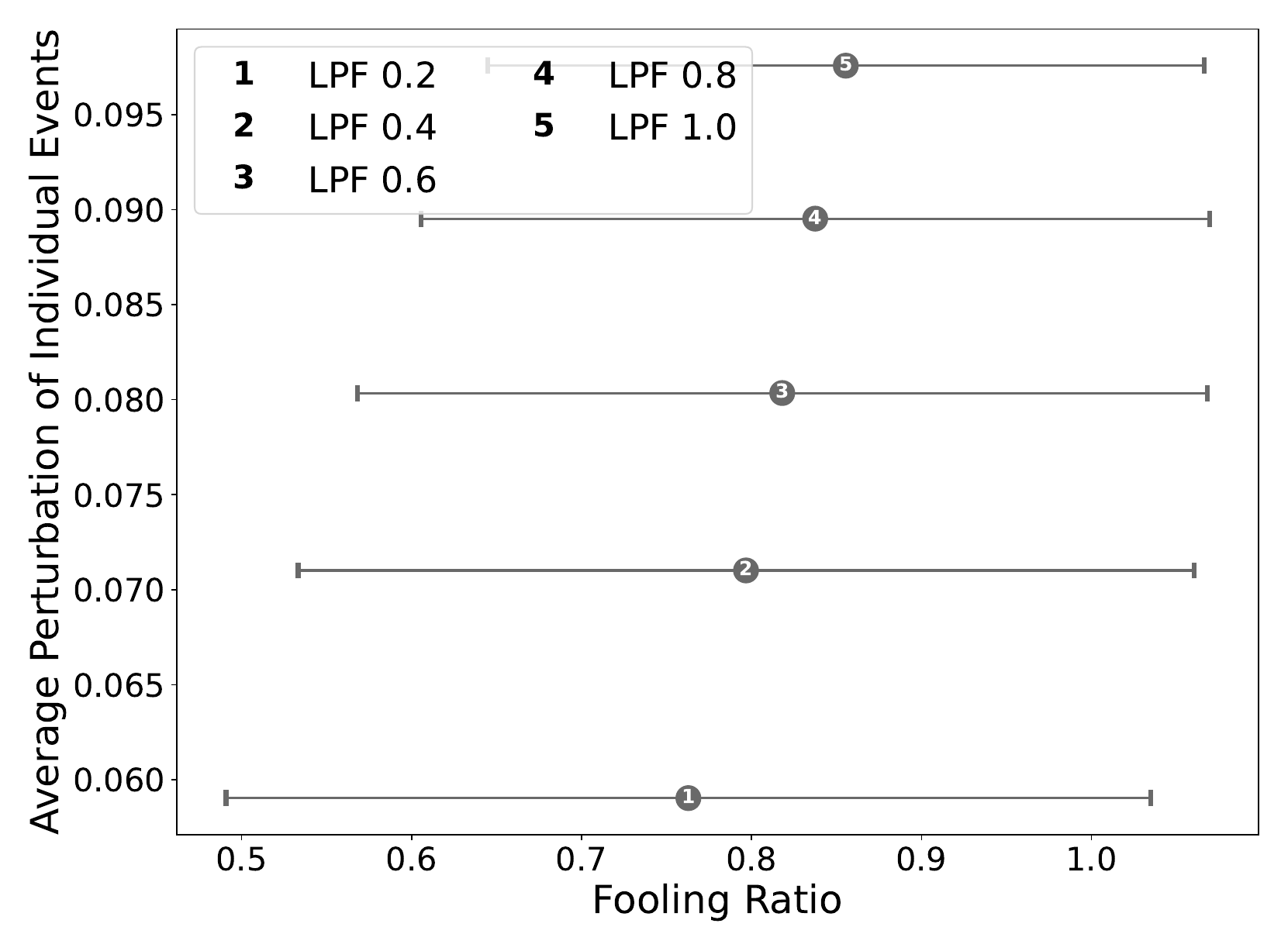}
         \caption{MIMIC-IV Mortality Model}
         \label{fig:Event_Diff_FR_LPF_MIMICIV}
     \end{subfigure}
     \hfill
     \begin{subfigure}[b]{0.45\textwidth}
         \centering
         \includegraphics[width=\textwidth]{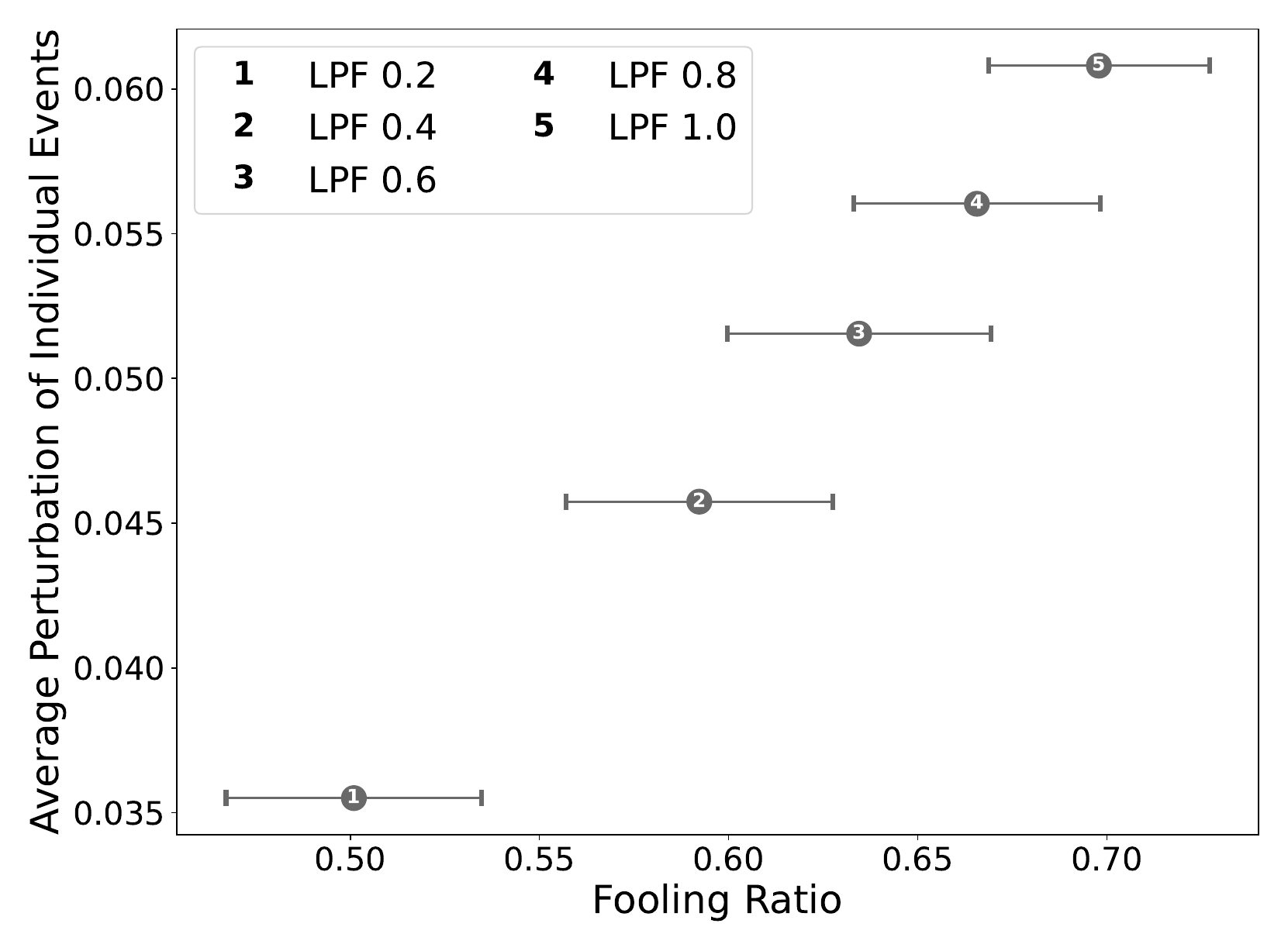}
         \caption{MNIST784 Model}
         \label{fig:Event_Diff_FR_LPF_MNIST784}
     \end{subfigure}
     \hfill
     \begin{subfigure}[b]{0.45\textwidth}
         \centering
         \includegraphics[width=\textwidth]{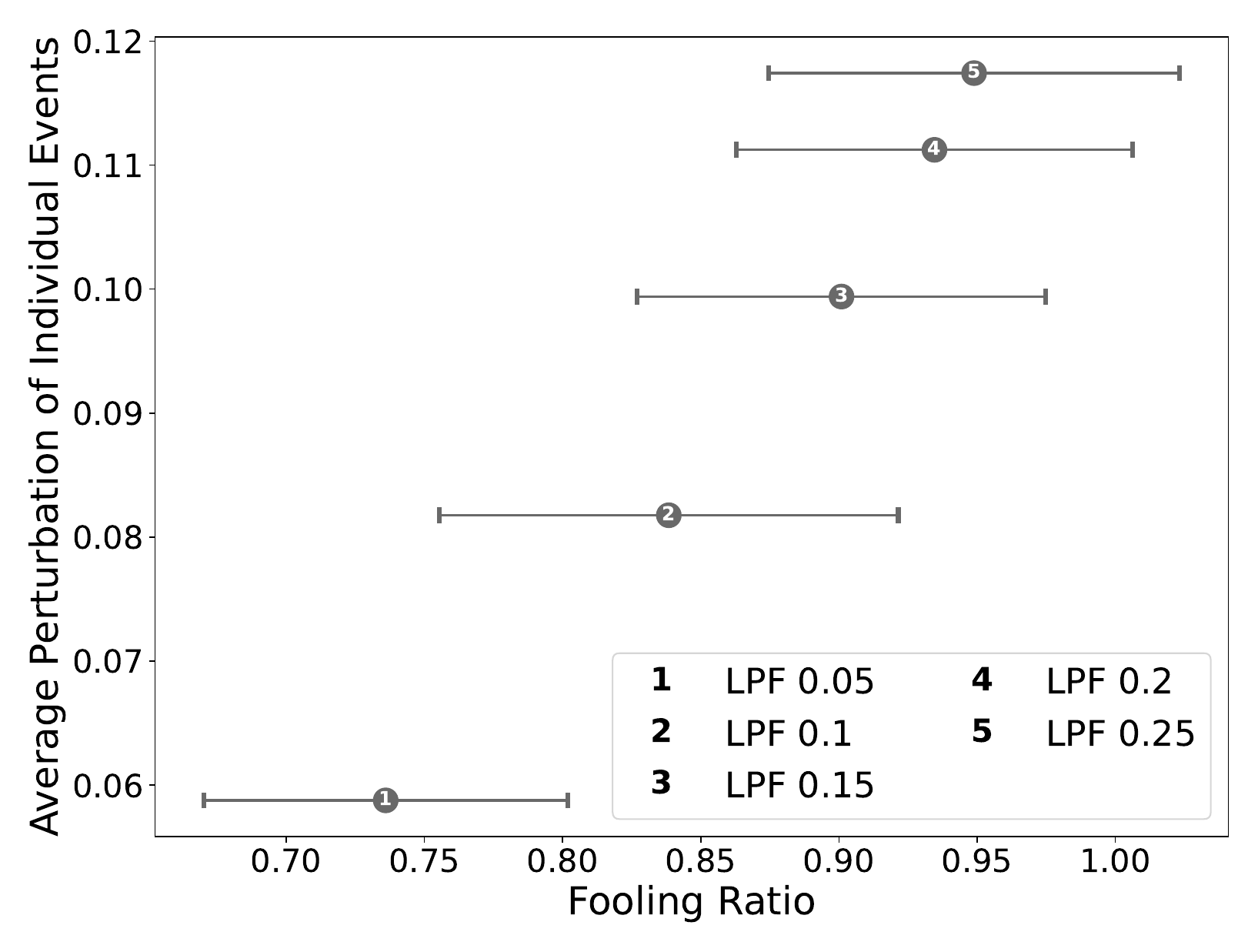}
         \caption{HAR Model}
         \label{fig:Event_Diff_FR_LPF_HAR}
     \end{subfigure}

        \caption{Average difference / perturbation between individual clean inputs (test set) and corresponding adversaries.}
        \label{fig:Event_Diff_FR_LPF_App}
\end{figure}

\subsection{Jensen-Shannon Distance}

\begin{figure}[H]
     \centering
     \begin{subfigure}[b]{0.45\textwidth}
         \centering
         \includegraphics[width=\textwidth]{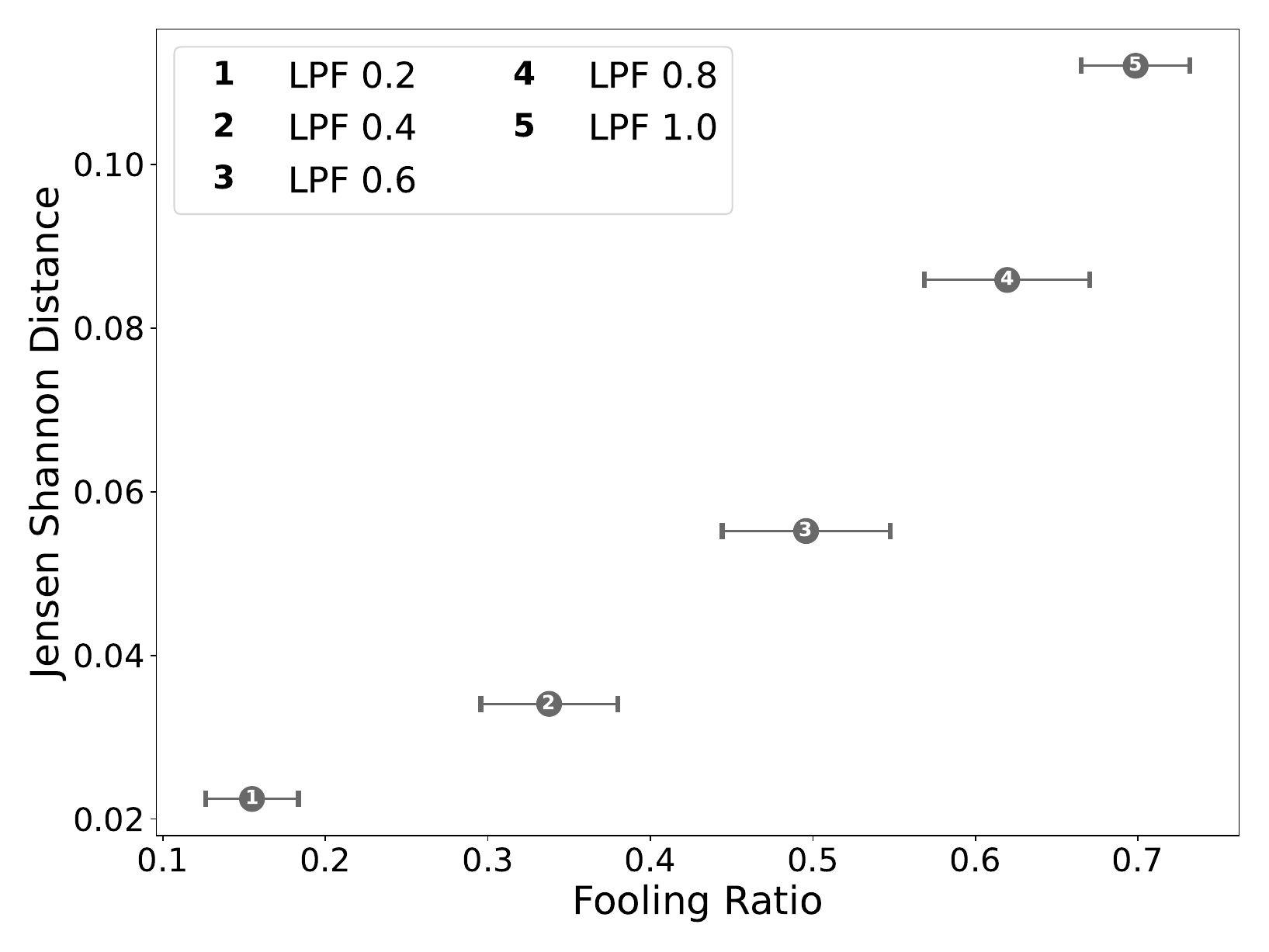}
         \caption{VBF Model}
         \label{fig:JSD_LPF_VBF}
     \end{subfigure}
     \hfill
     \begin{subfigure}[b]{0.45\textwidth}
         \centering
         \includegraphics[width=\textwidth]{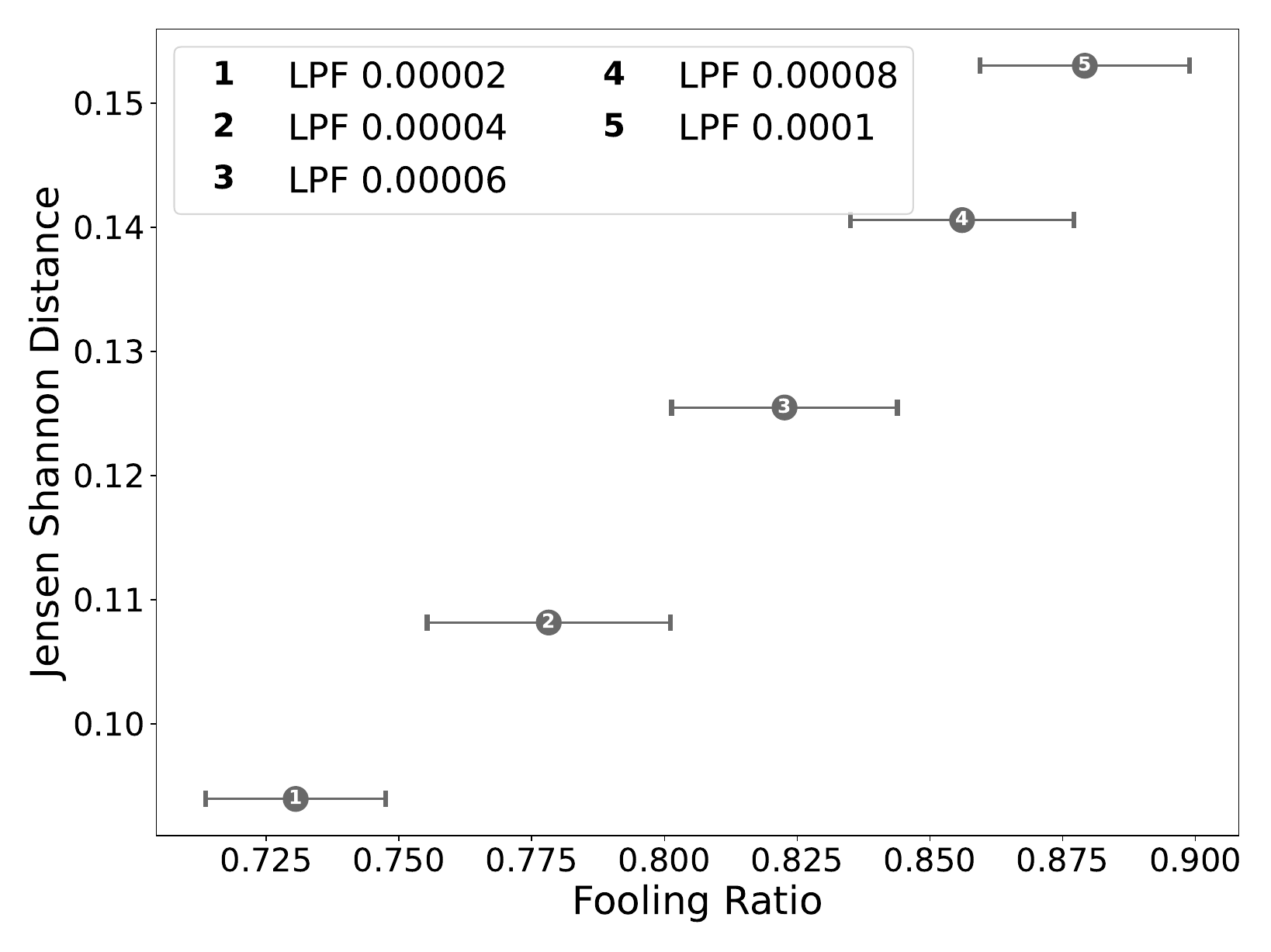}
         \caption{TopoDNN}
         \label{fig:JSD_LPF_Topo}
     \end{subfigure}
     \hfill
     \begin{subfigure}[b]{0.45\textwidth}
         \centering
         \includegraphics[width=\textwidth]{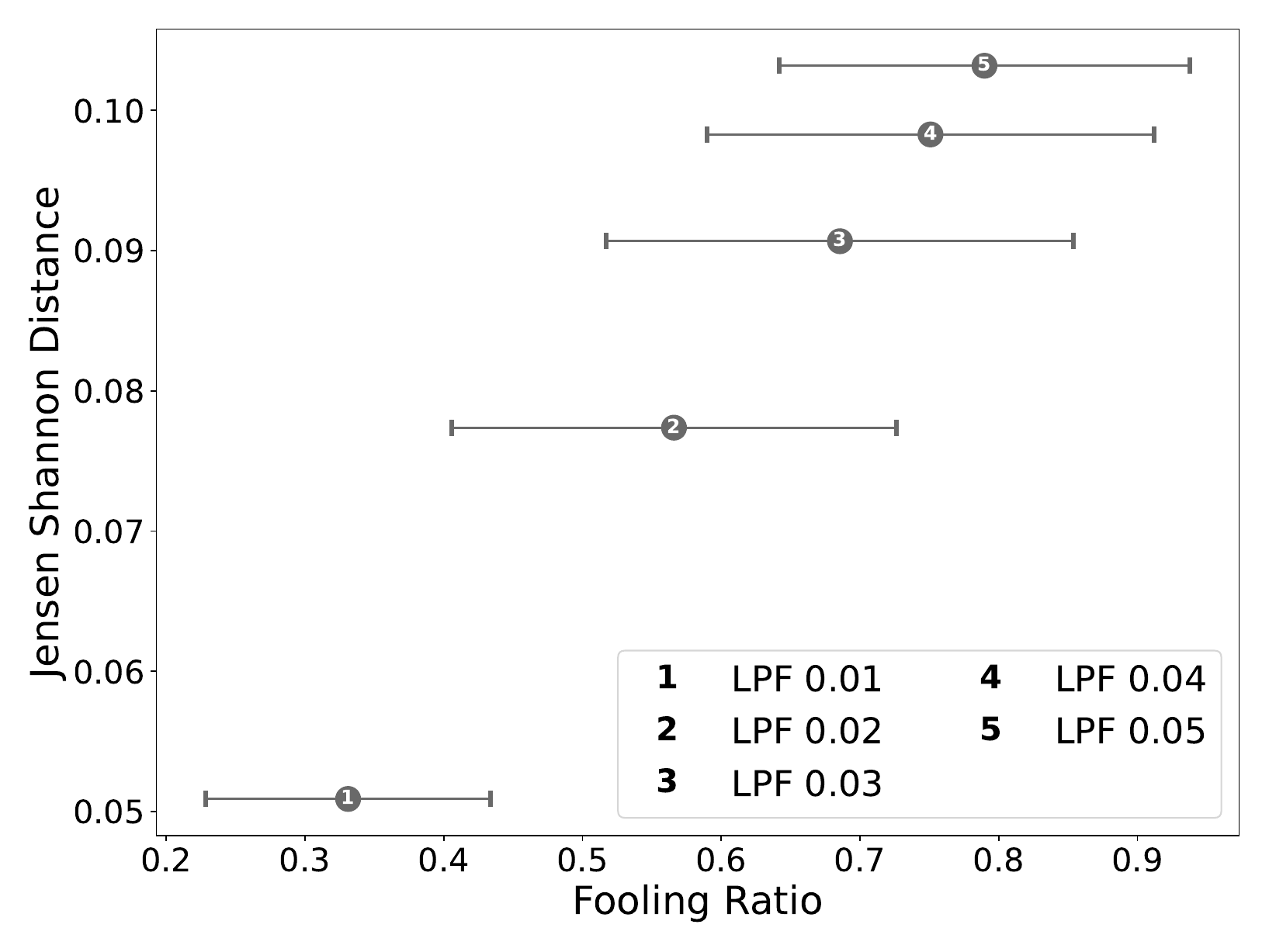}
         \caption{Rain in Australia Model}
         \label{fig:JSD_LPF_Rain}
     \end{subfigure}
     \hfill
     \begin{subfigure}[b]{0.45\textwidth}
         \centering
         \includegraphics[width=\textwidth]{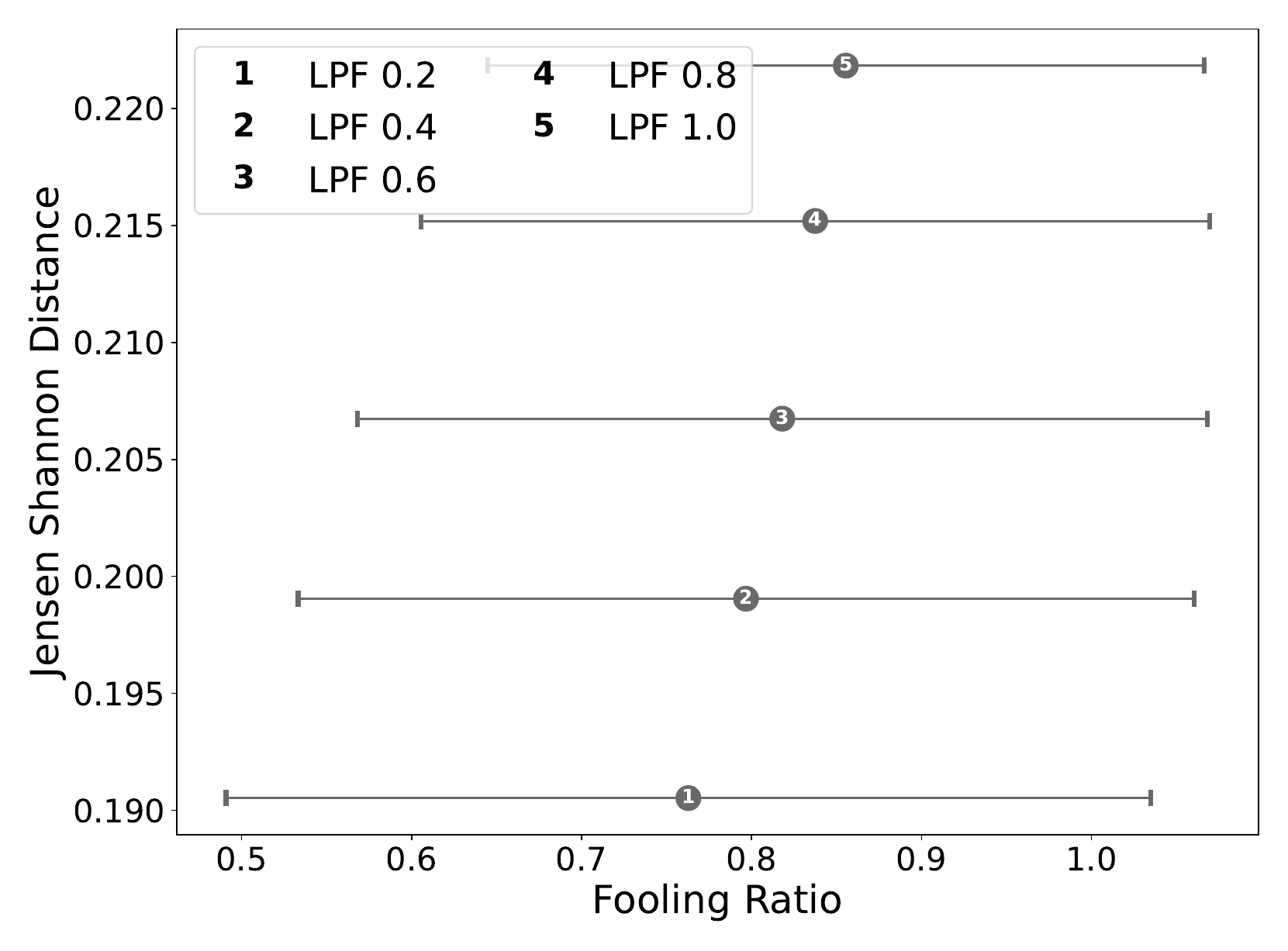}
         \caption{MIMIC-IV Mortality Model}
         \label{fig:JSD_LPF_MIMICIV}
     \end{subfigure}
     \hfill
     \begin{subfigure}[b]{0.45\textwidth}
         \centering
         \includegraphics[width=\textwidth]{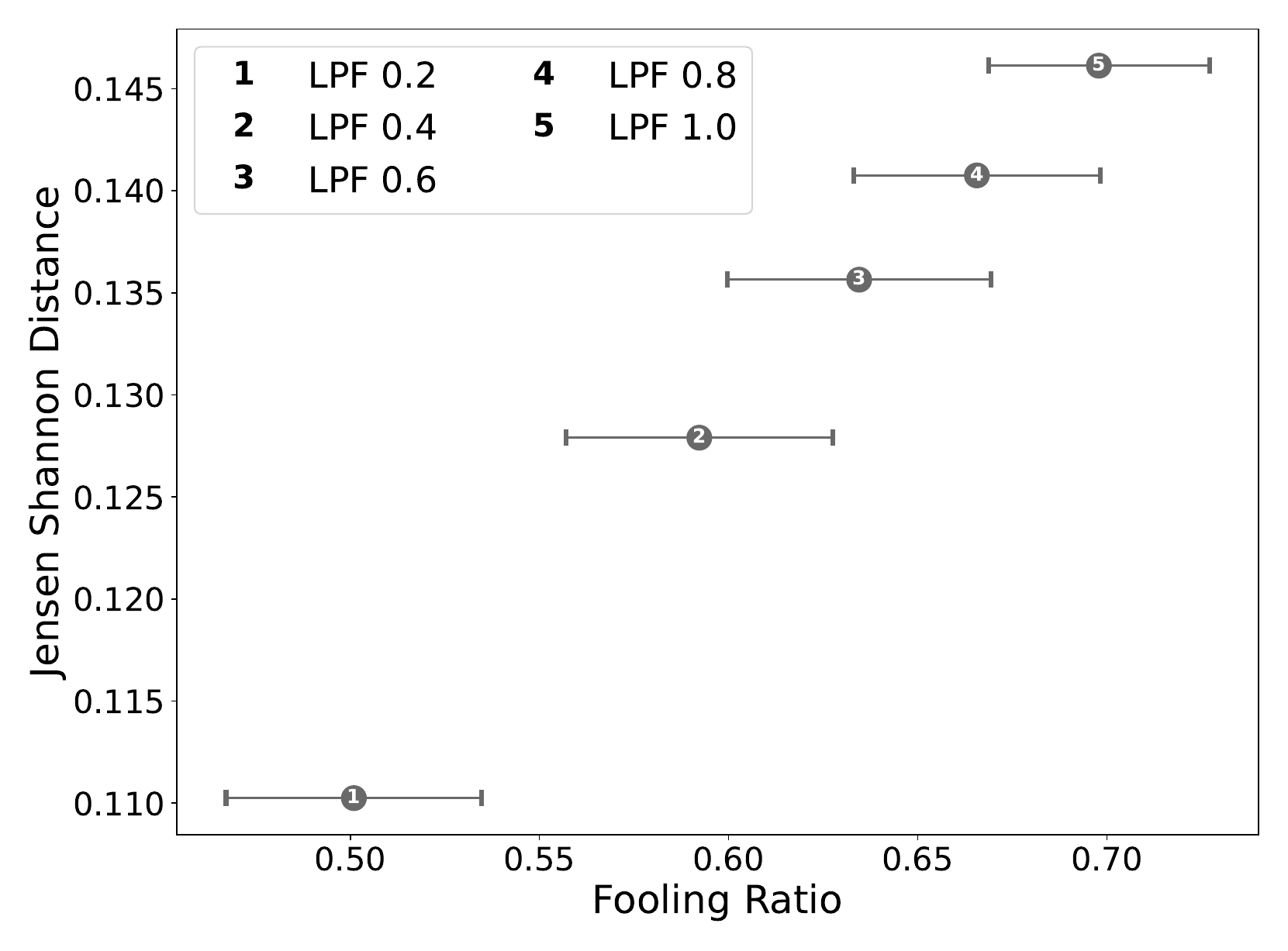}
         \caption{MNIST784 Model}
         \label{fig:JSD_LPF_MNIST784}
     \end{subfigure}
     \hfill
     \begin{subfigure}[b]{0.45\textwidth}
         \centering
         \includegraphics[width=\textwidth]{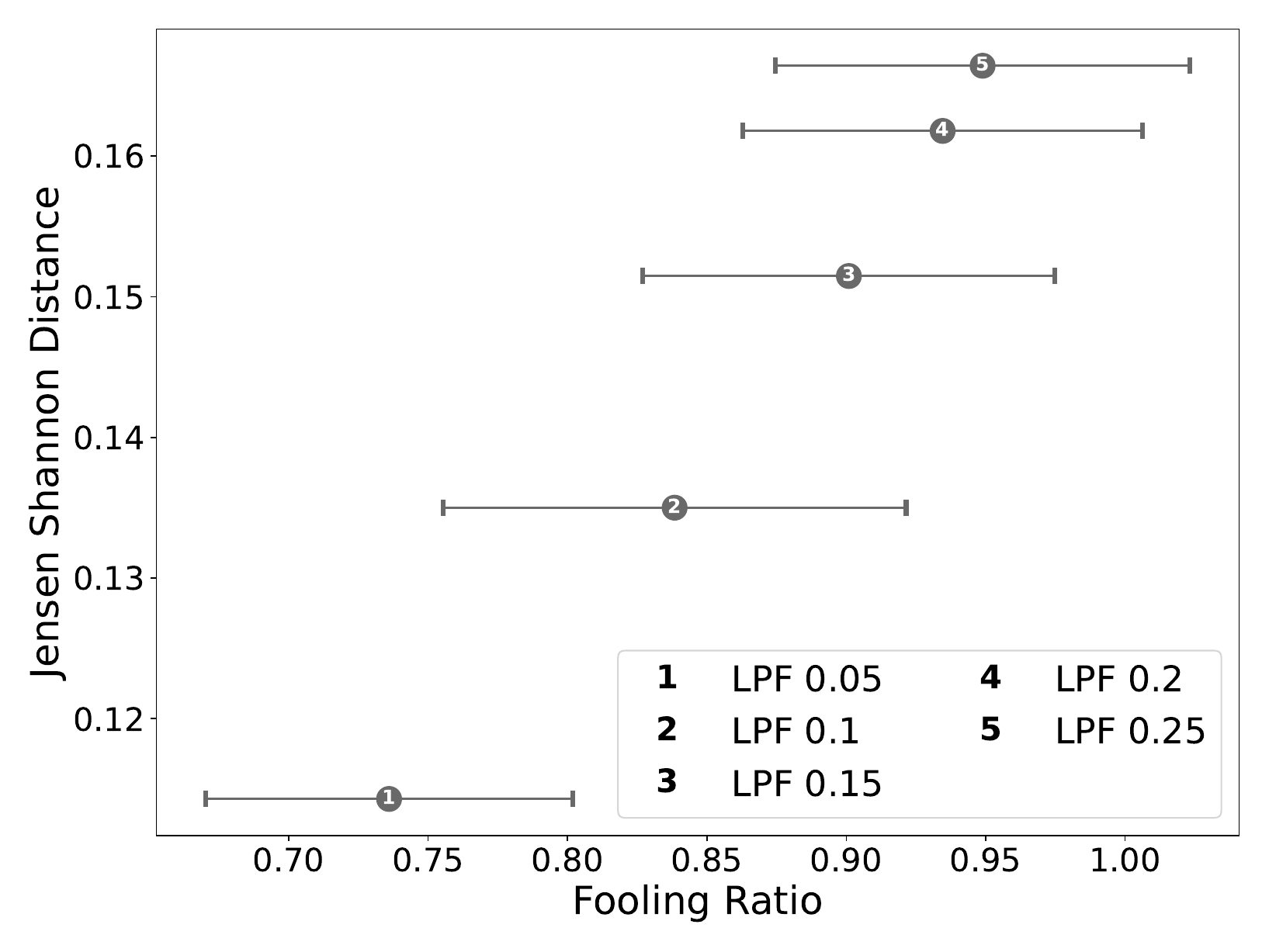}
         \caption{HAR Model}
         \label{fig:JSD_LPF_HAR}
     \end{subfigure}

        \caption{Average Jensen-Shannon Distances between the initial distributions and the adversarial distributions for different attacks applied on the models.}
        \label{fig:JSD_LPF_App}
\end{figure}

\subsection{Difference in Correlation Matrices}

\begin{figure}[H]
     \centering
     \begin{subfigure}[b]{0.45\textwidth}
         \centering
         \includegraphics[width=\textwidth]{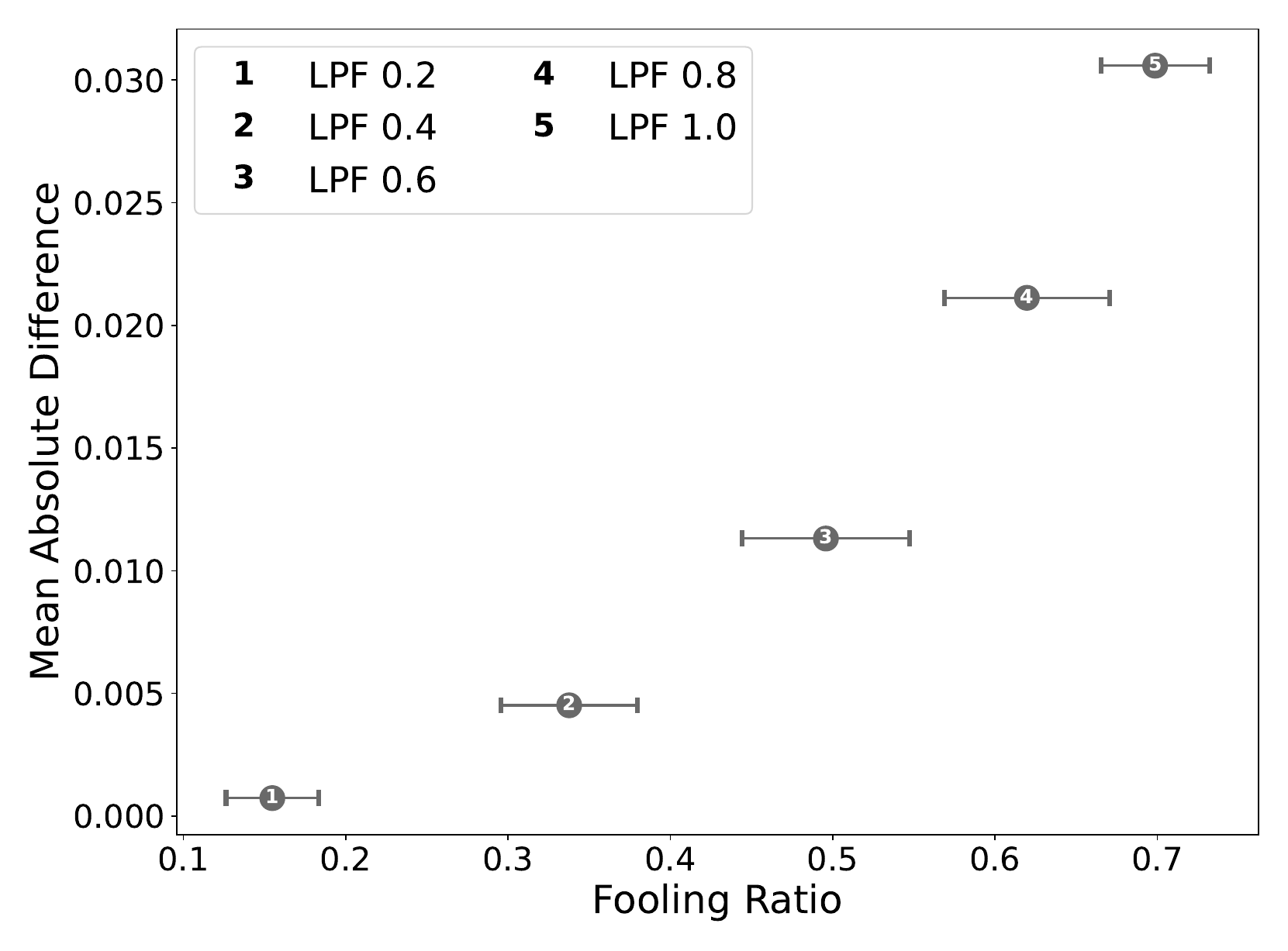}
         \caption{VBF Model}
         \label{fig:Mean_Difference_Correlation_LPF_VBF}
     \end{subfigure}
     \hfill
     \begin{subfigure}[b]{0.45\textwidth}
         \centering
         \includegraphics[width=\textwidth]{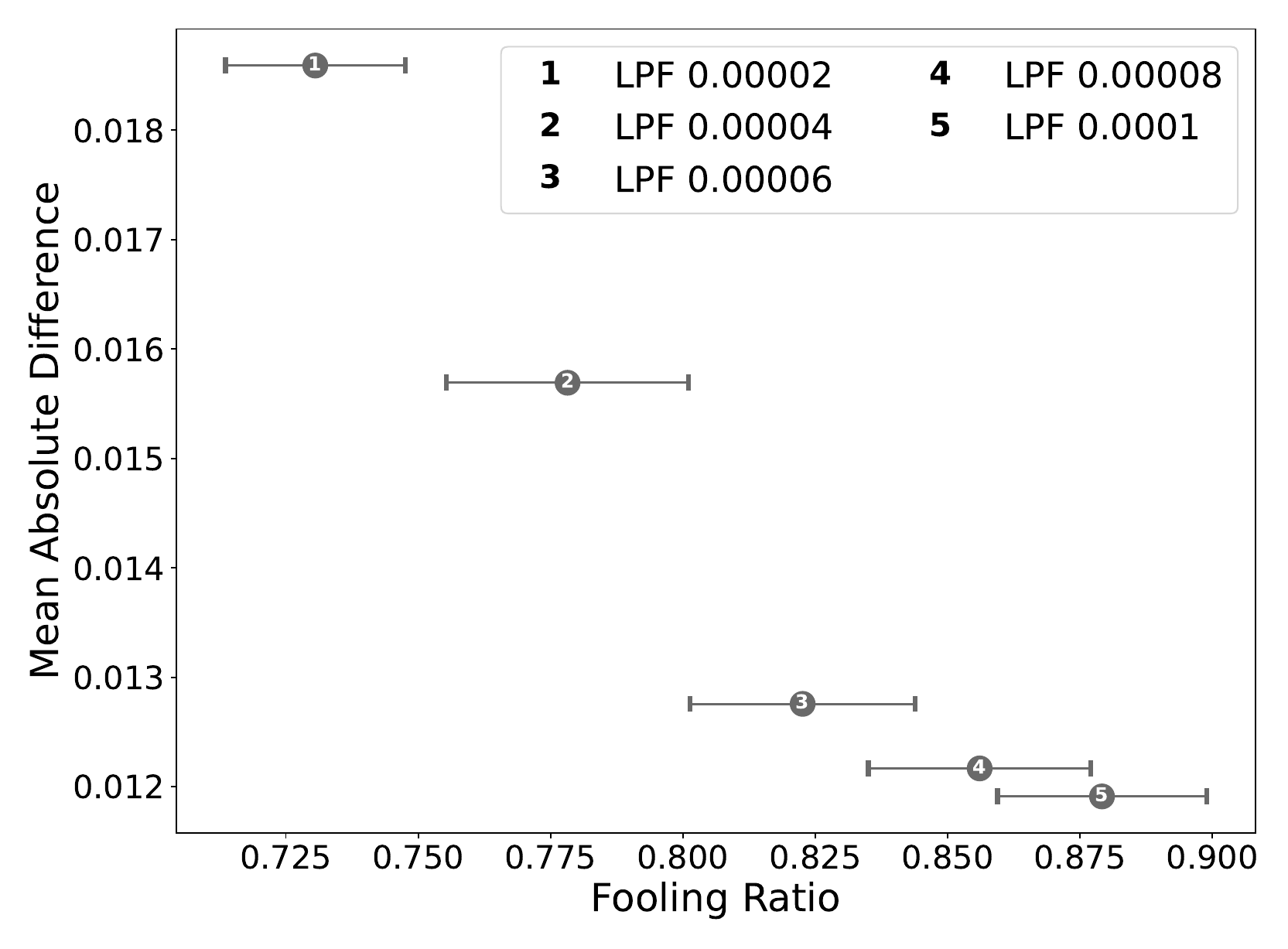}
         \caption{TopoDNN}
         \label{fig:Mean_Difference_Correlation_LPF_Topo}
     \end{subfigure}
     \hfill
     \begin{subfigure}[b]{0.45\textwidth}
         \centering
         \includegraphics[width=\textwidth]{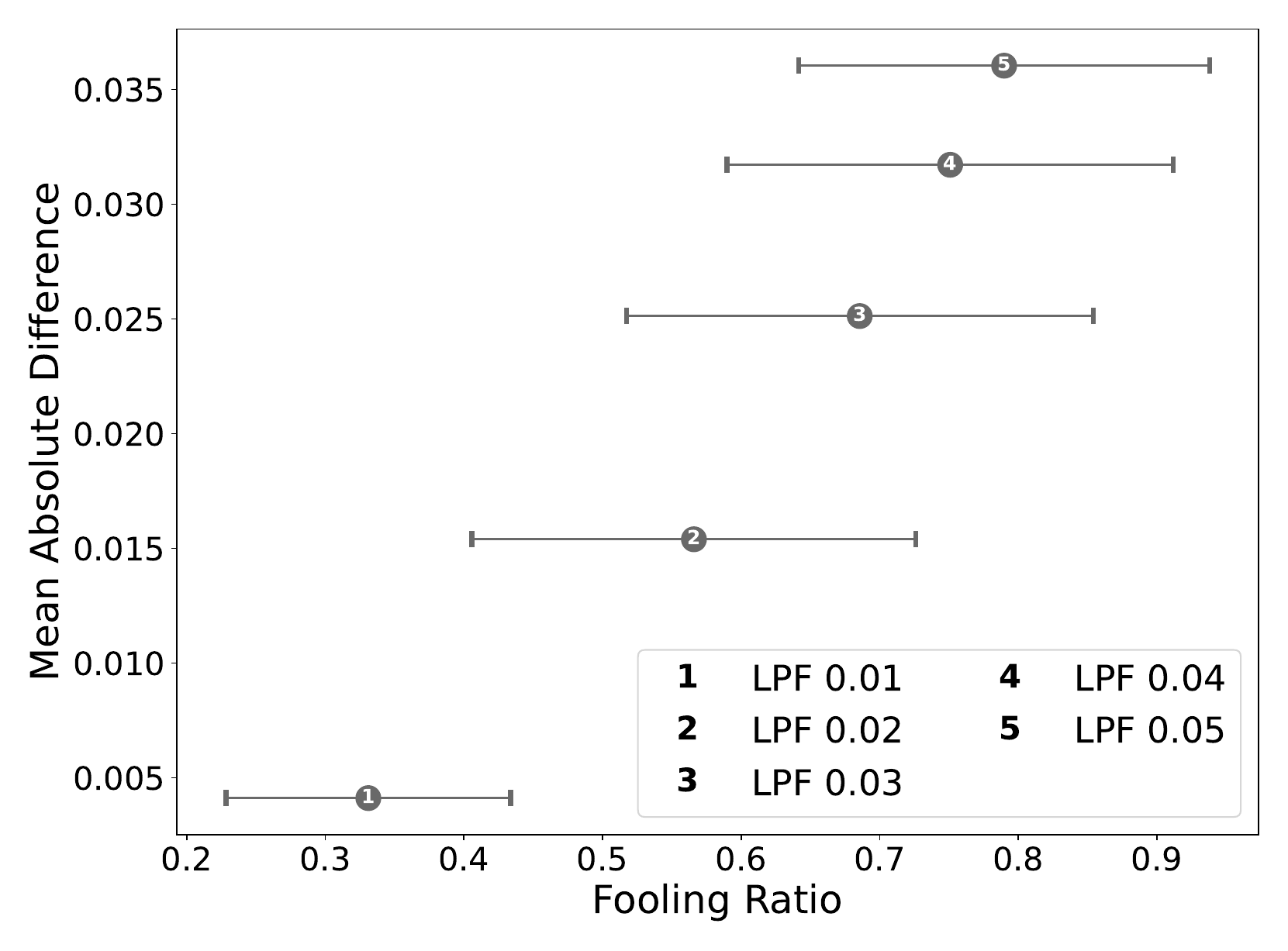}
         \caption{Rain in Australia Model}
         \label{fig:Mean_Difference_Correlation_LPF_Rain}
     \end{subfigure}
     \hfill
     \begin{subfigure}[b]{0.45\textwidth}
         \centering
         \includegraphics[width=\textwidth]{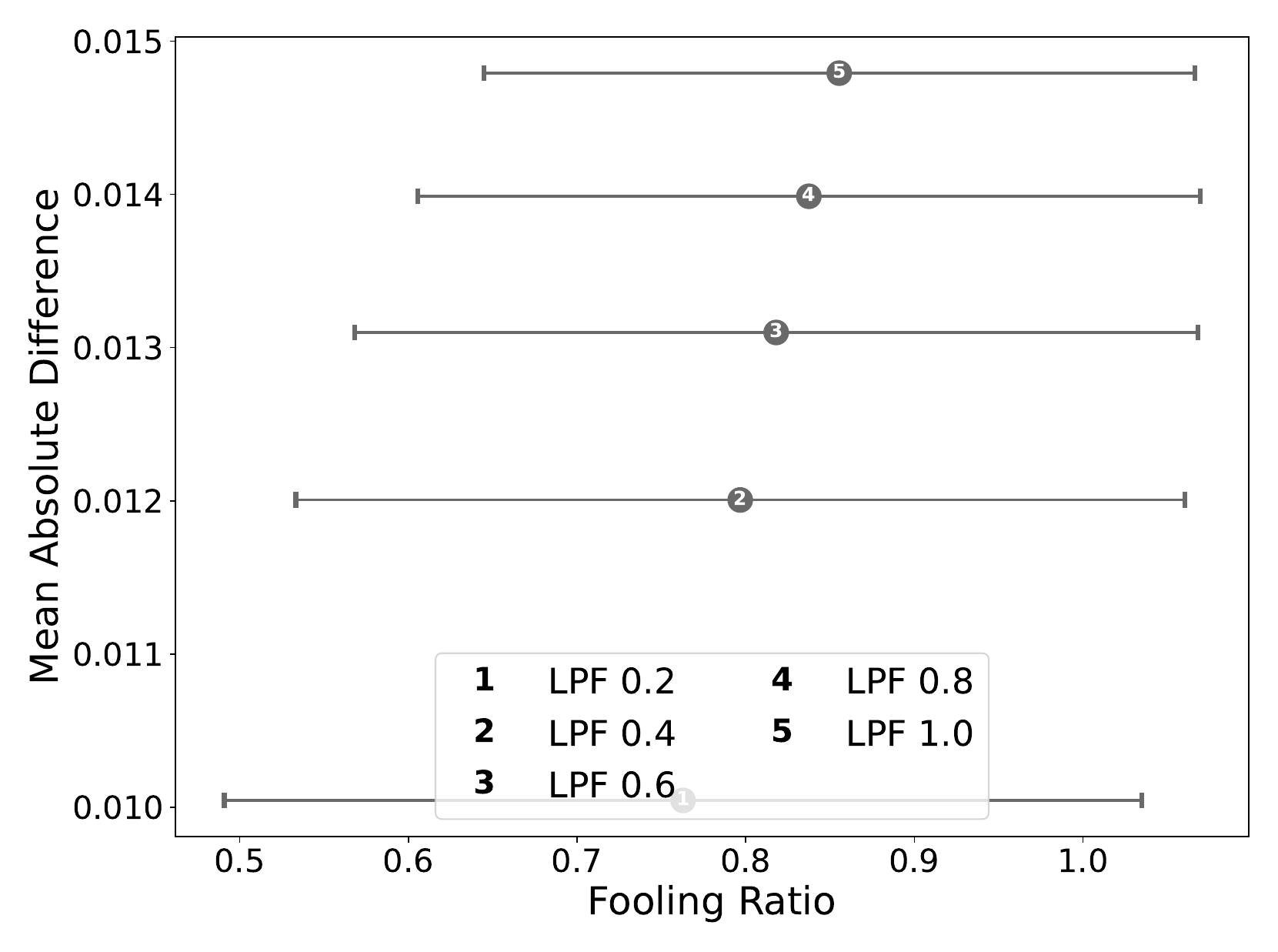}
         \caption{MIMIC-IV Mortality Model}
         \label{fig:Mean_Difference_Correlation_LPF_MIMICIV}
     \end{subfigure}
     \hfill
     \begin{subfigure}[b]{0.45\textwidth}
         \centering
         \includegraphics[width=\textwidth]{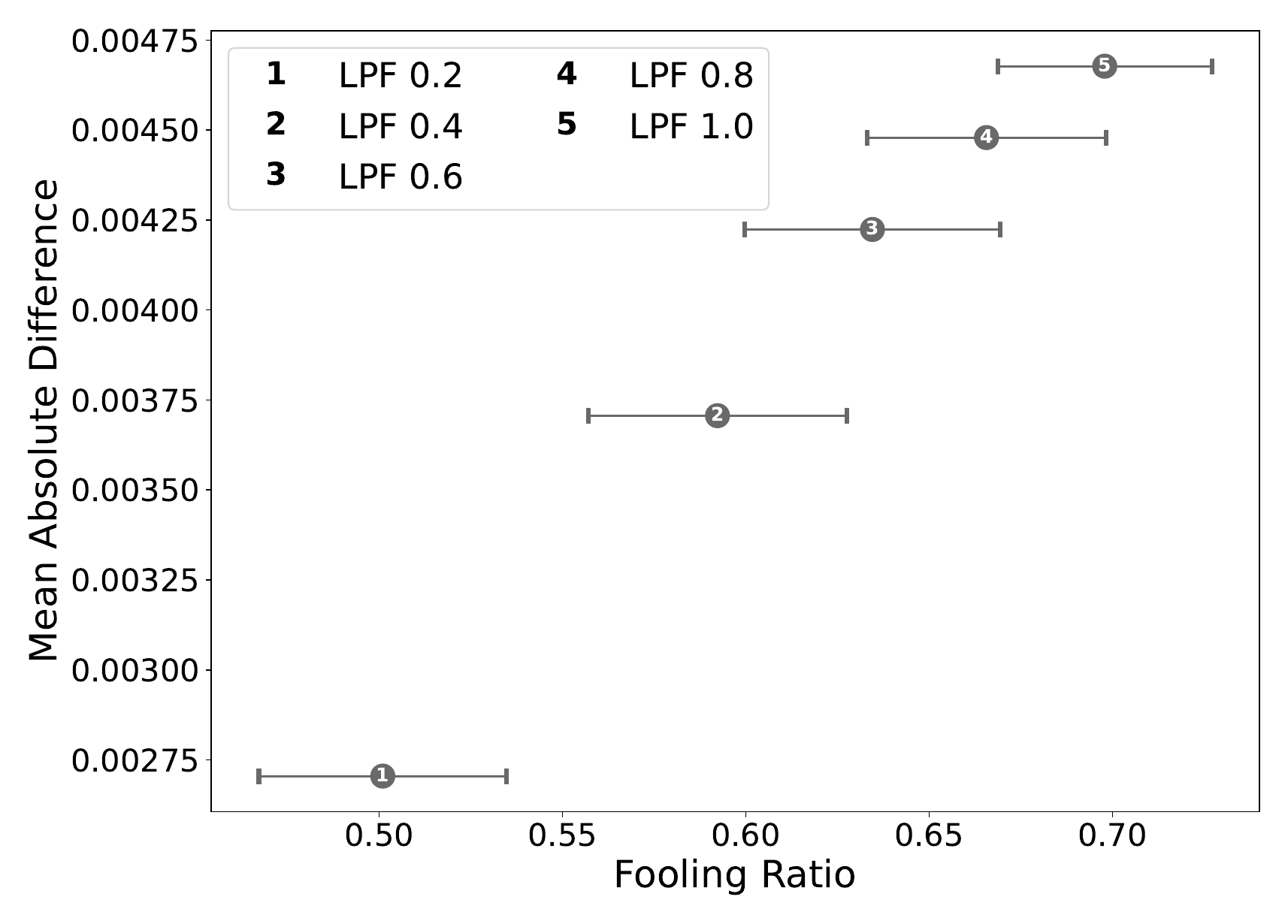}
         \caption{MNIST784 Model}
         \label{fig:Mean_Difference_Correlation_LPF_MNIST784}
     \end{subfigure}
     \hfill
     \begin{subfigure}[b]{0.45\textwidth}
         \centering
         \includegraphics[width=\textwidth]{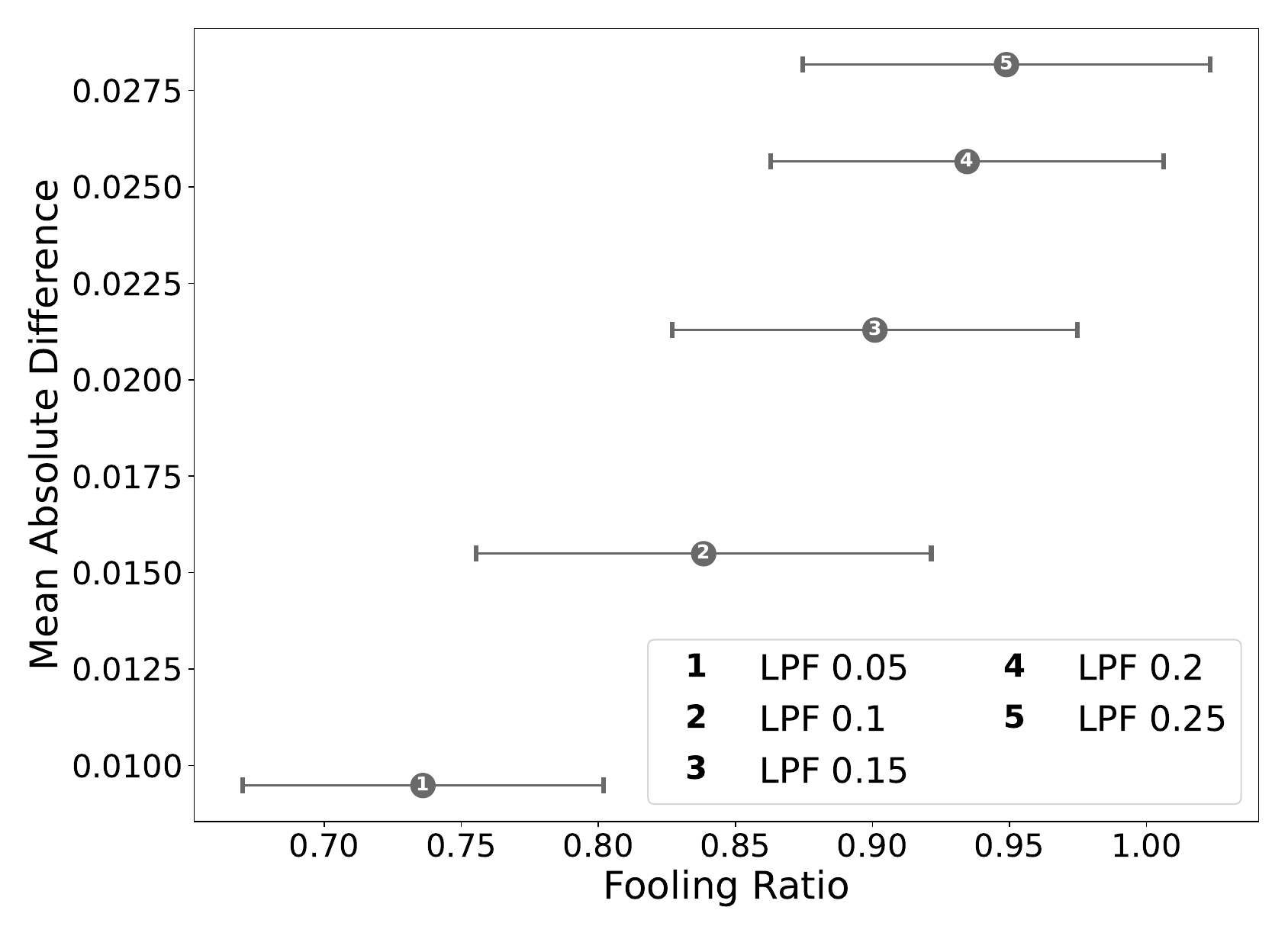}
         \caption{HAR Model}
         \label{fig:Mean_Difference_Correlation_LPF_HAR}
     \end{subfigure}

        \caption{Average absolute difference between the clean correlation matrices and the adversarial correlation matrices for different attacks applied on the models.}
        \label{fig:Mean_Difference_Correlation_LPF_App}
\end{figure}

\section{RDSA Pseudo-Code}

\begin{lstlisting}[language=Python, mathescape]
def get_vars_shuffle(input, nVars):
    return random.sample(range(0, len(input), nVars))

def get_frequencies(data, nBins):
    for each inputFeature:
        # Create finely binned histogram (with #bins = nBins) for the given data
        # Calculate frequencies for the bins
    return frequencies, binEdges

def sample_with_frequencies(inputFeature, frequency, binEdges):
    # Take frequencies and binEdges on the given input to sample randomly
    # But according to the underlying probabilities / frequencies
    return sampledValue

def random_distribution_shuffle_attack(input, inputLabel, frequencies, binEdges, nVars, model):
    varsToShuffle = get_vars_shuffle(input, nVars)
    adv = input

    for s in range(max_steps):
        for v in varsToShuffle:
            adv[v] = sample_with_frequencies(adv[v], frequencies[v], binEdges[v])
        if model.predict(adv) != inputLabel:
            return adv

    return None

if __name__ == "__main__":
    data = entireDataSet    # E.g. Test Dataset
    nBins = 1000
    nVars = 5

    model = load_model()
    frequencies, binEdges = get_frequencies(data, nBins)

    advs = []
    for i in range(len(data)):
        adv = random_distribution_shuffle_attack(data[i].input,
                                                 data[i].inputLabel,
                                                 frequencies, 
                                                 binEdges,
                                                 nVars,
                                                 model)
        advs.append(adv)
\end{lstlisting}
\end{document}